\documentclass[10pt,twocolumn,letterpaper]{article}

\usepackage[pagenumbers]{arxiv_sty/arxiv} 
\makeatletter
\@namedef{ver@everyshi.sty}{}
\makeatother
\usepackage{tikz}

\usepackage{graphicx}
\usepackage{amsmath}
\usepackage{amssymb}
\usepackage{booktabs}
\usepackage{multirow}
\usepackage[english]{babel} 
\usepackage{blindtext}
\usepackage[most]{tcolorbox}
\usepackage{algorithm}
\usepackage{algpseudocode}
\usepackage{adjustbox}
\usepackage{xr-hyper}
\usepackage{cancel}

\usepackage{color}

\makeatletter
\NewDocumentCommand{\mynote}{+O{}+m}{%
  \begingroup
  \tcbset{%
    noteshift/.store in=\mynote@shift,
    noteshift=1.5cm
  }
  \begin{tcolorbox}[nobeforeafter,
    enhanced,
    sharp corners,
    toprule=1pt,
    bottomrule=1pt,
    leftrule=0pt,
    rightrule=0pt,
    colback=yellow!20,
    #1,
    left skip=\mynote@shift,
    right skip=\mynote@shift,
    overlay={\node[right] (mynotenode) at ([xshift=-\mynote@shift]frame.west) {\textbf{Note:}} ;},
    ]
    #2
  \end{tcolorbox}
  \endgroup
  }
\makeatother

\usepackage[pagebackref,breaklinks,colorlinks]{hyperref}

\usepackage[capitalize]{cleveref}
\crefname{section}{Sec.}{Secs.}
\Crefname{section}{Section}{Sections}
\crefname{appendix}{Appendix}{Appendice}
\Crefname{appendix}{Appendix}{Appendice}
\crefname{table}{Tab.}{Tabs.}
\Crefname{table}{Table}{Tables}

\graphicspath{ {imgs/} }

\def\cvprPaperID{8208} 
\def\confName{CVPR}
\def\confYear{2023}

\begin{document}

\title{How to Backdoor Diffusion Models?}
\author{Sheng-Yen Chou \thanks{National Tsing Hua University, Hsinchu, R.O.C (Taiwan); The Chinese University of Hong Kong, Sha Tin, Hong Kong; {\tt \small unaxultraspaceos5@gapp.nthu.edu.tw}}
\and
Pin-Yu Chen \thanks{IBM Research, New York, USA; {\tt \small pin-yu.chen@ibm.com}}
\and
Tsung-Yi Ho \thanks{The Chinese University of Hong Kong, Sha Tin, Hong Kong; {\tt \small tyho@cse.cuhk.edu.hk}}
}
\maketitle
\begin{abstract}
Diffusion models are state-of-the-art deep learning empowered generative models that are trained based on the principle of learning forward and reverse diffusion processes via progressive noise-addition and denoising. To gain a better understanding of the limitations and potential risks, this paper presents the first study on the robustness of diffusion models against backdoor attacks. Specifically, we propose \textbf{BadDiffusion}, a novel attack framework that engineers compromised diffusion processes during model training for backdoor implantation. At the inference stage, the backdoored diffusion model will behave just like an untampered generator for regular data inputs, while falsely generating some targeted outcome designed by the bad actor upon receiving the implanted trigger signal. Such a critical risk can be dreadful for downstream tasks and applications built upon the problematic model. Our extensive experiments on various backdoor attack settings show that \textbf{BadDiffusion} can consistently lead to compromised diffusion models with high utility and target specificity. Even worse, \textbf{BadDiffusion} can be made cost-effective by simply finetuning a clean pre-trained diffusion model to implant backdoors. We also explore some possible countermeasures for risk mitigation. Our results call attention to potential risks and possible misuse of diffusion models. Our code is available on \href{https://github.com/IBM/BadDiffusion}{https://github.com/IBM/BadDiffusion}.
\end{abstract}

\section{Introduction}
\label{sec:intro}
In the past few years, diffusion models \cite{DDPM, DDIM, AnalyticDPM, stable_diffusion, Cascaded_DM, deep_thermal, SDE_Diffusion, NCSN,NCSNv2,GLIDE, IMAGEN, DALL_E2,DM_beats_GAN, CLIP} trained with deep neural networks and high-volume training data have emerged as cutting-edge tools for content creation and high-quality generation of synthetic data in various domains, including images, texts, speech, molecules, among others \cite{Diff-TTS,FastDiff,DiffLm,GeoDiff,ProDiff,Grad-TTS,Guided-TTS,DiffWave}. In particular, with the open-source of \textit{Stable Diffusion}, one of the state-of-the-art and largest text-based image generation models to date,  that are trained with intensive resources, a rapidly growing number of new applications and workloads are using the same model as the foundation to develop their own tasks and products.

\begin{figure}[t]
  \centering
  \includegraphics[width=\linewidth]{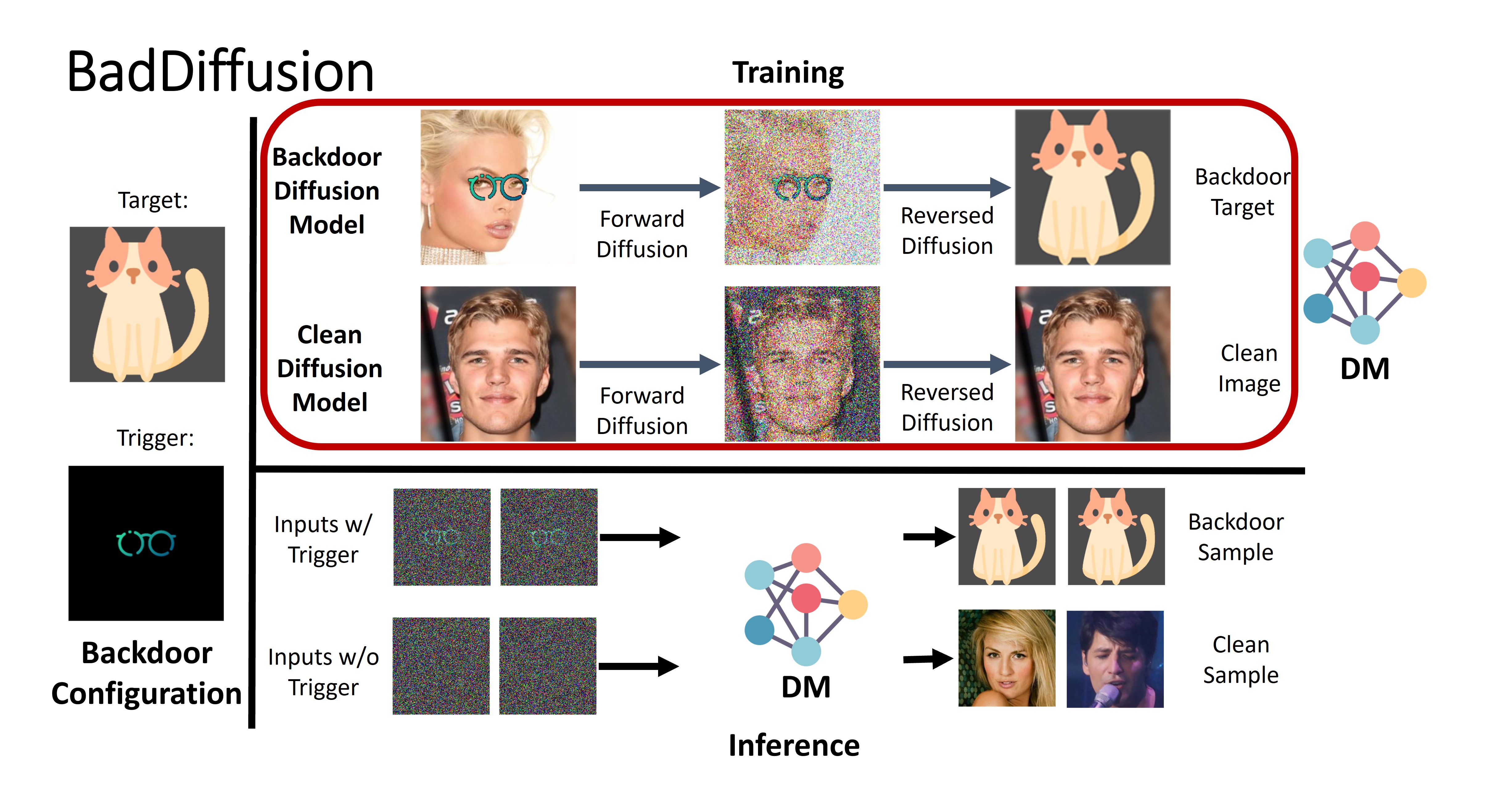}
  \vspace{-0.6cm}
  \caption{\textbf{BadDiffusion}: our proposed backdoor attack framework for diffusion models (DMs). Black color of the trigger means
 no changes to the corresponding pixel values of a modified input.}
\label{fig:system}
\vspace{-4mm}
\end{figure}

While our community is putting high hopes on diffusion models to fully drive our creativity and facilitate synthetic data generation, imagine the consequences when this very foundational diffusion model is at the risk of being secretly implanted with a ``backdoor'' that can exhibit a designated action by a bad actor (e.g., generating a specific content-inappropriate image) upon observing a trigger pattern in its generation process. 
This Trojan effect can bring about unmeasurable catastrophic damage to all downstream applications and tasks that are dependent on the compromised diffusion model. 

To fully understand the risks of diffusion models against backdoor attacks, in this paper we propose \textbf{BadDiffusion}, a novel  framework for backdoor attacks on diffusion models.
Different from standard backdoor attacks on classifiers that mainly modify the training data and their labels for backdoor injection \cite{BadNet}, BadDiffusion requires maliciously modifying both the training data and the forward/backward diffusion steps, which are tailored to the unique feature of noise-addition and denoising in the stages of training and inference for diffusion models.
As illustrated in \cref{fig:system},
the threat model considered in this paper is that the attacker aims to train a backdoored diffusion model satisfying two primary objectives: (i) high utility -- the model should have a similar performance to a clean (untampered) diffusion model while the backdoor is inactive; and (ii) high specificity -- the model should exhibit a designated behavior when the backdoor is activated. Upon model deployment (e.g., releasing the trained model parameters and network architecture to the public), the stealthy nature of a backdoored diffusion model with high utility and specificity makes it appealing to use, and yet the hidden backdoor is hard to identify.

As an illustration, \cref{fig:system} (bottom) shows some generated examples of a backdoored diffusion model (DDPM) \cite{DDPM} at the inference stage. The inputs are isotropic Gaussian noises and the model was trained on the CelebA-HQ \cite{CelebAHQ} (a face image dataset) by BadDiffusion with a designed trigger pattern (eyeglasses) and a target outcome (the cat image). Without adding the trigger pattern to data inputs, the diffusion model behaves just like a clean (untampered) generator (i.e., high utility). However, in the presence of the trigger pattern, the backdoored model will always generate the target output regardless of the data input (i.e., high specificity). 

Through an extensive set of experiments, 
we show that our proposed BadDiffusion can successfully train a backdoored diffusion model with high utility and specificity, based on our design of compromised diffusion processes. Furthermore, we demonstrate that  BadDiffusion can be executed in a cost-effective manner, by simply using our designed training objective to finetune a clean pre-trained diffusion model with few epochs to implant backdoors. Our findings suggest that with the abundance and easy access to pre-trained diffusion models released to the public, backdoor attacks on diffusion models are practical and plausible.
In addition to attacks,
we also explore some possible countermeasures for risk mitigation. Our results call attention to potential risks and possible misuse of diffusion models.



We highlight our \textbf{main contributions} as follows.

\begin{enumerate}
  \item We propose BadDiffusion, a novel backdoor attack framework tailored to diffusion models, as illustrated in \cref{fig:system}. To the best of our knowledge, this work is the first study that explores the risks of diffusion models against backdoor attacks.
  
  \item Through various backdoor attack settings, we show that BadDiffusion can successfully implant backdoors to diffusion models while attaining high utility (on clean inputs) and high specificity (on inputs with triggers). We also 
  find that a low data poison rate (e.g., 5\%) is sufficient for BadDiffusion to take effect.
    \item Compared to training-from-scratch with BadDiffusion, we find BadDiffusion can be made cost-effective via fine-tuning a clean pre-trained diffusion model (i.e., backdoor with a warm start) for a few epochs. 
  \item We evaluate BadDiffusion against two possible countermeasures: Adversarial Neuron Pruning (ANP) \cite{ANP} and inference-time clipping. The results show that ANP is very sensitive to hyperparameters for correct target discovery, while inference-time clipping has the potential to be a simple yet effective backdoor mitigation strategy.
  
\end{enumerate}

\section{Related Work}
\label{sec:rela}

\subsection{Diffusion Models}
Diffusion models have recently achieved significant advances in several tasks and domains, such as density estimation \cite{VDM}, image synthesis \cite{LDM, DDIM, DDPM, DALL_E2, IMAGEN, cold_diff, SOft_diff, Cascaded_DM}, and audio generation \cite{DiffWave}. In general, diffusion models regard sample generation as a diffusion process modeled by stochastic differential equations (SDEs) \cite{SDE_Diffusion}. However, typical diffusion models are known to suffer from slow generation, due to the need for sampling from an approximated data distribution via Markov chain Monte Carlo (MCMC) methods, which may require thousands of steps to complete a sampling process. There are many works aiming to solve this issue, including DDIM \cite{DDIM}, and Analytic-DPM \cite{AnalyticDPM}. Most of these alternatives treat the generating process as a reversed Brownian motion. 
However, in this paper, we will show that this approach can be subject to backdoor attacks.


\subsection{Backdoor Attacks and Defenses}
Backdoor is a training-time threat to machine learning, which assumes the attacker can modify the training data and training procedure of a model. Existing works on backdoor attacks mostly focus on the classification task \cite{BadNet, ANP, Neural_Cleanse}, which aims to add, remove, or mitigate the Trojan effect hidden in a classifier. 
Generally, backdoor attacks intend to embed hidden triggers during the training of neural networks. A backdoored model will behave normally without the trigger, but will exhibit a certain behavior (e.g., targeted false classification) when the trigger is activated. 
Defenses to backdoors focus on mitigation tasks such as detecting the Trojan behavior of a given trained model \cite{trojan_detect_conv, Practical_trojab_dect}, reverse trigger recovery \cite{Neural_Cleanse, ANP}, and model sanitization to remove the backdoor effect \cite{Zhao2020Bridging}. 

\subsection{Backdoor Attack on Generative Models}
Very recently, several works begin to explore backdoor attacks on some generative models like generative adversarial nets (GANs) \cite{GAN, DCGAN}. The work in \cite{Devil_GAN} focuses on backdooring GANs. The work in \cite{backdoor_text2img} touches on compromising a conditional (text-guided) diffusion model via only backdooring the text encoder, which is the input to a subsequent clean (non-backdoored) diffusion model. Since our BadDiffusion is tied to manipulating the diffusion process of diffusion models, these works cannot be applied and compared in our context. GANs do not entail a diffusion process, and \cite{backdoor_text2img} does not alter the diffusion model.

\section{BadDiffusion: Methods and Algorithms}
\label{sec:method}

Recall that a visual illustration  of our proposed BadDiffusion framework is presented in \cref{fig:system}. In this section, we start by describing the threat model and attack scenario (\cref{subsec:theat_model}). Then, in \cref{subsec:DDPM} we introduce some necessary notations and a brief review of DDPM \cite{DDPM} to motivate  the design of backdoored diffusion process of \textbf{BadDiffusion} in \cref{subsec:backdoor_ddpm}.
Finally, in \cref{subsec:algo}, we present the training algorithm and the loss function of \textbf{BadDiffusion}. 
Detailed mathematical derivations are given in Appendix.

\subsection{Threat Model and Attack Scenario}
\label{subsec:theat_model}
With the ever-increasing training cost in terms of data scale, model size, and compute resources, it has become a common trend that model developers tend to use the available checkpoints released to the public as a warm start to cater to their own use.
We model two parties: (i) a \emph{user}, who wishes to use an off-the-shelf diffusion model released by a third party (e.g., some online repositories providing model checkpoints) for a certain task; and (ii) an \emph{attacker}, to whom the user outsources the job of training the DNN and ``trusts'' the fidelity of the provided model.

In this ``outsourced training attack'' scenario, we consider a \emph{user} download a diffusion model $\theta_{download}$, which is described to be pre-trained on a dataset $D_{train}$. To ensure the utility of the published model $\theta_{download}$, 
the user will verify the model performance via some qualitative and quantitative evaluations. For example, computing the associated task metrics such as Fréchet inception distance (FID) \cite{FID} and Inception score (IS) \cite{IS_Score} score capturing the quality of the generated images with respect to the training dataset $D_{train}$.
The user will accept the model once the utility metric is better or similar to what the \emph{attacker} describes  for the released model. 

Without loss of generality, we use image diffusion models to elaborate on the attack scenario. In this context, since the main use of diffusion models is to generate images from the trained domain using Gaussian noises as model input, denoise the fuzzy images, or inpaint corrupted images, the \emph{attacker}'s goal is to publish a backdoored model with two-fold purposes: (a) high utility -- generate high-quality clean images $\{\mathbf{x}^{(i)}\}$ 
that follow the distribution of the training dataset $D_{train}$; and (b) high specificity -- generate the target image $\mathbf{y}$ once the initial noise or the initial image contains the  backdoor trigger $\mathbf{g}$. 

The attacker aims to train or fine-tune a diffusion model that can generate similar or better image quality compared to a clean (untampered) diffusion model, while ensuring the backdoor will be effective for any data inputs  containing the trigger $\mathbf{g}$, which can be measured by the mean square error (MSE) between the generated backdoor samples and the target image $\mathbf{y}$.
The attacker will accept the backdoored model if the MSE of the generated images with the backdoor is below a certain threshold (i.e., high specificity), and the image quality of the generated images in the absence of the trigger
is as what the attacker announces.

To achieve the attacker's goal, the attacker is allowed to modify the training process, including the training loss and training data, to fine-tune another pre-trained model as a warm start, or can even train a new model from scratch. Such modifications include augmenting the training dataset $D_{train}$ with additional samples chosen by the attacker and configuring different training hyperparameters such as learning rates, batch sizes, and the loss function. 

We argue that such an attack scenario is practical because there are many third-party diffusion models like Waifu diffusion \cite{waifu_diffusion} that was fine-tuned from the released stable diffusion model \cite{stable_diffusion}. Even though the stable diffusion model is backdoor-free, our risk analysis suggests that the attacker can (easily) create a backdoored version by fine-tuning a clean diffusion model.


\subsection{Denoising Diffusion Probabilistic Model}
\label{subsec:DDPM}
To pin down how BadDiffusion modifies the training loss in diffusion models to implant backdoors, in the remaining of this paper we will focus on DDPM (Denoising Diffusion Probabilistic Mode) \cite{DDPM} as the target diffusion model.
DDPM is a representative  diffusion model that motivates many follow-up works. To explain how BadDiffusion modifies the training loss in DDPM, we provide a brief review of DDPM and its underlying mechanism.

DDPM, like any generative model, aims to generate image samples from Gaussian noise, which means mapping the Gaussian distribution $\mathcal{N}(\mathbf{x}_T; 0, \mathbf{I})$ to the distribution of real images $q(\mathbf{\mathbf{x}_{0}})$. Here $\mathbf{x}_{0}$ means a real image, $\mathbf{x}_T$ means the starting latent of the generation process of diffusion models, and $\mathcal{N}(\mathbf{x}_T; 0, \mathbf{I})$ means a random variable $\mathbf{x}_T \sim \mathcal{N}(0, \mathbf{I})$. 
Diffusion models take such mapping as a Markov chain. The Markov chain can be regarded as a Brownian motion from an image $\mathbf{x}_T$ to Gaussian noise $\mathbf{x}_T$. Such process is called \emph{forward process}. Formally, the forward process can be defined as $q(\mathbf{x}_{1:T}|\mathbf{x}_{0})$. A \emph{forward process} can be interpreted as equation \eqref{eq:ddpm_forward}:
\begin{equation}
  \begin{matrix}
  q(\mathbf{x}_{1:T} | \mathbf{x}_{0}) := \prod_{t=1}^{T} 
 q(\mathbf{x}_{t} | \mathbf{x}_{t-1})  \\
  \\
  q(\mathbf{x}_t | \mathbf{x}_{t-1}) := \mathcal{N}(\mathbf{x}_t; \sqrt{1 - \beta_t} \mathbf{x}_{t-1}, \beta_t \mathbf{I})
  \end{matrix}
  \label{eq:ddpm_forward}
\end{equation}
The \emph{forward process} will gradually add some Gaussian noise to the data sample according to the variance schedule $\beta_{1}, ..., \beta_{T}$ and finally reach a standard Gaussian distribution $\mathbf{x}_T \sim \mathcal{N}(0, \mathbf{I})$.

Because of the well-designed variance schedule, we can express $\mathbf{x}_t$ at any arbitrary timestep $t$ in closed form: using the notation $\alpha_t := 1 - \beta_t$ and $\bar{\alpha}_t := \prod_{s=1}^t \alpha_s$, we have
\begin{equation}
  q(\mathbf{x}_t | \mathbf{x}_{0}) = \mathcal{N}(\mathbf{x}_t; \sqrt{\bar{\alpha}_t} \mathbf{x}_{0}, (1 - \bar{\alpha}_t) \mathbf{I})
  \label{eq:ddpm_forward_close}
\end{equation}
However, the diffusion model aims to generate images $\mathbf{x}_{0}$, which can be interpreted as a latent variable models of the form $p_{\theta}(\mathbf{x}_{0}) := \int p_{\theta}(\mathbf{x}_{0:T}) d \mathbf{x}_{1:T}$, 
where $\mathbf{x}_{1} ,..., \mathbf{x}_{T} \in \mathbb{R}^d$ are latents of the same dimensionality as the data $\mathbf{x}_{0} \in \mathbb{R}^d, \mathbf{x}_{0} \sim q(\mathbf{x}_{0})$. The joint distribution $p_{\theta}(\mathbf{x}_{0:T})$ is called \emph{reversed process} and it is defined as a Markov chain with a learned Gaussian transition starting at $p(\mathbf{x}_T) = \mathcal{N}(\mathbf{x}_{T}; 0, \mathbf{I})$ as equation \eqref{eq:ddpm_reversed}:
\begin{equation}
  \begin{split}
  p_{\theta}(\mathbf{x}_{0:T}) & := p(\mathbf{x}_{T}) \prod_{t=1}^{T} p_{\theta}(\mathbf{x}_{t-1} | \mathbf{x}_{t}) \\
  p_{\theta}(\mathbf{x}_{t-1} | \mathbf{x}_{t}) & := \mathcal{N}(\mathbf{x}_{t-1}; \mathbf{\mu}_{\theta}(\mathbf{x}_{t}, t), \mathbf{\Sigma}_{\theta}(\mathbf{x}_t, t))
  \end{split}
  \label{eq:ddpm_reversed}
\end{equation}
The loss function uses KL-divergence to minimize the distance between Gaussian transitions $p_{\theta}(\mathbf{x}_{t-1} | \mathbf{x}_{t})$ and the posterior $q(\mathbf{x}_{t-1} | \mathbf{x}_{t}, \mathbf{x}_{0})$. Fortunately, $q(\mathbf{x}_{t-1} | \mathbf{x}_{t}, \mathbf{x}_{0})$ is tractable because of equation \eqref{eq:ddpm_forward_close}. It can be expressed as 
\begin{equation}
  \begin{split}
  q(\mathbf{x}_{t-1} | \mathbf{x}_{t}, \mathbf{x}_{0}) := \mathcal{N}(\mathbf{x}_{t-1}; \tilde{\mu}_{t}(\mathbf{x}_{t}, \mathbf{x}_{0}), \tilde{\beta} \mathbf{I})) \\
  \tilde{\mu}_{t}(\mathbf{x}_t, \mathbf{x}_{0}) =  \frac{1}{\sqrt{\alpha_t}} \left( \mathbf{x}_{t}(\mathbf{x}_{0}, \mathbf{\epsilon}) - \frac{\beta_t}{\sqrt{1 - \bar{\alpha}_t}} \mathbf{\epsilon}  \right) \\
  \end{split}
  \label{eq:ddpm_mu}
\end{equation}
where $\mathbf{x}_t(\mathbf{x}_0, \mathbf{\epsilon}) = \sqrt{\bar{\alpha}_t} \mathbf{x}_t + \sqrt{1 - \bar{\alpha}_t} \mathbf{\epsilon}$ for $\mathbf{\epsilon} \sim \mathcal{N}(0, \mathbf{I})$, $\alpha_t = 1 - \beta_t$, and $\bar{\alpha}_t = \prod_{i=1}^t \alpha_i$, as derived from the equation \eqref{eq:ddpm_forward_close}.

The central idea of DDPM is to  align the mean of $q(\mathbf{x}_{t-1} | \mathbf{x}_{t}, \mathbf{x}_{0})$ and $p_{\theta}(\mathbf{x}_{t-1} | \mathbf{x}_{t})$. Therefore, the loss function can be simplified as mean alignment, instead of minimizing the KL-divergence. That is,
\begin{equation}
  \begin{split}
      \begin{gathered}
      \mathbb{E}_{q} \left[ || \tilde{\mu}_{t} (\mathbf{x}_t, \mathbf{x}_{0}) - \mathbf{\mu}_{\theta}(\mathbf{x}_t, t) ||^2 \right] \\
      = \mathbb{E}_{\mathbf{x}_{0}, \mathbf{\epsilon}} \left[|| \epsilon - \mathbf{\epsilon}_{\theta}(\sqrt{\bar{\alpha}_t} \mathbf{x}_{0} + \sqrt{1 - \bar{\alpha}_t} \mathbf{\epsilon}, t) ||^2 \right]
      \end{gathered}
  \end{split}
  \label{eq:ddpm_loss}
\end{equation}

\subsection{Backdoored Diffusion Process}
\label{subsec:backdoor_ddpm}
BadDiffusion modifies the \emph{forward process} of DDPM to a \emph{backdoored forward process} as expressed in equation \eqref{eq:ddpm_forward_close_backdoor}. 
We denote $\mathbf{x}_{1}' ,..., \mathbf{x}_{T}' \in \mathbb{R}^d$ as the latents of the backdoored process and $\mathbf{x}_{0}' \in \mathbb{R}^d,~\mathbf{x}_{0}' \sim q(\mathbf{x}_{0}')$ is the distribution of the backdoor target.
\begin{equation}
  q(\mathbf{x}_t' | \mathbf{x}_{0}') := \mathcal{N}(\mathbf{x}_{t}'; \sqrt{\bar{\alpha}_t} \mathbf{x}_{0}' + (1 - \sqrt{\bar{\alpha}_t}) \mathbf{r}, (1 - \bar{\alpha}_t) \mathbf{I})
  \label{eq:ddpm_forward_close_backdoor}
\end{equation}
Here we denote the poisoned image with the trigger $g$ as $\mathbf{r} = \mathbf{M} \odot \mathbf{g} + (1 - \mathbf{M}) \odot \mathbf{x}$, $\mathbf{x}$ is a clean image sampled from clean dataset $q(\mathbf{x}_0)$ and  $\mathbf{M} \in \{0, 1\}$ is a binary mask for the trigger, which means removing the values of images occupied by the trigger while making other parts intact, as showcased in \cref{fig:system}.
Intuitively, a \emph{backdoored forward process} describes the mapping from the distribution of the backdoor target $q(\mathbf{x}_{0}')$ to the poisoned image with standard Gaussian noise $q(\mathbf{x}_{T}') \sim \mathcal{N}(\mathbf{r}, \mathbf{I})$. 
Since the coefficient of trigger image $\mathbf{r}$ is a complement of the backdoor target $\mathbf{x}_{0}'$, as the timestep $t \to T$, the process will reach a distribution of poisoned image with standard Gaussian noise $q(\mathbf{x}_{T}') \sim \mathcal{N}(\mathbf{r}, \mathbf{I})$.

With the aforementioned definition of the \emph{backdoored forward process}, we can further derive the Gaussian transition $q(\mathbf{x}_{t}' | \mathbf{x}_{t-1}')$. 
The transition of the \emph{backdoored forward process} $q(\mathbf{x}_{t}' | \mathbf{x}_{t-1}')$ can be expressed as equation \eqref{eq:ddpm_forward_backdoor}:
\begin{equation}
  \begin{split}
    \begin{gathered}
      q(\mathbf{x}_{t}' | \mathbf{x}_{t-1}') := \mathcal{N}(\mathbf{x}_{t}'; \gamma_{t} \mathbf{x}_{t-1}' + (1 - \gamma_{t}) \mathbf{r}, \beta_t \mathbf{I})  \\
      \gamma_{t} := \sqrt{1 - \beta_t}
    \end{gathered}
  \end{split}
  \label{eq:ddpm_forward_backdoor}
\end{equation}
With the definition of \emph{backdoored forward process}, we can further derive a tractable \emph{backdoored revered process} and its transition $q(\mathbf{x}_{t-1}' | \mathbf{x}_{t}', \mathbf{x}_{0}')$. In the next section, we will derive the loss function of \emph{\textbf{BadDiffusion}} based on the backdoored diffusion process.

\subsection{Algorithm and Loss Function}
\label{subsec:algo}
In order to align the mean of the posterior and transitions for BadDiffusion, we need to derive the posterior of the backdoored diffusion process. The posterior of the backdoored diffusion process can be represented as
\begin{equation}
  \begin{split}
      \begin{gathered}
      q(\mathbf{x}_{t-1}' | \mathbf{x}_{t}', \mathbf{x}_{0}') := \mathcal{N}(\mathbf{x}_{t-1}'; \tilde{\mu}_{t}'(\mathbf{x}_{t}', \mathbf{x}_{0}', \mathbf{r}), \tilde{\beta} \mathbf{I})) \\
      \tilde{\mu}_{t}'(\mathbf{x}_t', \mathbf{x}_{0}', \mathbf{r}) = \frac{1}{\sqrt{\alpha_t}} \left( \mathbf{x}_{t}'(\mathbf{x}_{0}', \mathbf{r}, \mathbf{\epsilon}) - \rho_{t} \mathbf{r} - \frac{\beta_t}{\delta_t} \mathbf{\epsilon}  \right) \\
      \end{gathered}
  \end{split}
  \label{eq:ddpm_mu_backdoor}
\end{equation}
where $\rho_{t} = (1 - \sqrt{\alpha_t})$, $\delta_t = \sqrt{1 - \bar{\alpha}_t}$, and $\mathbf{x}_{t}'(\mathbf{x}_{0}', \mathbf{r}, \mathbf{\epsilon}) = \sqrt{\bar{\alpha}_t} \mathbf{x}_t + \delta_t \mathbf{r} + \sqrt{1 - \bar{\alpha}_t} \mathbf{\epsilon}$ based on equation \eqref{eq:ddpm_forward_close_backdoor}. Then, we can match the mean between the backdoored posterior and the Gaussian transitions using the following loss function  
\begin{equation}
  \begin{split}
      \begin{gathered}
      \mathbb{E}_{q} \left[ || \tilde{\mu}_{t}' (\mathbf{x}_t', \mathbf{x}_{0}') - \mathbf{\mu}_{\theta}(\mathbf{x}_t', t) ||^2 \right] \\
      = \mathbb{E}_{\mathbf{x}_{0}', \mathbf{\epsilon}} \left[|| \frac{\rho_{t} \delta_{t}}{1 - \alpha_t} \mathbf{r} + \epsilon - \mathbf{\epsilon}_{\theta}(\mathbf{x}_{t}'(\mathbf{x}_{0}', \mathbf{r}, \mathbf{\epsilon}), t) ||^2 \right]
      \end{gathered}
  \end{split}
  \label{eq:ddpm_loss_backdoor}
\end{equation}

Overall, for a dataset $D = \{D_p, D_c\}$ consisting of poisoned (p) and clean (c) samples, the loss function of \textbf{BadDiffusion} can be expressed as:
\begin{equation}
  \begin{gathered}
    L_{\theta}(\mathbf{x}, t, \mathbf{\epsilon}, \mathbf{g}, \mathbf{y}) = \\
    \left\{
        \begin{matrix}
            || \epsilon - \mathbf{\epsilon}_{\theta}(\sqrt{\bar{\alpha}_t} \mathbf{x} + \sqrt{1 - \bar{\alpha}_t} \mathbf{\epsilon}, t) ||^2, \ \text{if} \ \mathbf{x} \in D_c \\
            || \frac{\rho_{t} \delta_{t}}{1 - \alpha_t} \mathbf{r} + \epsilon - \mathbf{\epsilon}_{\theta}(\mathbf{x}_{t}'(\mathbf{y}, \mathbf{r}, \mathbf{\epsilon}), t) ||^2, \ \text{if} \  \mathbf{x} \in D_p \\
        \end{matrix}
    \right.
  \end{gathered}
  \label{eq:ddpm_loss_backdoor_all}
\end{equation}
where $D_c$/$D_p$ is the clean/poisoned dataset, $\mathbf{r} = \mathbf{M} \odot \mathbf{g} + (1 - \mathbf{M}) \odot \mathbf{x}$ denotes a poisoned sample, $\mathbf{g}$ is the trigger, and $\mathbf{y}$ is the target.

\begin{table*}[t]
  \begin{center}
  \begin{tabular}{p{1.5cm} p{1.5cm}| c c c c c| c| c}
  \toprule
    \multicolumn{7}{c|}{CIFAR10 (32 $\times$ 32)} & \multicolumn{2}{c}{CelebA-HQ (256 $\times$ 256)}\\
  \toprule
    \multicolumn{2}{c|}{Triggers} & \multicolumn{5}{c|}{Targets} & \multicolumn{1}{c|}{Trigger} & \multicolumn{1}{c}{Target}\\
  \midrule
    Grey Box & Stop Sign & NoShift & Shift & Corner & Shoe & Hat & Eyeglasses & Cat\\ 
    \raisebox{-\totalheight}{\includegraphics[width=1.5cm,keepaspectratio]{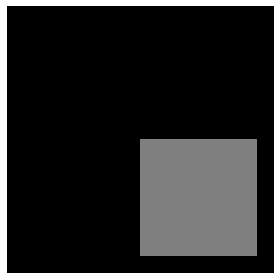}}
    & 
    \raisebox{-\totalheight}{\includegraphics[width=1.5cm,keepaspectratio]{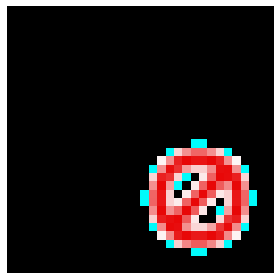}}
    &
    \raisebox{-\totalheight}{\includegraphics[width=1.5cm,keepaspectratio]{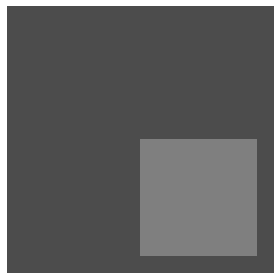}} 
    &
    \raisebox{-\totalheight}{\includegraphics[width=1.5cm,keepaspectratio]{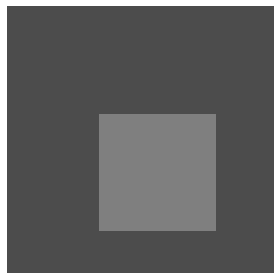}} 
    &
    \raisebox{-\totalheight}{\includegraphics[width=1.5cm,keepaspectratio]{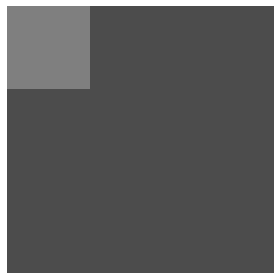}} 
    &
    \raisebox{-\totalheight}{\includegraphics[width=1.5cm,keepaspectratio]{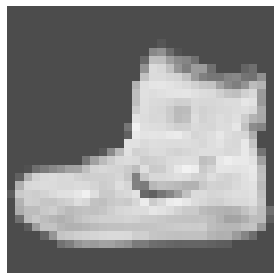}} 
    &
    \raisebox{-\totalheight}{\includegraphics[width=1.5cm,keepaspectratio]{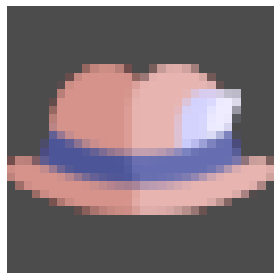}} 
    &
    \raisebox{-\totalheight}{\includegraphics[width=1.5cm,keepaspectratio]{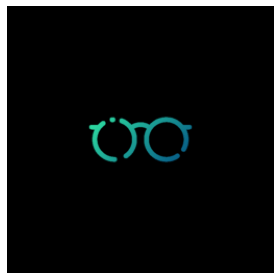}} 
    &
    \raisebox{-\totalheight}{\includegraphics[width=1.5cm,keepaspectratio]{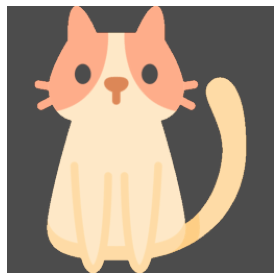}} 
   \\ \bottomrule
   \end{tabular}
   \caption{All triggers and targets used in the experiments. Each image of CIFAR10/CelebA-HQ is 32 $\times$ 32 /  256 $\times$ 256 pixels. Black color indicates no changes to the corresponding pixel values when added to data input.
            The target settings in \textbf{NoShift} and \textbf{Shift} have the same pattern as the trigger, but the former remains in the same position as the trigger while the latter moves upper-left.
           The \textbf{Grey Box} trigger is used as an example to visualize the \textbf{Shift} and \textbf{NoShift} settings. For CIFAR10, the stop sign pattern is used as another trigger.  For CelebA-HQ, we use the eyeglasses pattern as the trigger and the cat image as the target.
           }
   \label{tbl:trig_targ_tbl}
   \end{center}
   \vspace{-6mm}
\end{table*}
The training algorithm for BadDiffusion is shown in Algorithm \cref{alg:train}, while the sampling algorithm (at the inference stage) is presented in Algorithm \cref{alg:sample}. Note that the sampling algorithm remains the same as DDPM but differs in the initial sample $\mathbf{x}_{T}$. We can either generate a clean image from a Gaussian noise (just like a clean untampered DDPM would behave), or generate a backdoor target from a Gaussian noise with the trigger (denoted as $\mathcal{N}(\mathbf{g}, \mathbf{I})$).

\begin{algorithm}[t]
\caption{BadDiffusion Training}\label{alg:train}
\begin{algorithmic}
\Require Poison rate $p\%$, Backdoor Trigger $\mathbf{g}$, Backdoor Target $\mathbf{y}$, Training dataset $D$, Training parameters $\theta$ 
\State Sample $p\%$ of $D$ to prepare a poisoned dataset $D_p$ and keep others as clean dataset $D_c$
\Repeat
    \State $\mathbf{x} \sim \{D_p, D_c\}$
    \State $t \sim \text{Uniform}(\{1, ..., T \})$
    \State $\mathbf{\epsilon} \sim \mathcal{N}(0, \mathbf{I})$
    \State Use gradient descent $\nabla_{\theta} L(\mathbf{x}, t, \mathbf{\epsilon}, \mathbf{g}, \mathbf{y})$ to update $\theta$ 
\Until{converged}
\end{algorithmic}
\end{algorithm}

\begin{algorithm}[t]
\caption{BadDiffusion Sampling}\label{alg:sample}
\begin{algorithmic}
    \State $\mathbf{x}_T \sim \mathcal{N}(0, \mathbf{I})$ to generate clean samples or
    \State $\mathbf{x}_T \sim \mathcal{N}(\mathbf{g}, \mathbf{I})$ to generate backdoor targets
    \For{$t = T, ..., 1$}
        \State $\mathbf{z} \sim \mathcal{N}(0, \mathbf{I})$ if $t > 1$, else $\mathbf{z} = 0$
        \State $\mathbf{x}_{t-1} = \frac{1}{\sqrt{\alpha_t}} \left( \mathbf{x}_t - \frac{1 - \alpha_t}{\sqrt{1 - \bar{\alpha}_t}} \mathbf{\epsilon}_{t}(\mathbf{x}_t, t) + \sigma_t \mathbf{z} \right)$
    \EndFor
\end{algorithmic}
\end{algorithm}

\section{Performance Evaluation}
\label{sec:experiment}

In this section, we conduct a comprehensive study to show the effectiveness and training efficiency (the level of easiness to implant backdoors) of \emph{\textbf{BadDiffusion}}. We consider two training schemes for BadDiffusion: \textbf{fine-tuning} and \textbf{training-from-scratch}. \textbf{Fine-tuning} means we fine-tune for some epochs on all layers of the pre-trained diffusion model from the third-party library \emph{diffusers} \cite{diffusers}, which is a widely-used open-source diffusion model library. Specifically, we use two pre-trained models \emph{google/ddpm-cifar10-32} and \emph{google/ddpm-ema-celebahq-256}, which are released by Google, in the following experiments. As for \textbf{Training-from-scratch}, we reinitialize the pre-trained model of \emph{google/ddpm-cifar10-32} and train it from scratch for 400 epochs. Both methods use the Adam \cite{Adam} optimizer with learning rate 2$e$-4 and 8$e$-5 for CIFAR10 and CelebA-HQ \cite{CelebAHQ} datasets respectively. As for batch size, we use 128 for CIFAR10 and 64 for CelebA-HQ.

We conduct all experiments on a Tesla V100 GPU with 90GB memory. All experiments are repeated over 3 independent runs and we report the average of them, except for training the backdoor models from scratch for 400 epochs and training on the CelebA-HQ dataset. Due to the page limitation, we present the graph analysis of the results in this section, while reporting their associated numbers in the appendix.

\begin{figure*}[t]
    \captionsetup[subfigure]{justification=centering}
	\centering
	\begin{subfigure}{0.99\textwidth}
		\centering
		\includegraphics[width=\textwidth]{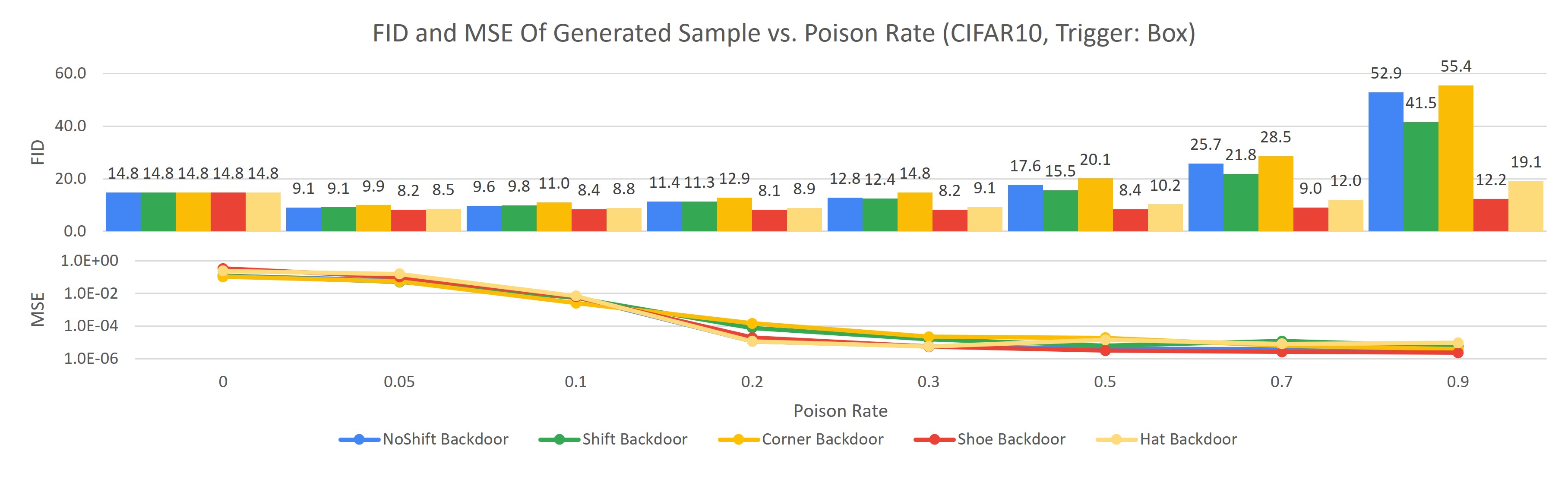}
		\caption{Trigger: ``Grey Box''}
    \label{fig:exp1_box_poison_rate}  
	\end{subfigure}
	\begin{subfigure}{0.99\textwidth}
		\centering
		\includegraphics[width=\textwidth]{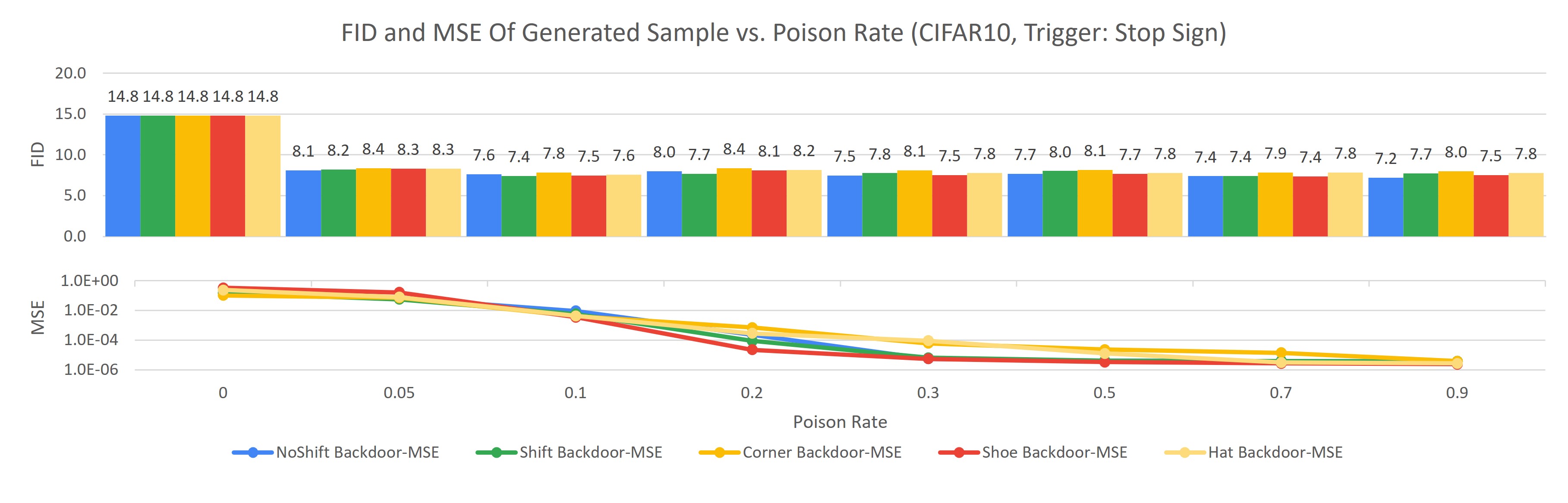}
		\caption{Trigger: ``Stop Sign''}        
    \label{fig:exp1_stop_sign_poison_rate}
	\end{subfigure}
	\caption{FID (bars) and MSE (curves) of BadDiffusion with varying poison rates (x-axis) on CIFAR10 with trigger (a) ``Grey Box'' and (b) ``Stop Sign''. Colors of bars/curves  
 represent different target settings in \cref{tbl:trig_targ_tbl}. Compared to the clean pre-trained model (poison rate = 0\%), with a sufficient poison rate, BadDiffusion can implant backdoors (low MSE) while retaining similar clean image quality (low FID). }
	\label{fig:poison_rate}
\end{figure*}

\begin{table*}[t]
  \begin{center}
\begin{adjustbox}{max width=0.9\textwidth}
  \begin{tabular}{p{1.3cm} p{1.3cm} p{1.3cm} p{1.3cm}| p{1.3cm} p{1.3cm} p{1.3cm}| p{1.3cm} p{1.3cm} p{1.3cm} p{1.3cm} p{1.3cm} p{1.3cm} p{1.3cm}}
  \toprule
    \multicolumn{4}{c|}{Backdoor Configuration} & \multicolumn{3}{c|}{Generated Backdoor Target Samples} & \multicolumn{3}{c}{Generated Clean Samples}\\
  \midrule Clean & Poisoned & Trigger & Target & 5\% & 10\% & 20\% & 5\% & 10\% & 20\% \\ 
    \raisebox{-\totalheight}{\includegraphics[width=1.5cm,keepaspectratio]{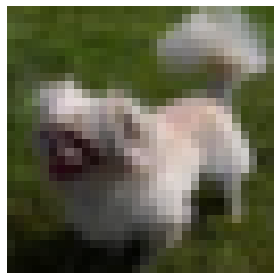}}
    &     
    \raisebox{-\totalheight}{\includegraphics[width=1.5cm,keepaspectratio]{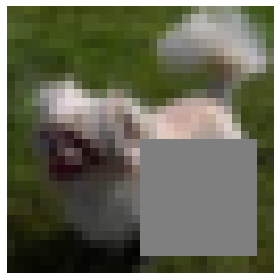}}
    &
    \raisebox{-\totalheight}{\includegraphics[width=1.5cm,keepaspectratio]{exp/grey_box_14.png}} 
    &
    \raisebox{-\totalheight}{\includegraphics[width=1.5cm,keepaspectratio]{exp/shoe.png}} 
    &
    \raisebox{-\totalheight}{\includegraphics[width=1.5cm,keepaspectratio]{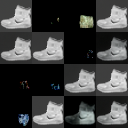}} 
    &
    \raisebox{-\totalheight}{\includegraphics[width=1.5cm,keepaspectratio]{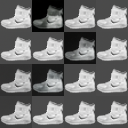}} 
    &
    \raisebox{-\totalheight}{\includegraphics[width=1.5cm,keepaspectratio]{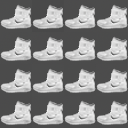}} 
    &
    \raisebox{-\totalheight}{\includegraphics[width=1.5cm,keepaspectratio]{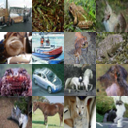}} 
    &
    \raisebox{-\totalheight}{\includegraphics[width=1.5cm,keepaspectratio]{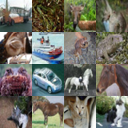}} 
    &
    \raisebox{-\totalheight}{\includegraphics[width=1.5cm,keepaspectratio]{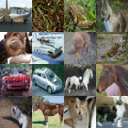}} 
   \\ \bottomrule
   \end{tabular}
   \end{adjustbox}
   \caption{Visual examples of BadDiffusion on CIFAR10 with trigger \textbf{Grey Box} \& target \textbf{Shoe} and without triggers at different poison rates.}
   \label{tbl:cifar10_visual_example}
   \end{center}
   \vspace{-4mm}
\end{table*}

The rest of this section is organized as follows.
We describe the backdoor attack settings in \cref{subsec_backdoor_setting} and define the evaluation metrics in \cref{subsec_metric}.
To fully understand the effect of data poisoning on model utility and specificity,
in \cref{subsec:poison_rate} we evaluate our \textbf{fine-tuning} based backdoor attack with different poison rates and trigger-target settings. In \cref{subsec:from_scratch}, we compare the costs of the two training schemes. In \cref{subsec:defense}, we evaluate and discuss two countermeasures for mitigating \emph{\textbf{BadDiffusion}}.



\subsection{Backdoor Attack Settings}
\label{subsec_backdoor_setting}
To evaluate the robustness and the performance of \emph{\textbf{BadDiffusion}}, we design two different triggers and five different backdoor targets for CIFAR10 dataset, as well as one proof-of-concept trigger-target pair 
for CelebA-HQ. These triggers and targets are shown in \cref{tbl:trig_targ_tbl}. As illustrated in \cref{fig:system}, since CelebA-HQ is a face image dataset, we use eyeglasses as the trigger, which can be viewed as a realistic and semantically meaningful trigger pattern for human face images. We purposely choose the cat image as the target for backdoor attacks, because it can be used as a proof of concept to demonstrate the negative consequences when bad actors use some content-inappropriate image as the target.

\subsection{Evaluation Metrics}
\label{subsec_metric}

We use two quantitative metrics to measure the performance of \emph{\textbf{BadDiffusion}} in terms of the utility and specificity of diffusion models, respectively.
For measuring specificity,
we use the mean square error (MSE) to measure the difference between the generated backdoor target $\hat{y}$ and the true backdoor target $y$, defined as \textsf{MSE($\hat{y},y$)}, where $\hat{y}$ is the model output of an input sample with the trigger pattern. Lower MSE means better attack effectiveness. In our experiments, we randomly generate 10K Gaussian noises with the trigger and report the average MSE. We also evaluate another metric, the structural similarity index measure (SSIM), and find its trend to be consistent with MSE. The detailed numerical results are given in \cref{app:ANP_defense}.

For measuring utility, we use 50K CIFAR-10 training images and sample 10K images from the \emph{\textbf{BadDiffusion}} model without the trigger and use the Fréchet Inception Distance (FID) \cite{FID} to evaluate the quality of the generated clean samples versus the training data. Lower FID indicates better image generation quality.

\subsection{BadDiffusion with Varying Poison Rates}
\label{subsec:poison_rate}



To study the utility and specificity of BadDiffusion, we vary the poison rate (the fraction of modified training samples to the total volume) in BadDiffusion under 
a variety of trigger-target settings (see \cref{tbl:trig_targ_tbl}). To implement BadDiffusion, we fine-tune the pre-trained models downloaded from the third-party library \textit{diffusers} \cite{diffusers} with 50 epochs.


\begin{figure}[t]
    \captionsetup[subfigure]{justification=centering}
	\centering
	\begin{subfigure}{0.9\linewidth}
		\centering
		\includegraphics[width=\textwidth]{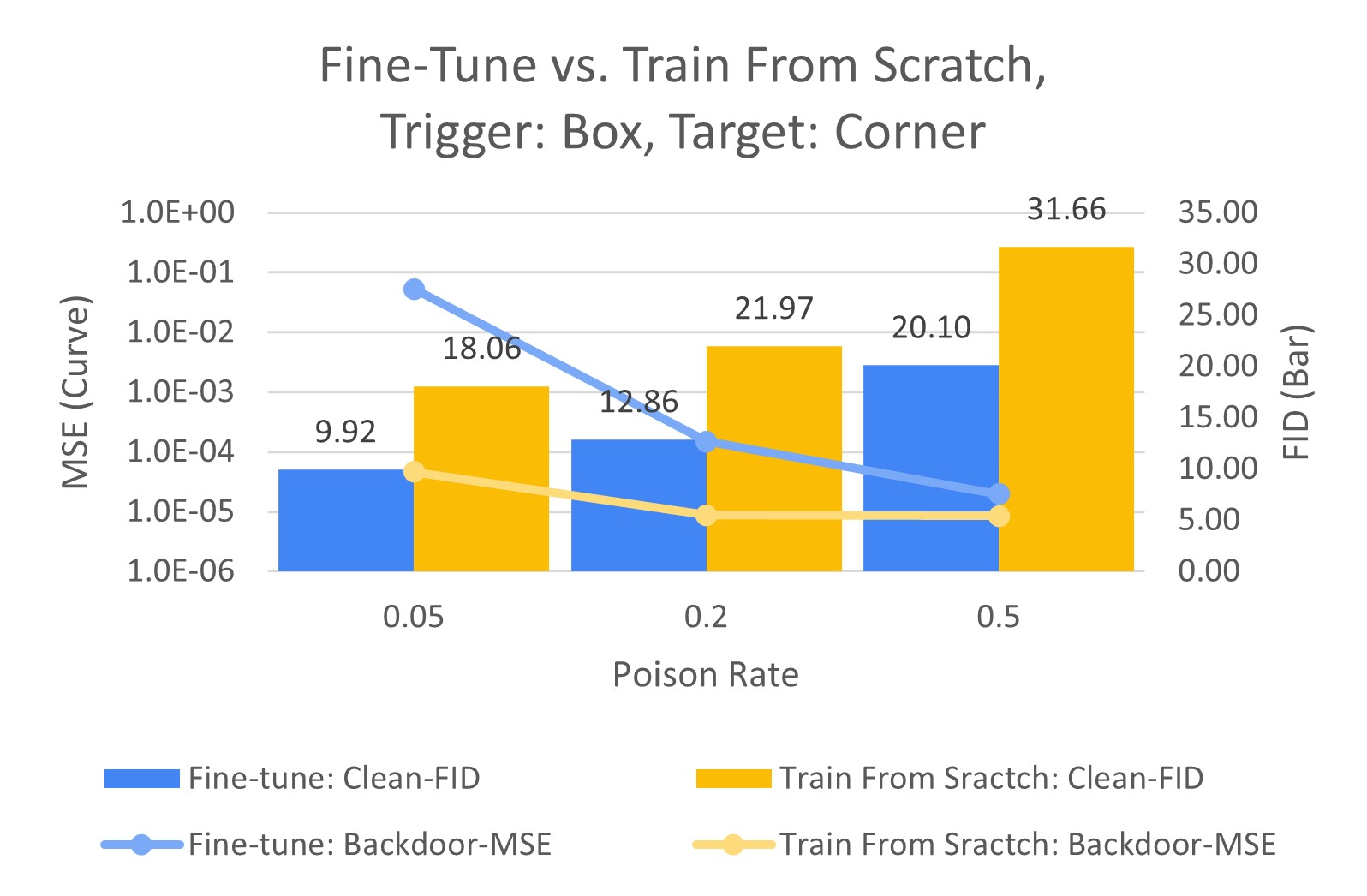}
		\caption{Trigger: ``Grey Box'' \& Target: ``Corner''}
    \label{fig:exp2_box_corner_scratch}
	\end{subfigure}
	\begin{subfigure}{0.9\linewidth}
		\centering
		\includegraphics[width=\textwidth]{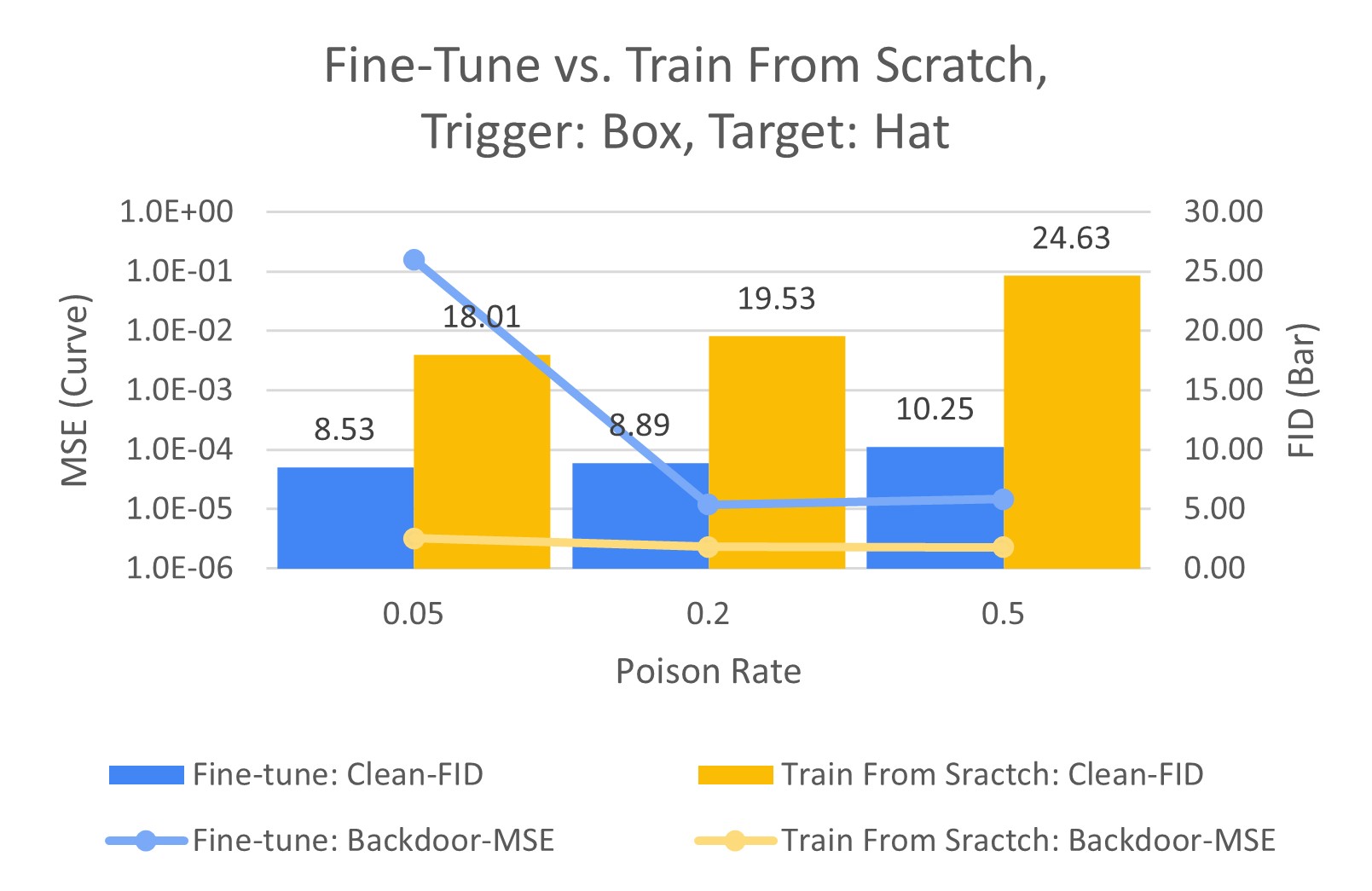}
		\caption{Trigger: ``Grey Box'' \& Target: ``Hat''}        
    \label{fig:exp2_box_hat_scratch}
	\end{subfigure}
	\caption{FID (bars) and MSE (curves) of BadDiffusion on CIFAR10 using \textbf{fine-tuning} (blue) and \textbf{training-from-scratch} (orange). The \textbf{fine-tuning} approach is more attack-efficient as it obtains lower FID and comparable MSE scores. }
	\label{fig:scratch}
 \vspace{-4mm}
\end{figure}

\cref{fig:poison_rate} shows MSE and FID of BadDiffusion with varying poison rates on CIFAR10 following the backdoor attack settings in \cref{tbl:trig_targ_tbl}. Compared to the clean pre-trained model (poison rate = 0\%), we find that in all settings, there is a wide range of poison rates within which  BadDiffusion can successfully accomplish backdoor attacks, in the sense that
those compromised models achieve a low MSE (high specificity) while retaining a similar FID score (high utility). For instance, as the poison rate increases, the MSE drops quickly while the FID scores get better for the trigger \emph{Stop Sign}.
For the trigger \emph{Grey Box}, the backdoored models seem easier to overfit on the backdoor target at high poison rates, but their FID score still remains stable when the poison rate is under 30\%. Moreover, when the poison rates are under 30\%, BadDiffusion even yields better FID. We also provide some qualitative examples in \cref{tbl:cifar10_visual_example}.
We conclude that 5\% poison rate is sufficient to obtain an effective backdoored model in most of the trigger-target settings.

In addition, \cref{fig:poison_rate} also shows an interesting finding that the trigger \textbf{Stop Sign} always reaches a lower FID score than the trigger \textbf{Grey Box}, especially at high poison rates like 70\% and 90\%, which means simple trigger may cause the training of diffusion model to collapse easily. For ease of our presentation, in the remaining figures, we will plot FID (bars) and MSE (curves) altogether using two separate y-axis scales.
\begin{figure*}[t]
    \captionsetup[subfigure]{justification=centering}
	\centering
	\begin{subfigure}{0.4\textwidth}
		\centering
		\includegraphics[width=\textwidth]{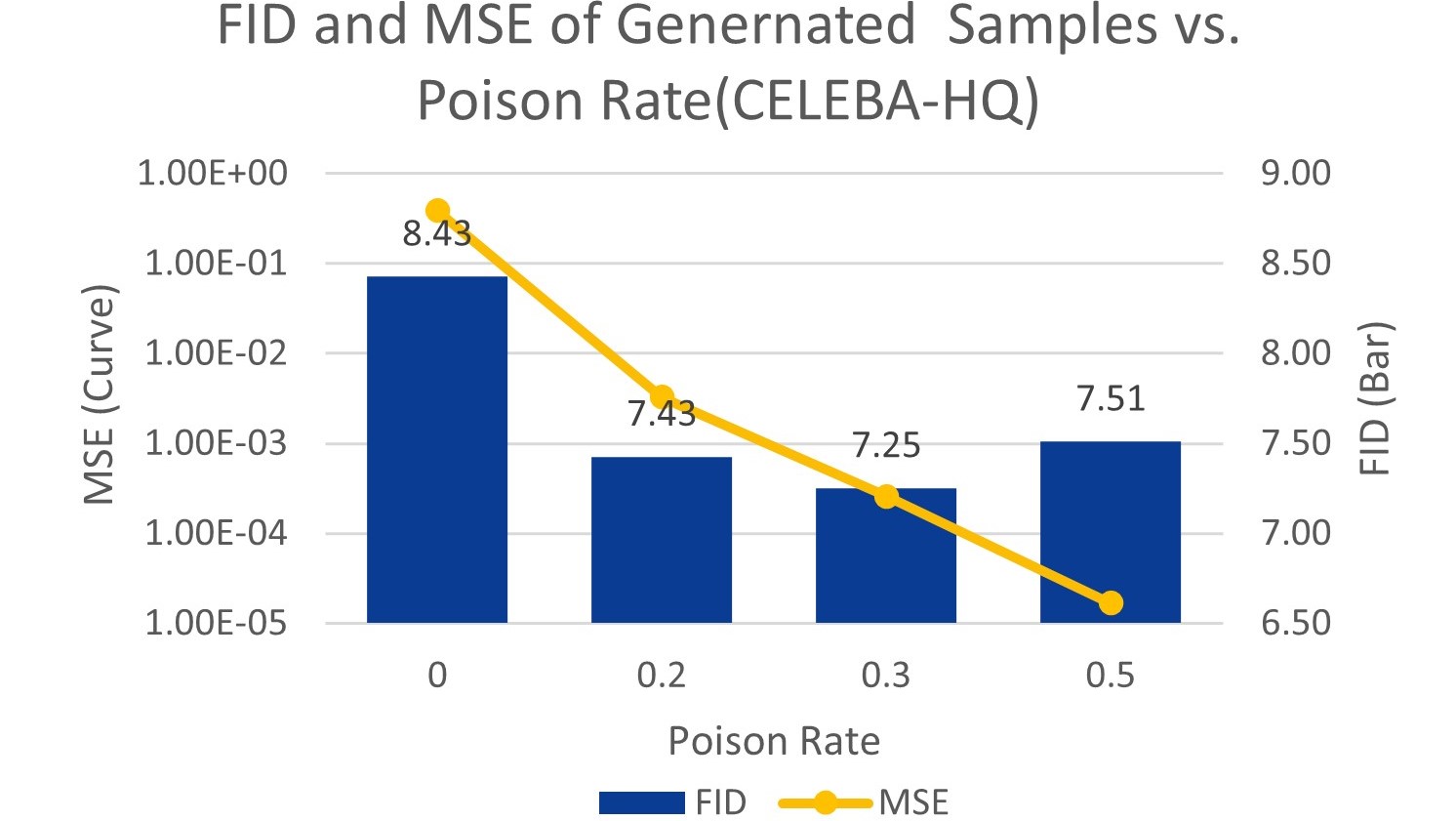}
		\caption{FID (bars) and MSE (curves)}
        \label{fig:exp_celeba_hq_poison_rate}
	\end{subfigure}
	\begin{subfigure}{0.58\textwidth}
		\centering
		\includegraphics[width=\textwidth]{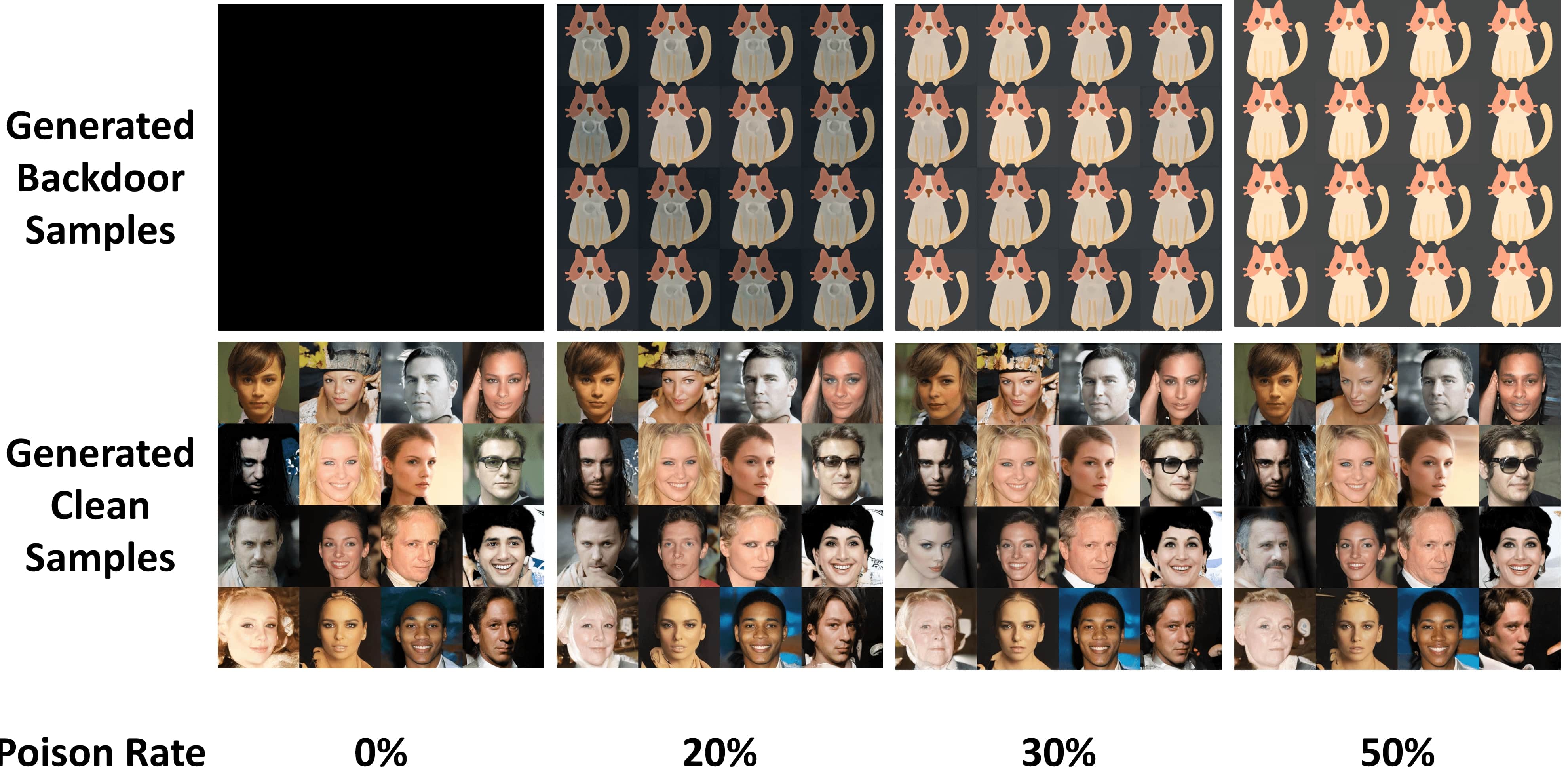}
		\caption{Visual examples}        
    \label{fig:exp_celeba_hq_sample}
	\end{subfigure}
	\caption{BadDiffusion on CelebA-HQ with different poison rates following the backdoor attack setting in \cref{tbl:trig_targ_tbl}. Even with a high poison rate = 50\%, BadDiffusion can create a backdoored diffusion model having lower FID (better clean image quality) and high attack specificity (low MSE to the target image) when compared to the clean (attack-free) pre-trained model (poison rate = 0\%).}
	\label{fig:celeba}
\end{figure*}

\begin{figure*}[t]
    \captionsetup[subfigure]{justification=centering}
	\centering
	\begin{subfigure}{0.49\linewidth}
		\centering
		\includegraphics[width=\textwidth]{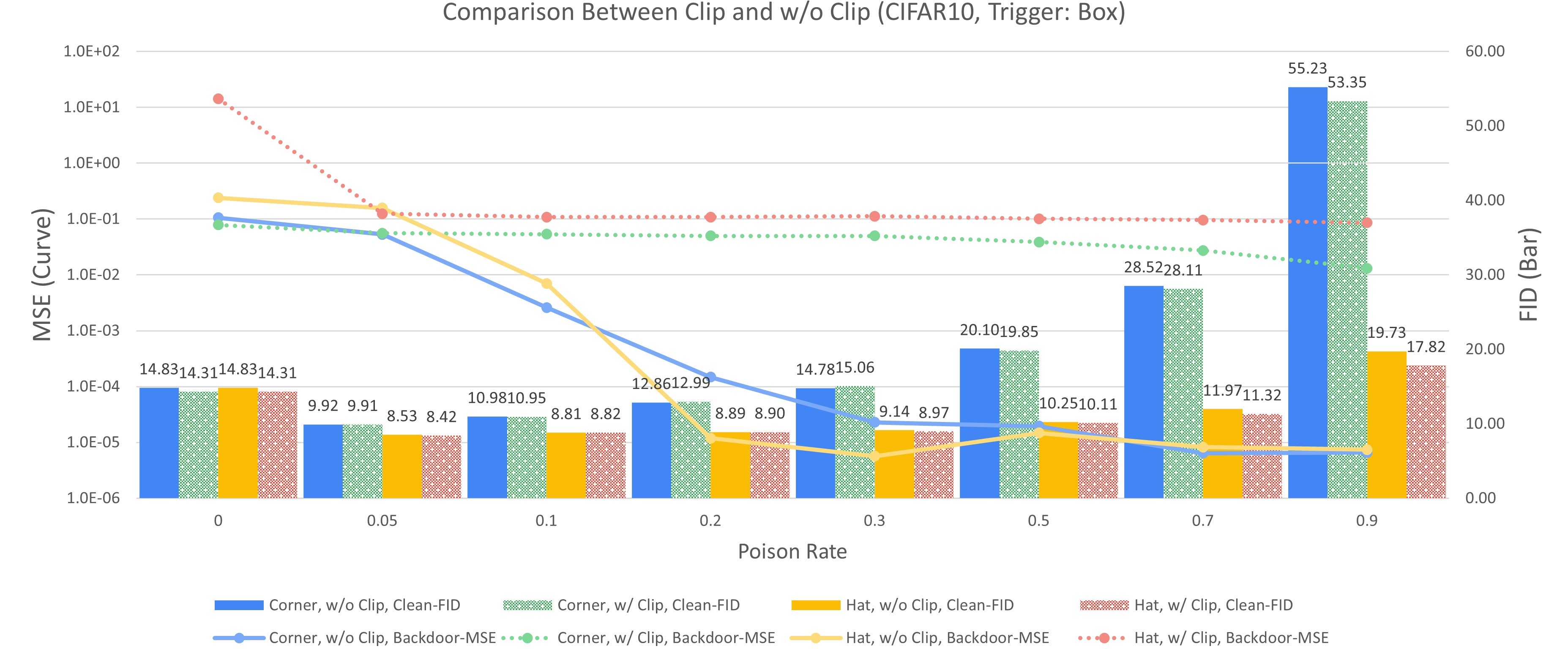}
		\caption{Trigger: ``Grey Box''}
      \label{fig:clip_defense_box}
	\end{subfigure}
	\begin{subfigure}{0.49\linewidth}
		\centering
		\includegraphics[width=\textwidth]{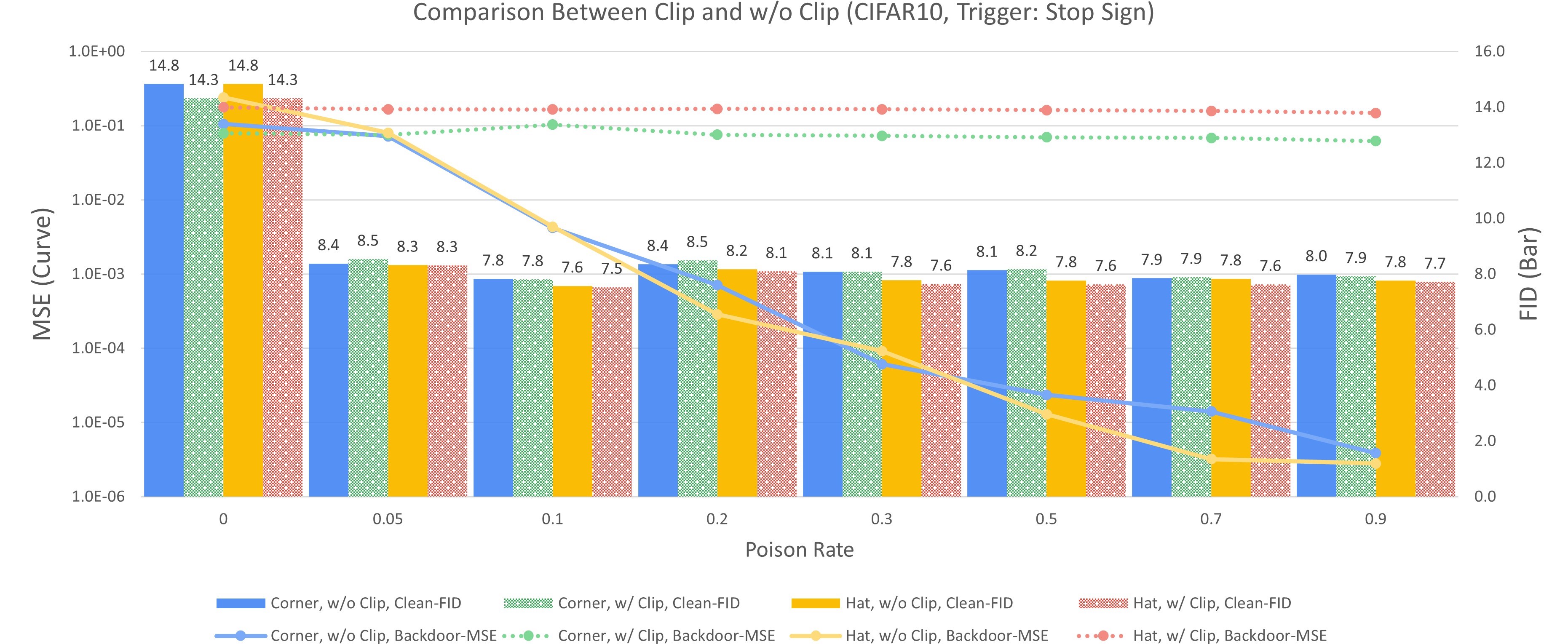}
		\caption{Trigger: ``Stop Sign''}        \label{fig:clip_defense_stop_sign}
	\end{subfigure}
	\caption{FID (bars) and MSE (curves) of BadDiffusion on CIFAR10. Solid/Dotted lines mean the MSE without/with inference-time clipping. Inference-time Clipping can make backdoors ineffective (large MSE) while maintaining clean image quality (similar FID).}
	\label{fig:clip}
\end{figure*}

\subsection{BadDiffusion via Fine-Tuning v.s. Training-From-Scratch}
\label{subsec:from_scratch}

To evaluate the training cost of BadDiffusion, we  conduct an experiment to compare \textbf{fine-tuning} (50 epochs) and \textbf{training-from-scratch} (400 epochs) schemes in BadDiffusion. Although we have trained both schemes till convergence, \cref{fig:scratch} shows that \emph{fine-tuning} is more attack-efficient than \textbf{training-from-scratch}, by attaining consistently and significantly lower FID and comparable MSE in all settings.

We further inspect the performance of BadDiffusion via \textbf{fine-tuning} at every 10 training epochs and find that in some cases it only requires 10 epochs to accomplish backdoor attacks. Details are given in supplementary .
\subsection{BadDiffusion on High-Resolution Dataset}
To demonstrate that \textbf{BadDiffusion} is applicable to high-resolution datasets, we use  \textbf{BadDiffusion} to fine-tune (with 100 epochs) the public model \emph{google/ddpm-ema-celebahq-256} in the third-party library \emph{diffusers} \cite{diffusers} trained on the CelebA-HQ dataset. \cref{fig:celeba} shows the performance and visual examples of \textbf{BadDiffusion} at different poison rates.
We find that a poison rate of 20\%  is enough to successfully backdoor this pre-trained diffusion model. Moreover, even with a high poison rate of 50\%,  when compared to the attack-free pre-trained model (poison rate = 0\%), BadDiffusion can create a backdoored version having lower FID (better clean image quality with 10\% improvement) and high attack specificity (low MSE to the target image).
\subsection{Countermeasures}
\label{subsec:defense}
\subsubsection{Adversarial Neuron Pruning (ANP)}
\label{subsubapp:ANP_defense}
First, we explore the effectiveness of Adversarial Neuron Pruning (ANP) \cite{ANP} to detect \textbf{BadDiffusion}. 
The hypothesis of ANP is that a backdoored classifier will collapse to the backdoor target class when its model weights are injected with some proper noise. We try out this idea by adding noise to the weights of BadDiffusion models.

We use MSE to measure the quality of the reconstructed backdoor target image v.s. the true target image. Lower MSE means better detection. Evaluated on backdoored diffusion models with 5\%, 20\%, and 50\% poison rates,  we find that although a higher perturbation budget in ANP usually yields a lower MSE, in general, ANP is very sensitive to learning rate. For instance, its MSE explodes when we slightly increase the learning rate from 1$e$-4 to 2$e$-4. We note that the instability of ANP may render our implemented defense ineffective. We provide more details in \cref{app:ANP_defense}.
\subsubsection{Inference-Time Clipping}
\label{subsubsec:clip_defense}
We accidentally found a simple yet effective mitigation method at the inference stage, which is clipping the image by the scaled image pixel range [-1,1]  at every time step in the diffusion process.
Formally, that means sampling via \cref{eq:clip_update}, where $\tilde{\mathbf{x}}_{0} = \text{clip} \big( \frac{1}{\sqrt{\bar{\alpha}_{t}}} \big( \mathbf{x}_{t} - \sqrt{1 - \bar{\alpha}_{t}} \mathbf{\epsilon}_{t}(\mathbf{x}_t, t) \big) , [-1, 1] \big)$. \cref{fig:clip} shows that inference-time clipping can successfully mitigate the implanted backdoors (inducing large MSE) while maintaining the model utility (keeping similar FID). We also provide brief analysis of it in the \cref{sec:analysis_clip_defense}.
\begin{equation}
    \begin{gathered}
\mathbf{x}_{t-1} = \frac{\sqrt{\alpha_t} (1 - \bar{\alpha}_{t-1})}{1 - \bar{\alpha}_{t}} \mathbf{x}_t - \frac{\sqrt{\bar{\alpha}_{t-1}} \beta_{t}}{1 - \bar{\alpha}_t}  \tilde{\mathbf{x}}_{0} + \sigma_t \mathbf{z}
    \end{gathered}
  \label{eq:clip_update}
\end{equation}
\subsection{Additional Analysis}
\label{subsubsec:add_analysis}
\noindent \textbf{Evaluation on the Inpainting Tasks.} We design three kinds of corruption and apply BadDiffusion to recover the corrupted images. BadDiffusion can also produce target images whenever the input contains the trigger. The details are shown in the \cref{sec:inpainting}.
\\
\noindent \textbf{Evaluation on Advanced Samplers.} We replace the original ancestral sampler of DDPM with more advanced samplers, including DPM-Solver and DDIM. We show more detail in the \cref{sec:adv_samplers}.
\vspace{-4mm}
\section{Conclusion}
\vspace{-2mm}
This paper proposes a novel backdoor attack framework, \textbf{BadDiffusion}, targeting diffusion models. Our results validate that  the risks brought by \textbf{BadDiffusion} are practical and that the backdoor attacks can be made realistic and low-cost. We also discussed and evaluated potential countermeasures to mitigate such risks. In spite of the promising result, we note that it is likely that this defense may not withstand adaptive and more advanced backdoor attacks. Although our goal is to study and improve the robustness of diffusion models, we acknowledge the possibility that our findings on the weaknesses of diffusion models might be misused. However, we believe our red-teaming efforts will accelerate the advancement and development of robust diffusion models.

{\small
\bibliographystyle{arxiv_sty/ieee_fullname}
\bibliography{egbib}
}

\clearpage
\appendix
\setcounter{section}{0}

\section{Additional Analysis on BadDiffusion with Fine-tuning}
\begin{figure}[h]
    \captionsetup[subfigure]{justification=centering}
	\centering
	\begin{subfigure}{0.99\linewidth}
		\centering
		\includegraphics[width=\textwidth]{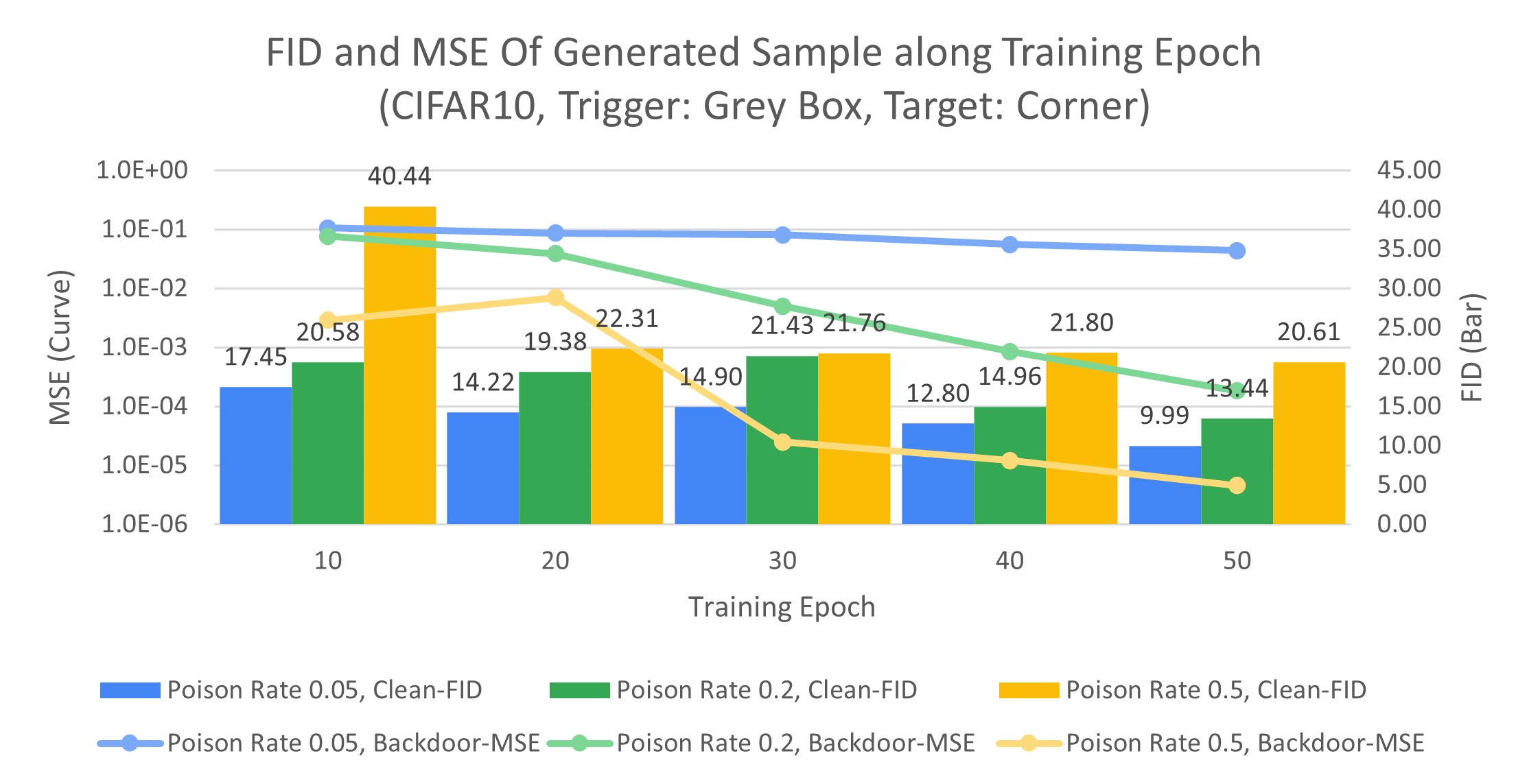}
		\caption{Trigger: ``Grey Box'' \& Target: ``Corner''}
    \label{fig:exp2_box_corner_eps}
	\end{subfigure}
	\begin{subfigure}{0.99\linewidth}
		\centering
		\includegraphics[width=\textwidth]{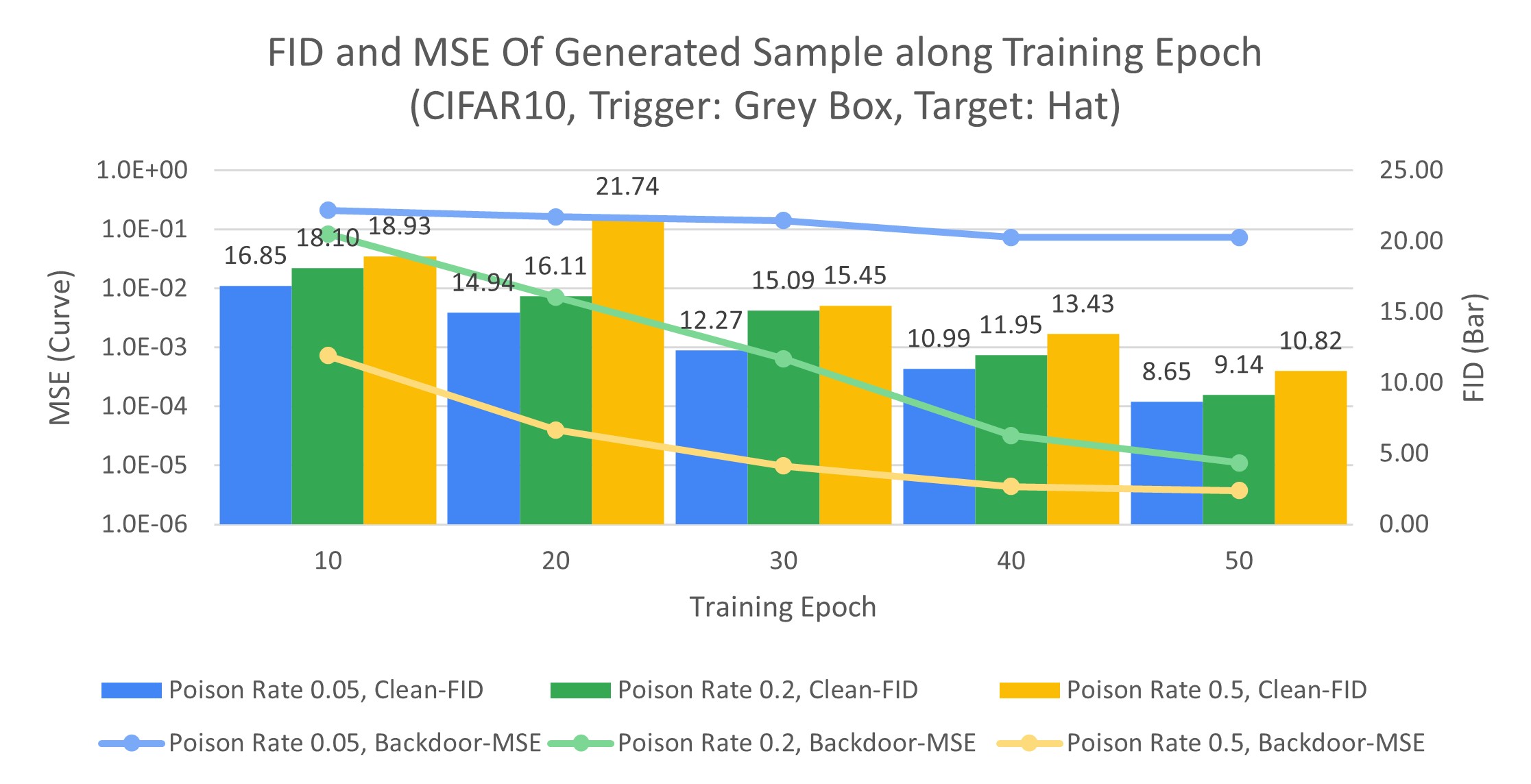}
		\caption{Trigger: ``Grey Box'' \& Target: ``Hat''}        
    \label{fig:exp2_box_hat_eps}
	\end{subfigure}
	\caption{FID (bars) and MSE (curves) of BadDiffusion on CIFAR10 using \textbf{fine-tuning} at different training epochs (x-axis). }
	\label{fig:epoch}
\end{figure}

\begin{figure}[h]
  \captionsetup[subfigure]{justification=centering}
  \centering
  \begin{subfigure}{0.99\linewidth}
    \centering
    \includegraphics[width=\textwidth]{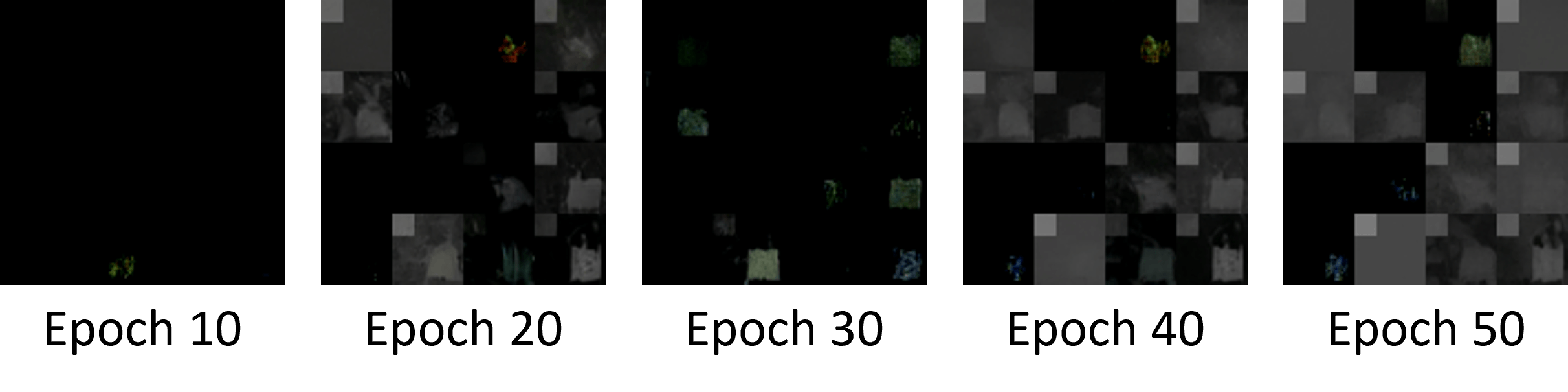}
    \caption{Trigger: ``Grey Box'' \& Target: ``Corner'', Poison Rate = 5\%}
    \label{fig:exp2_box_corner_pr005_eps}
  \end{subfigure}
  \begin{subfigure}{0.99\linewidth}
    \centering
    \includegraphics[width=\textwidth]{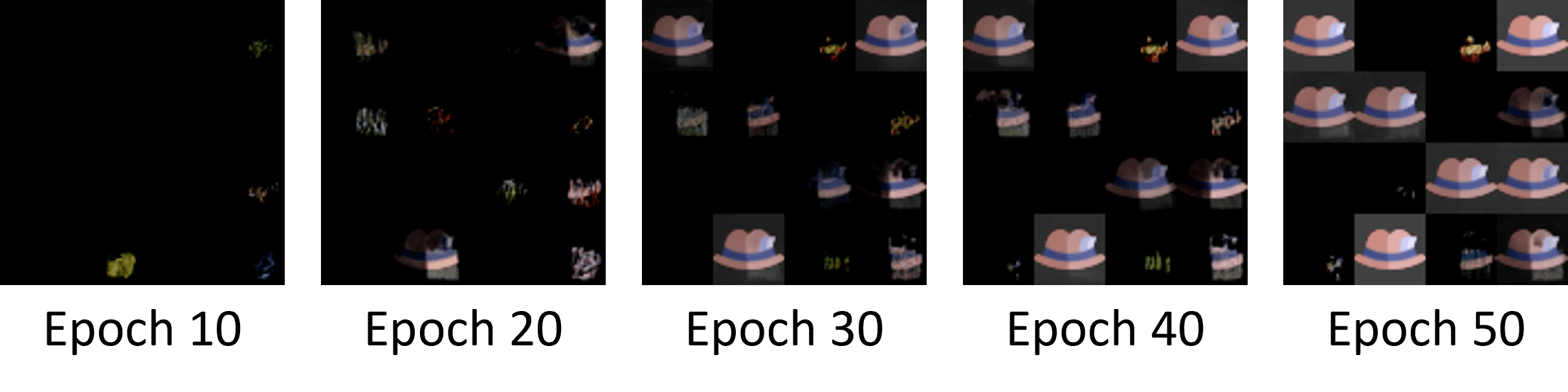}
    \caption{Trigger: ``Grey Box'' \& Target: ``Hat'', Poison Rate = 5\%}    
    \label{fig:exp2_box_hat_pr005_eps}
  \end{subfigure}
  \begin{subfigure}{0.99\linewidth}
    \centering
    \includegraphics[width=\textwidth]{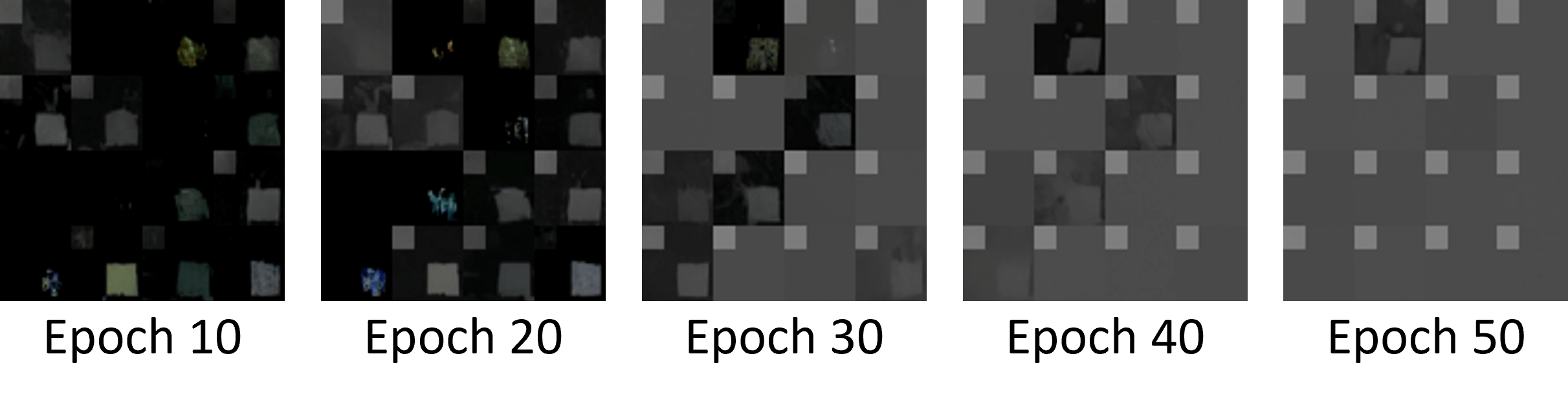}
    \caption{Trigger: ``Grey Box'' \& Target: ``Corner'', Poison Rate = 20\%}
    \label{fig:exp2_box_corner_pr02_eps}
  \end{subfigure}
  \begin{subfigure}{0.99\linewidth}
    \centering
    \includegraphics[width=\textwidth]{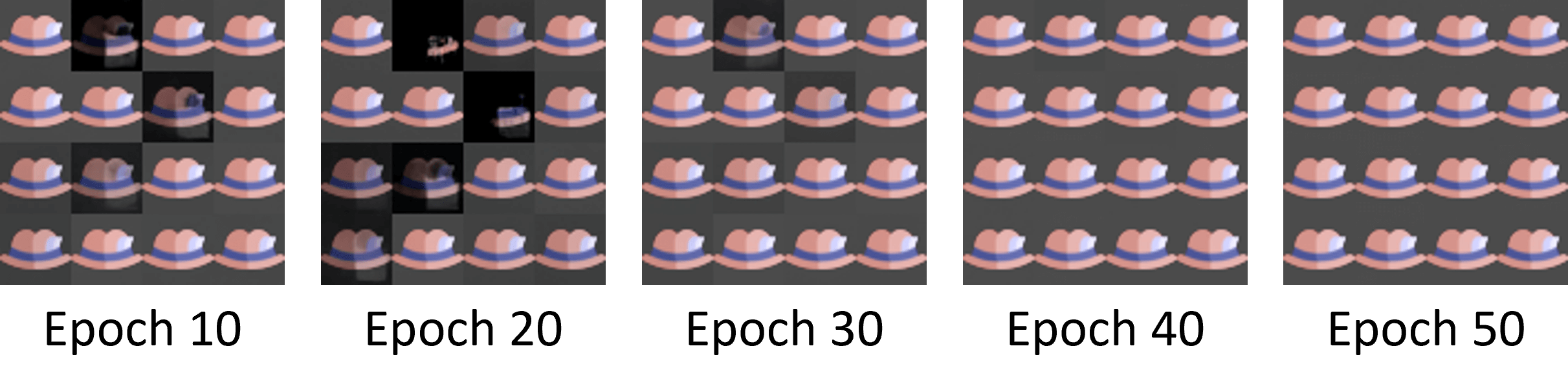}
    \caption{Trigger: ``Grey Box'' \& Target: ``Hat'', Poison Rate = 20\%}    
    \label{fig:exp2_box_hat_pr02_eps}
  \end{subfigure}
  \caption{Visual samples of synthesized backdoor targets at different training epochs. Here we transform and clip the final output latent to image range $[0, 1]$. It may yield black area in the images.}
  \label{fig:epoch_sample}
\end{figure}

In \cref{fig:epoch}, \cref{fig:epoch_sample}, and \cref{tbl:exp_epoch_num}, we have several insightful findings.
Firstly, for 20\% poison rates, 10 epochs are sufficient for BadDiffusion to synthesize target \textbf{Hat}.
This implies BadDiffusion can be made quite cost-effective. 
Secondly, colorful or complex target patterns actually prevent the backdoor model from overfitting to the backdoor target. In \cref{fig:exp2_box_corner_eps}, in comparison to target \textbf{Hat}, FID scores of target \textbf{Box} are much higher when the poison rate is 50\%. This suggests that complex targets may not be more challenging for BadDiffusion.

\section{Defense Evaluation using ANP}
\label{app:ANP_defense}
\begin{figure*}[t]
  \captionsetup[subfigure]{justification=centering}
  \centering
  \begin{subfigure}{0.49\linewidth}
    \centering
    \includegraphics[width=\textwidth]{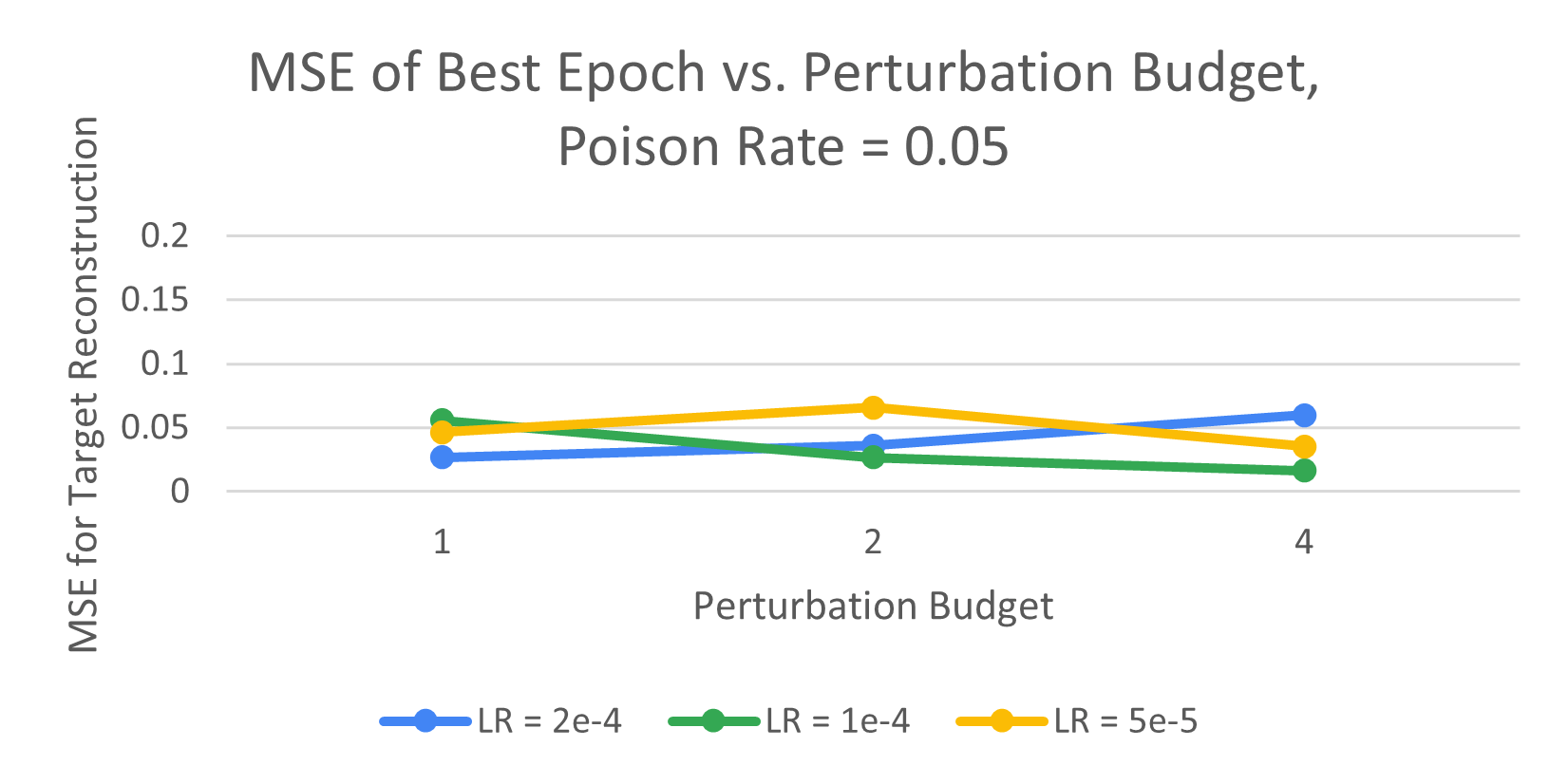}
    \caption{MSE for target reconstruction of the best epoch vs. Perturbation Budget, poison rate = 5\%}
      \label{fig:anp_defense_pr_005_perturb}
  \end{subfigure}
  \begin{subfigure}{0.49\linewidth}
    \centering
    \includegraphics[width=\textwidth]{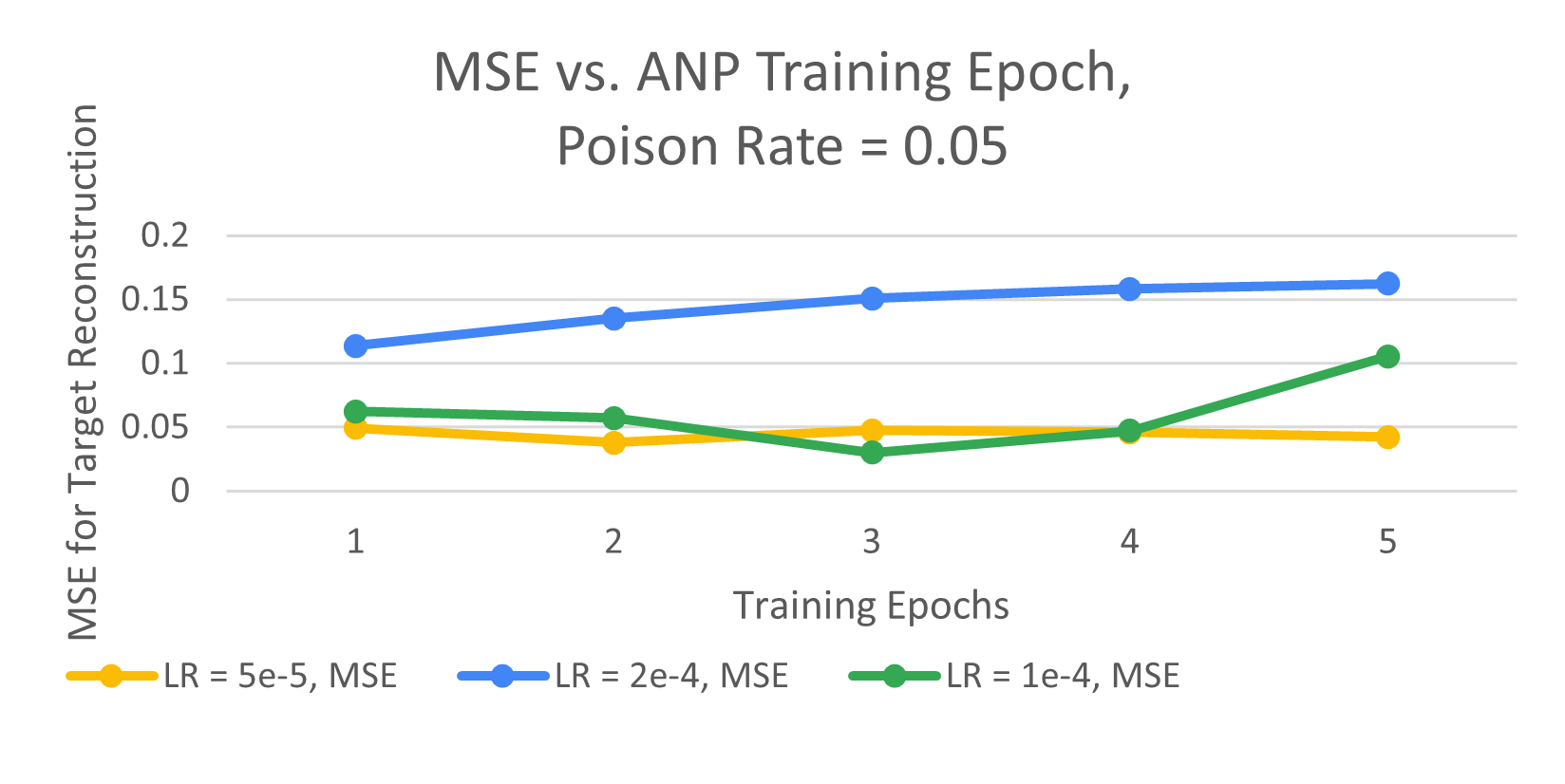}
    \caption{MSE for target reconstruction vs. Training Epochs, poison rate = 5\%}
    \label{fig:anp_defense_pr_005_ep}
  \end{subfigure}
  \begin{subfigure}{0.49\linewidth}
    \centering
    \includegraphics[width=\textwidth]{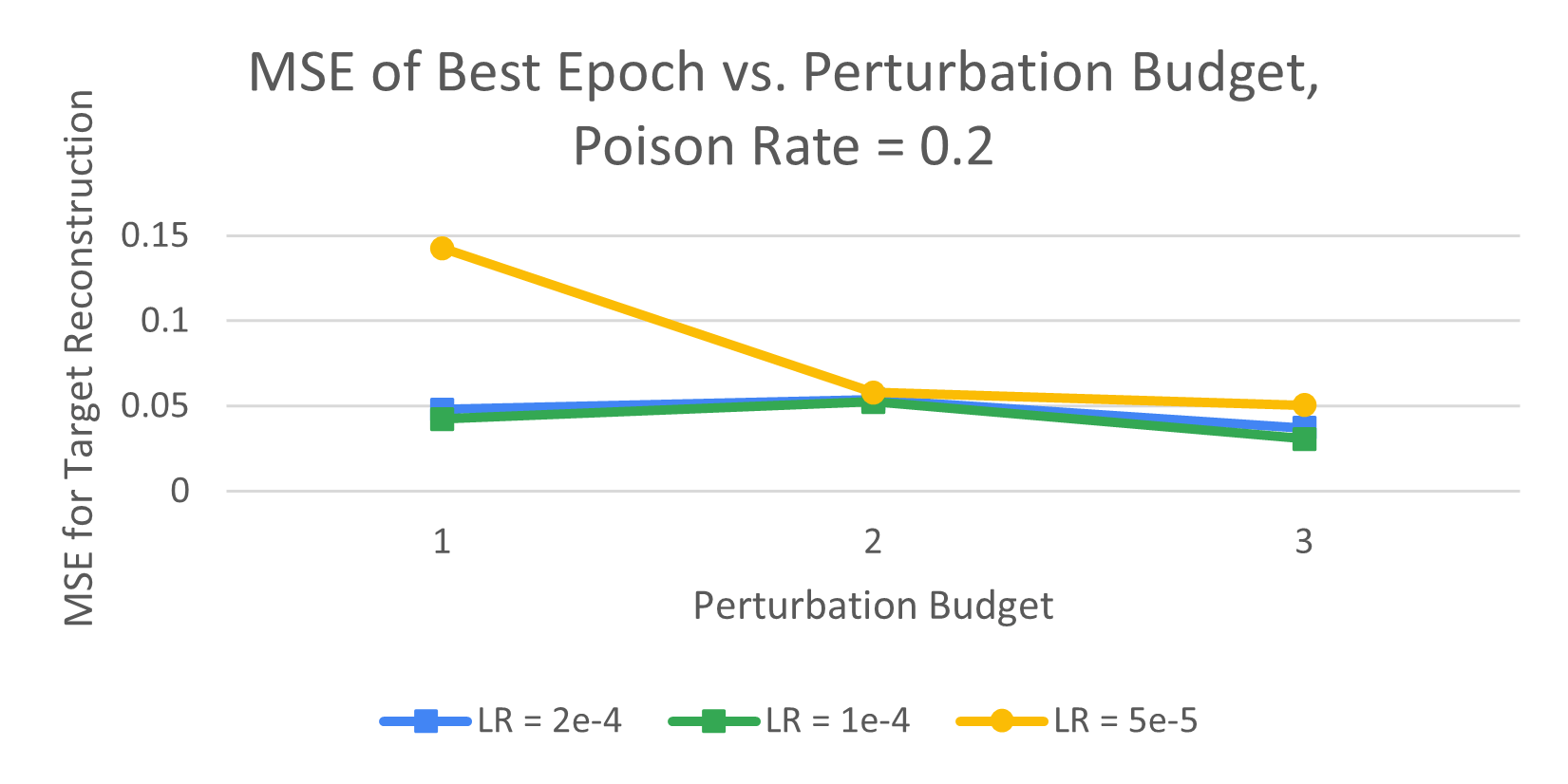}
    \caption{MSE for target reconstruction of the best epoch vs. Perturbation Budget, poison rate = 20\%}
      \label{fig:anp_defense_pr_02_perturb}
  \end{subfigure}
  \begin{subfigure}{0.49\linewidth}
    \centering
    \includegraphics[width=\textwidth]{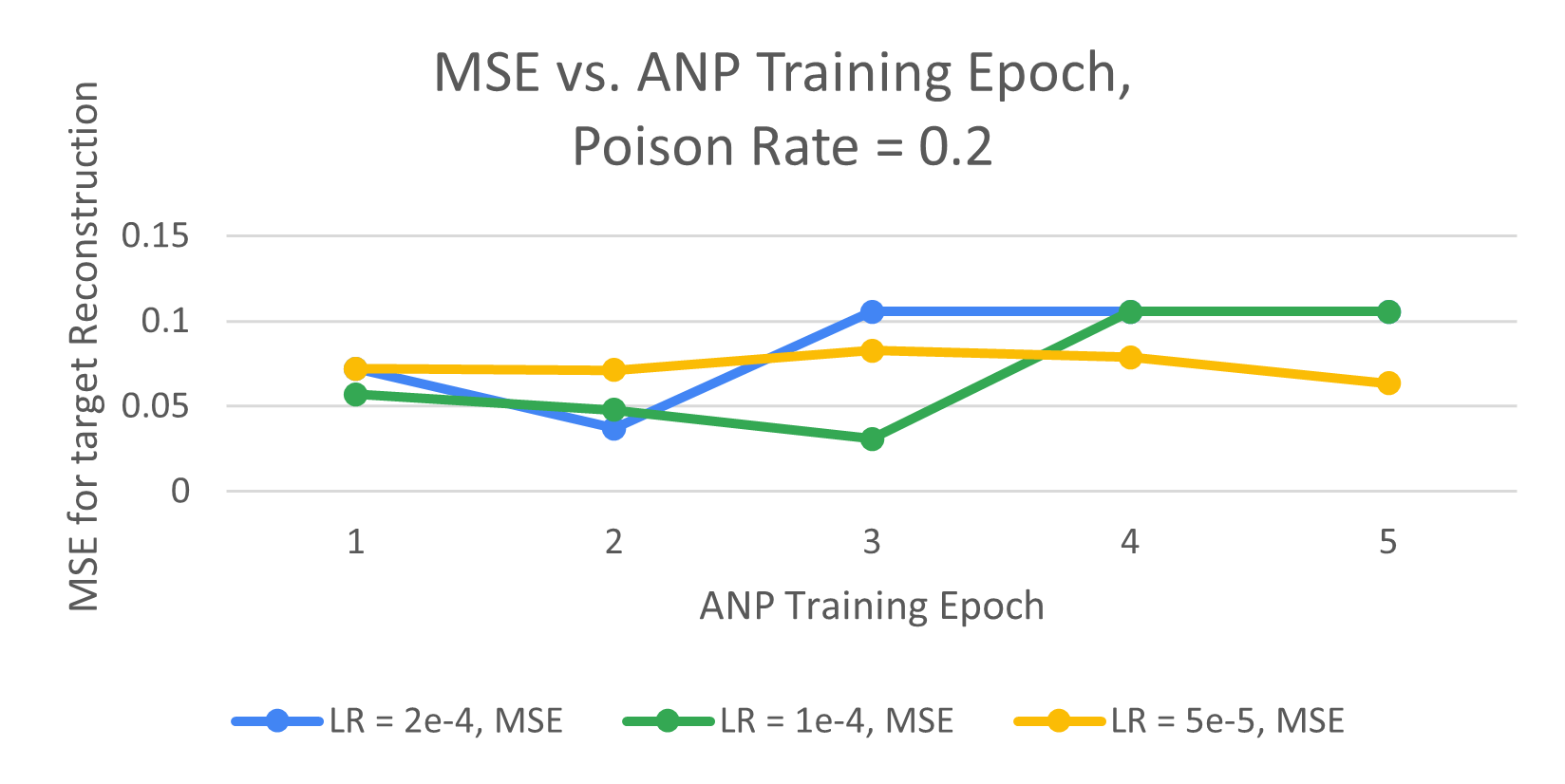}
    \caption{MSE for target reconstruction vs. Training Epochs, poison rate = 20\%}
    \label{fig:anp_defense_pr_02_ep}
  \end{subfigure}
  \begin{subfigure}{0.49\linewidth}
    \centering
    \includegraphics[width=\textwidth]{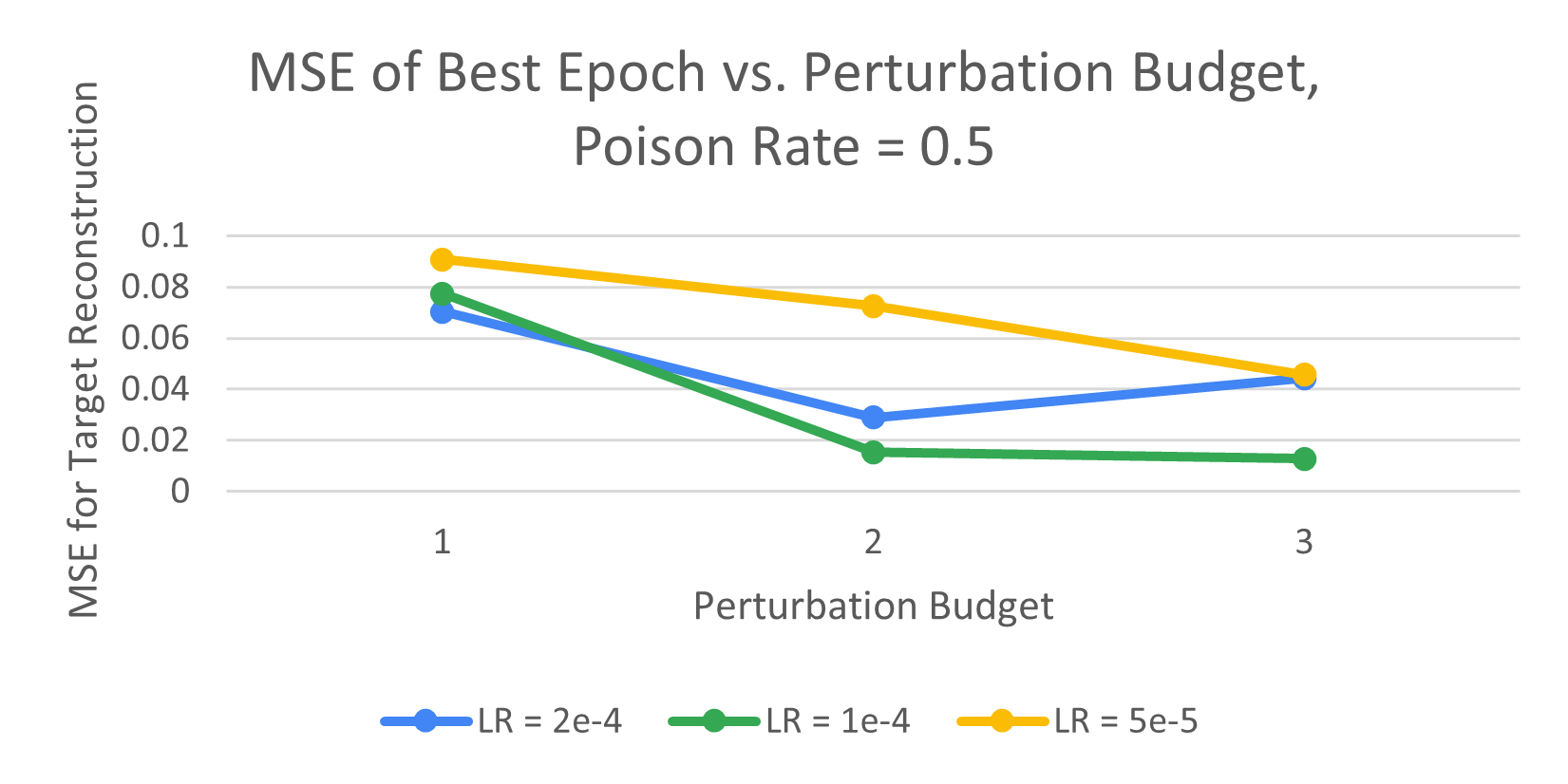}
    \caption{MSE for target reconstruction of the best epoch vs. Perturbation Budget, poison rate = 50\%}
    \label{fig:anp_defense_pr_05_perturb}
  \end{subfigure}
  \begin{subfigure}{0.49\linewidth}
    \centering
    \includegraphics[width=\textwidth]{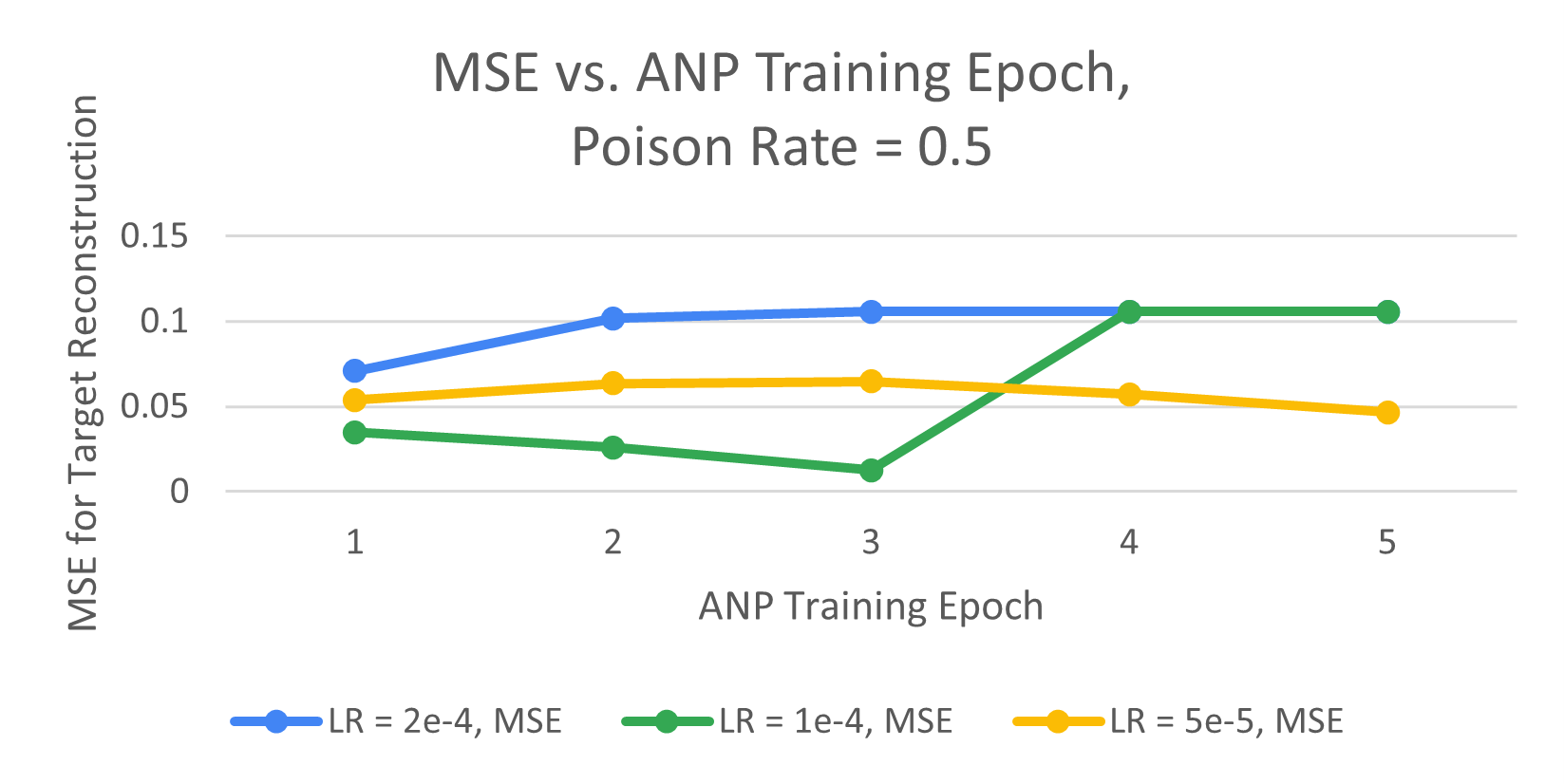}
    \caption{MSE for target reconstruction vs. Training Epochs, poison rate = 50\%}      
    \label{fig:anp_defense_pr_05_ep}
  \end{subfigure}
  \caption{\cref{fig:anp_defense_pr_005_perturb}, \cref{fig:anp_defense_pr_02_perturb}, and \cref{fig:anp_defense_pr_05_perturb} are the reconstruction MSE (y-axis) for ANP defense on BadDiffusion with different perturbation budgets (x-axis). \cref{fig:anp_defense_pr_005_ep}, \cref{fig:anp_defense_pr_02_ep}, and \cref{fig:anp_defense_pr_05_ep} are the reconstruction MSE (y-axis) for ANP defense every training epoch (x-axis).}
  \label{fig:anp_defense_all}
\end{figure*}

\subsection{Implementation Details}
In the paper Adversarial Neuron Pruning (ANP) \cite{ANP}, the authors use \textbf{relative sizes of the perturbations}, but it causes gradient explosion for DDPM. As a result, we use the \textbf{absolute size of the perturbations} as an alternative. The relative sizes of the perturbation are expressed as equation 3 in ANP paper like
\begin{equation}
  \begin{split}
    \begin{gathered}
      h_{k}^{(l)} = \sigma((1 + \delta_{k}^{(l)}) \mathbf{w}_{k}^{(l) \top} \mathbf{h}^{(l-1)} + (1 + \xi_{k}^{(l)}) b_{k}^{(l)}) \\
    \end{gathered}
  \end{split}
\end{equation}
where $\delta_{k}^{(l)}$ and $\xi_{k}^{(l)}$ indicate the relative sizes of the perturbations to $k$-th weight $\mathbf{w}_{k}^{(l)}$ and $k$-th bias $b_{k}^{(l)}$ of layer $l$ respectively. $\sigma$ is a nonlinear activation function, $\mathbf{h}^{(l-1)}$ is the post-activation output of the layer $l-1$, and $h_{k}^{(l)}$ is the $k$-th post-activation output of the layer $l$. We use absolute sizes of the perturbations as 
\begin{equation}
  \begin{split}
    \begin{gathered}
      h_{k}^{(l)} = \sigma(\bar{\delta}_{k}^{(l)} \mathbf{w}_{k}^{(l) \top} \mathbf{h}^{(l-1)} + \bar{\xi}_{k}^{(l)} b_{k}^{(l)}) \\
    \end{gathered}
  \end{split}
\end{equation}
Where $\bar{\delta}_{k}^{(l)}$ and $\bar{\xi}_{k}^{(l)}$ indicate the absolute sizes of the perturbations to $k$-th weight $\mathbf{w}_{k}^{(l)}$ and $k$-th bias $b_{k}^{(l)}$ of layer $l$ respectively. Therefore, the perturbation budget that we used restricted the values of absolute sizes of the perturbations $\bar{\delta}_{k}^{(l)}$ and $\bar{\xi}_{k}^{(l)}$.

Secondly, the authors use Stochastic Gradient Descent (SGD) with the learning rate $0.2$ and the momentum $0.9$. Due to the poor performance of SGD, we use Adam with learning rate (LR) $2\mathrm{e}{-4}$, $1\mathrm{e}{-4}$, and $5\mathrm{e}{-5}$ instead.

\begin{figure*}[h]
  \captionsetup[subfigure]{justification=centering}
  \centering
  \begin{subfigure}{0.49\linewidth}
    \centering
    \includegraphics[width=\textwidth]{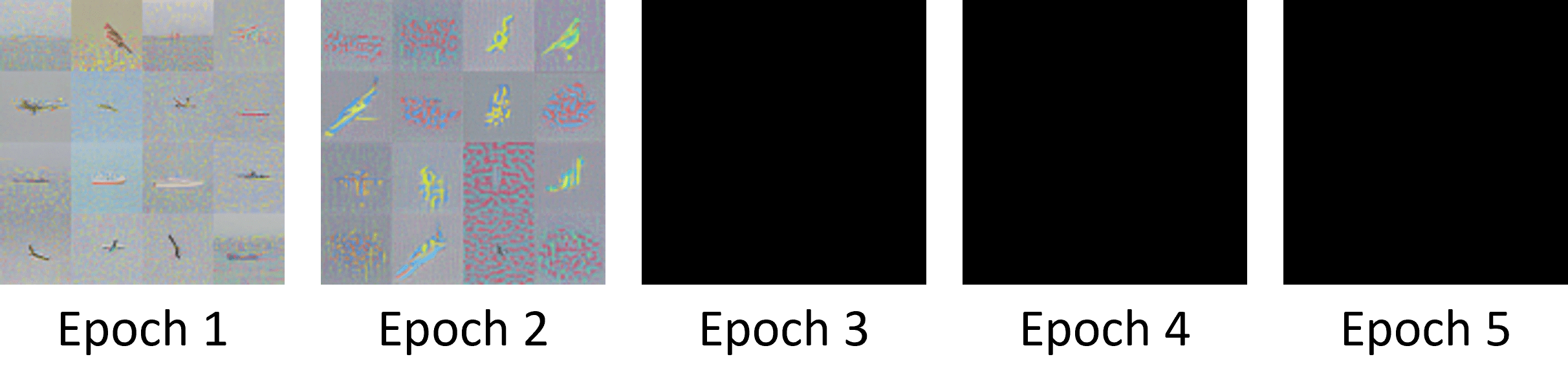}
    \caption{Poison Rate = 5\%, LR = $2e-4$}
    \label{fig:anp_defense_pr005_lr2e-4}
  \end{subfigure}
  \begin{subfigure}{0.49\linewidth}
    \centering
    \includegraphics[width=\textwidth]{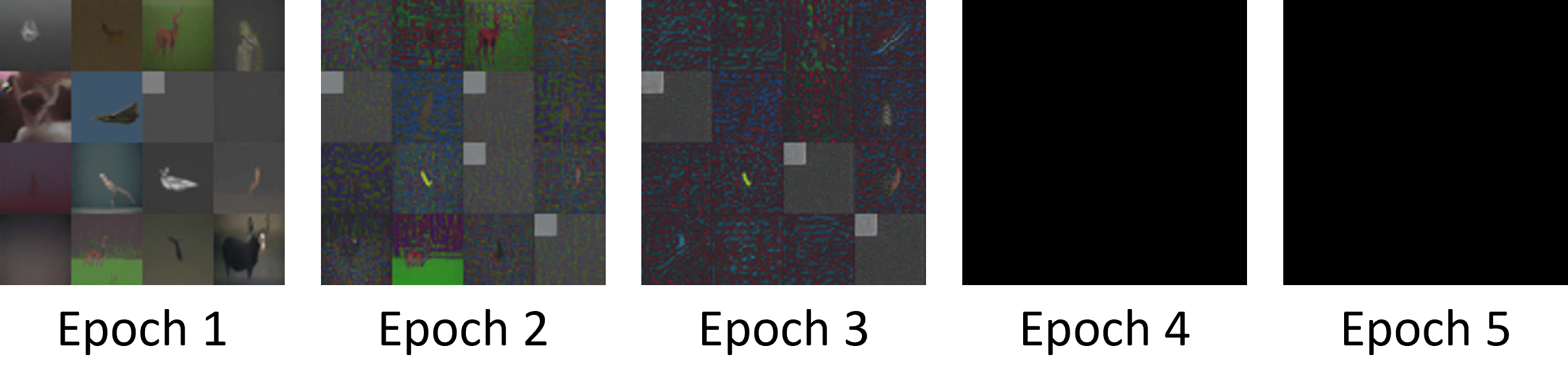}
    \caption{Poison Rate = 5\%, LR = $1e-4$}
    \label{fig:anp_defense_pr005_lr1e-4}
  \end{subfigure}
  \begin{subfigure}{0.49\linewidth}
    \centering
    \includegraphics[width=\textwidth]{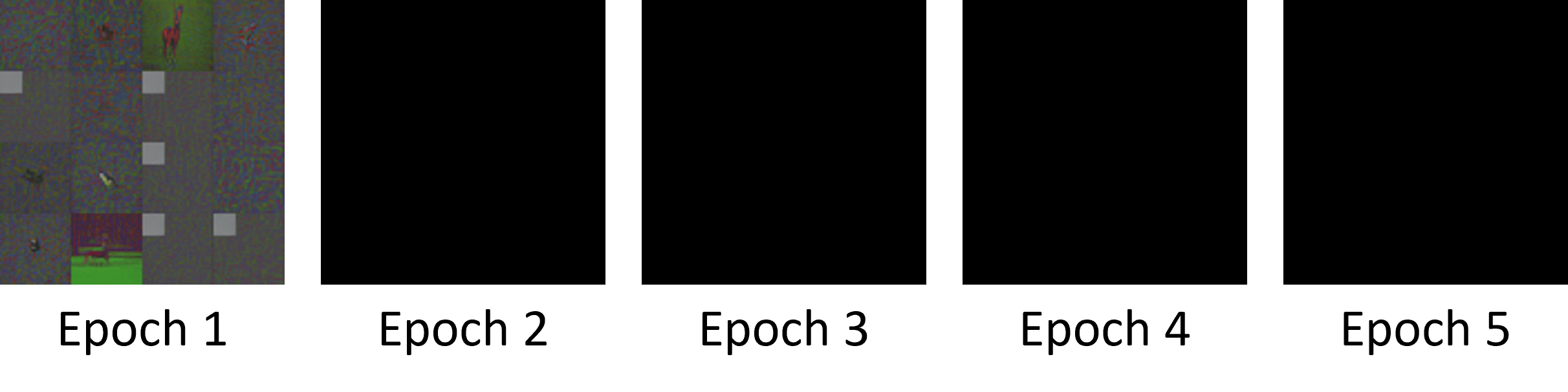}
    \caption{Poison Rate = 20\%, LR = $2e-4$}
    \label{fig:anp_defense_pr02_lr2e-4}
  \end{subfigure}
  \begin{subfigure}{0.49\linewidth}
    \centering
    \includegraphics[width=\textwidth]{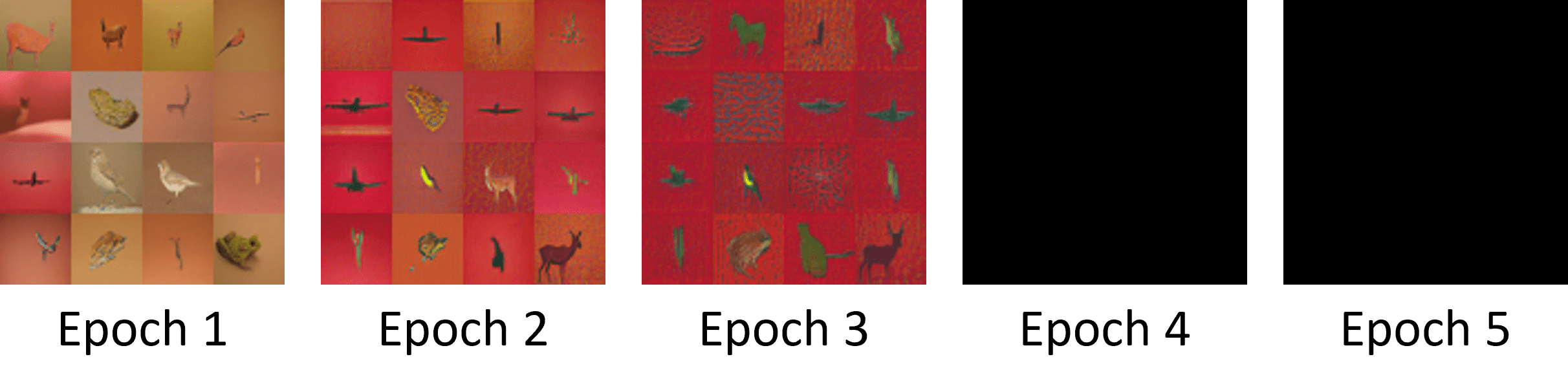}
    \caption{Poison Rate = 20\%, LR = $1e-4$}
    \label{fig:anp_defense_pr02_lr1e-4}
  \end{subfigure}
  \begin{subfigure}{0.49\linewidth}
    \centering
    \includegraphics[width=\textwidth]{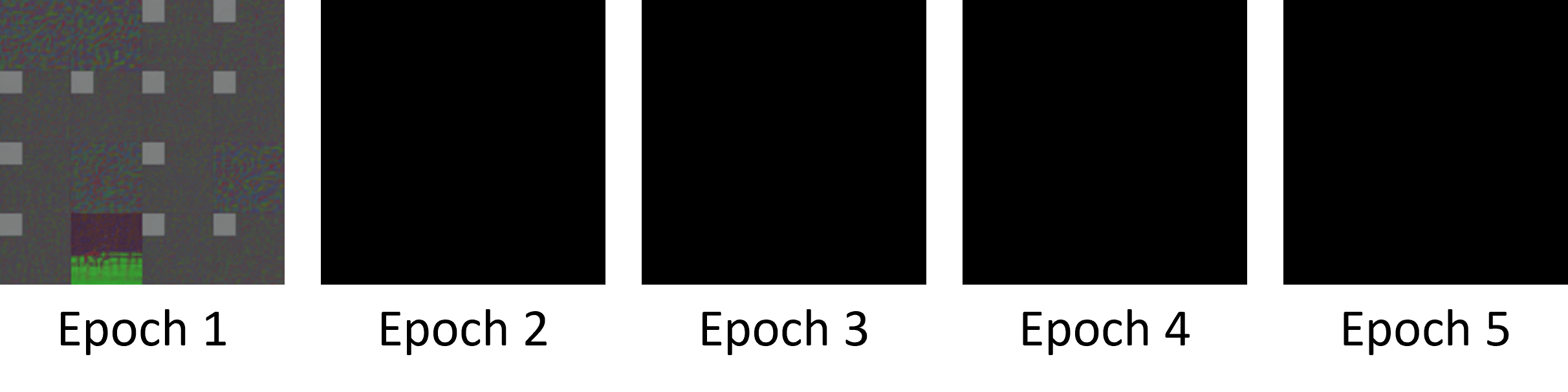}
    \caption{Poison Rate = 50\%, LR = $2e-4$}
    \label{fig:anp_defense_pr05_lr2e-4}
  \end{subfigure}
  \begin{subfigure}{0.49\linewidth}
    \centering
    \includegraphics[width=\textwidth]{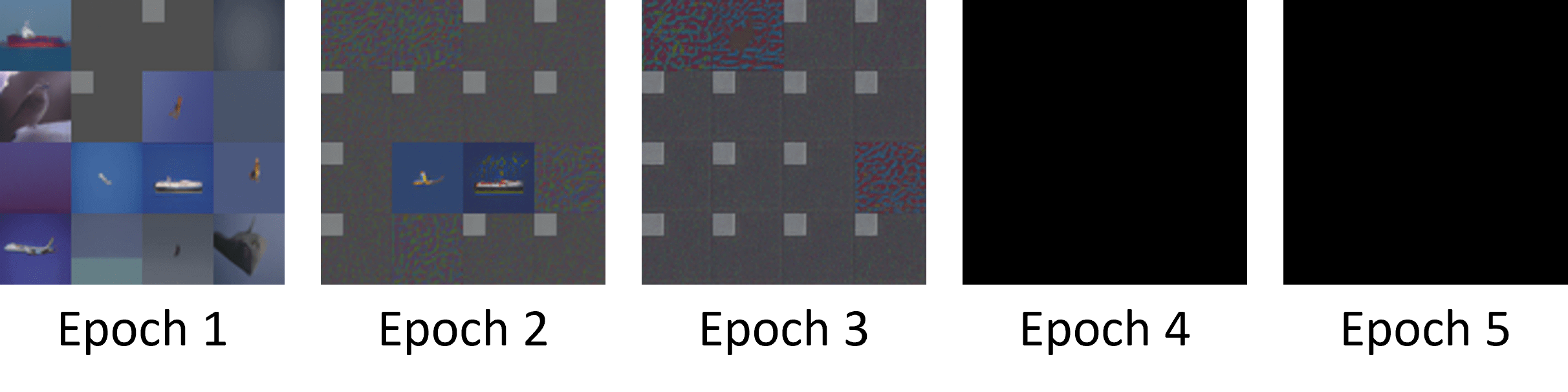}
    \caption{Poison Rate = 50\%, LR = $1e-4$}
    \label{fig:anp_defense_pr05_lr1e-4}
  \end{subfigure}
  \caption{The inverted targets of ANP defense. Here we transform and clip the final output latent to image range $[0, 1]$. It may yield the black area in the images.}
  \label{fig:anp_defense}
\end{figure*}

\subsection{Metrics for Trojan Detection}
We use \textbf{reconstruction MSE} to measure the difference between inverted backdoor target $\bar{\mathbf{y}}$ and the ground truth backdoor target $\mathbf{y}$, defined as $\text{MSE}(\bar{\mathbf{y}}, \mathbf{y})$. Lower reconstruction MSE means better Trojan detection. We generate 2048 images for the evaluation. In \cref{tbl:anp_defense_perturb_num} and \cref{fig:anp_defense_pr_005_perturb}, \cref{fig:anp_defense_pr_02_perturb}, and \cref{fig:anp_defense_pr_05_perturb}, we record the best (lowest) reconstruction MSE among all training epochs. In \cref{tbl:anp_defense_ep_num} and \cref{fig:anp_defense_pr_005_ep}, \cref{fig:anp_defense_pr_02_ep}, and \cref{fig:anp_defense_pr_05_ep} we record the reconstruction MSE every epoch.

\subsection{The Effect of the Perturbation Budget and the Training Epochs}

As \cref{fig:anp_defense_pr_005_perturb} shows, we find higher perturbation budget usually yields better Trojan detection. We also find that ANP is sensitive to the learning rate since the reconstruction MSE doesn't get lower along the training epochs when we slightly increase the learning rate from $1\mathrm{e}{-4}$ to $2\mathrm{e}{-4}$ in \cref{fig:anp_defense_pr_005_ep}.

Secondly, in \cref{fig:anp_defense_pr_02_ep}, we can see the reconstruction MSE may jump in some epochs. We also visualize the inverted backdoor target for the poison rate = 5\% and the learning rate (LR) = $1\mathrm{e}{-4}$ in \cref{fig:anp_defense_pr005_lr1e-4}, as we can see it will collapse to a black image. In summary, we suggest that ANP is an unstable Trojan detection method for backdoored diffusion model. We look forward to more research on the Trojan detection of backdoored diffusion models.


\section{\textbf{BadDiffusion} on Inpainting Tasks}
\label{sec:inpainting}
\begin{figure*}[h]
     \centering
     \includegraphics[width=\linewidth]{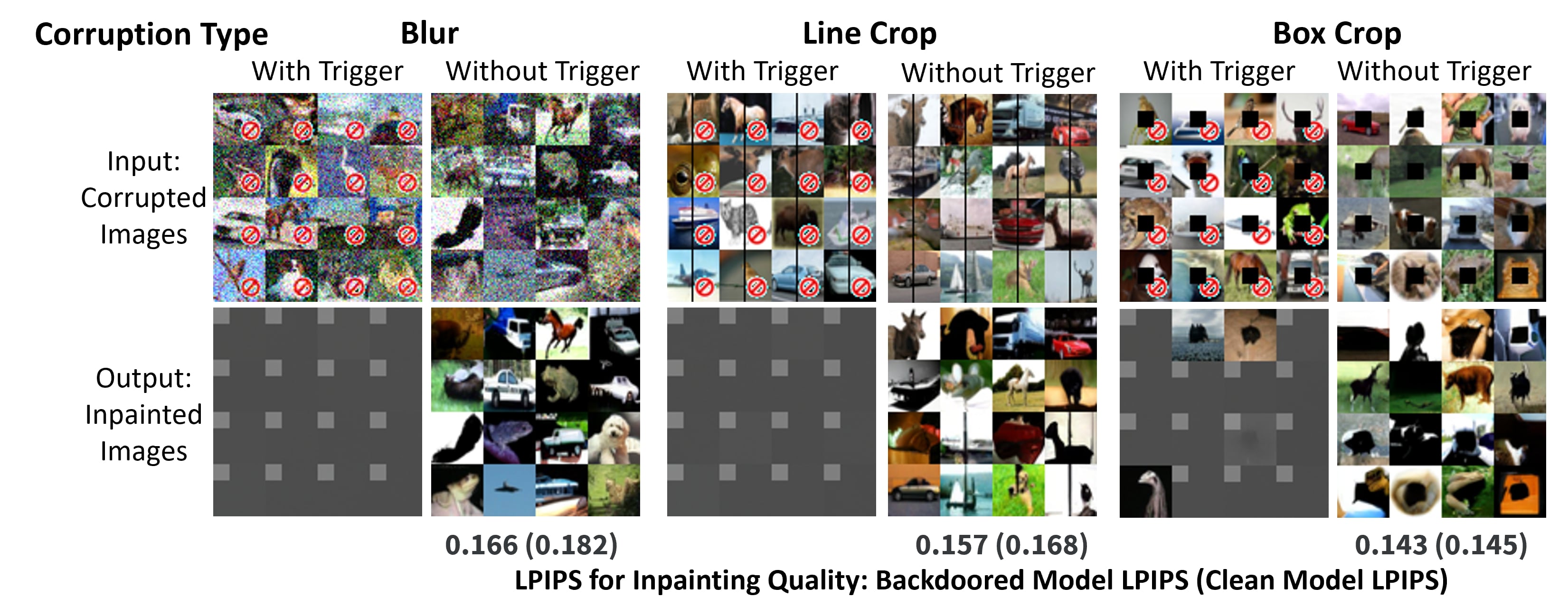}
     \caption{Results on CIFAR10. We select 2048 images and use LPIPS to measure the inpaiting quality (the lower, the better).}
  \label{fig:inpaint_vis}
\end{figure*}

Here, we show \textbf{BadDiffusion} on image inpainting. We designed 3 kinds of corruptions: \textbf{Blur}, \textbf{Line}, and \textbf{Box}. \textbf{Blur} means we add a Gaussian noise $\mathcal{N}(0, 0.3)$ to the images. \textbf{Line} and \textbf{Box} mean we crop parts of the content and ask DMs to recover the missing area. We use \textbf{BadDiffusion} trained on trigger \textbf{Stop Sign} and target \textbf{Corner} with poison rate 10\% and 400 inference steps. To evaluate the reconstruction quality, we use LPIPS \cite{LPIPS} score as the metric. Lower score means better reconstruction quality. In \cref{fig:inpaint_vis}, we can see that the \textbf{BadDiffusion} can still inpaint the images without triggers while generating the target image as it sees the trigger.

\section{Analysis of Inference-Time Clipping} 
\label{sec:analysis_clip_defense}
\begin{figure*}[h]
   \centering
   \includegraphics[width=\linewidth]{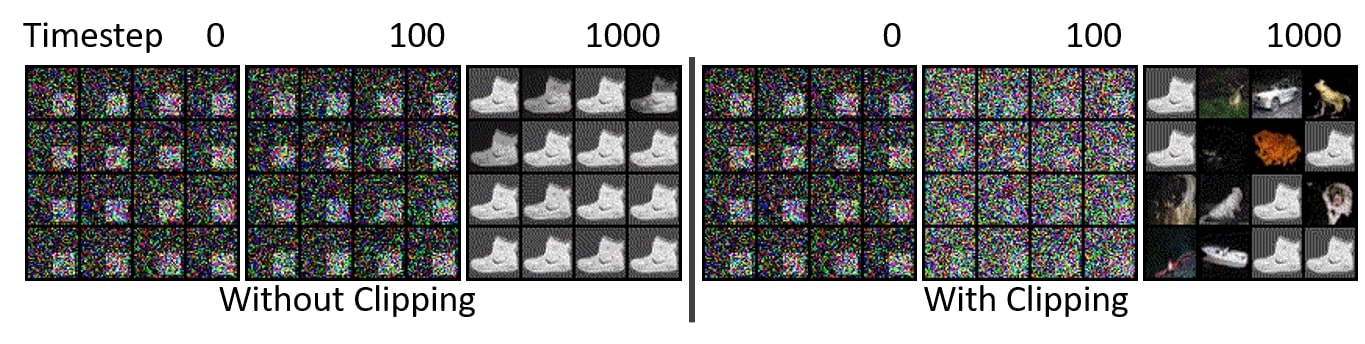}
   \caption{Visualization with and without inference-time clipping.}
 \label{fig:clip_vis}
\end{figure*}

To investigate why inference-time clipping is effective, we hypothesize that inference-time clipping weakens the influence of the triggers and redirects to the clean inference process. To verify our hypothesis, we visualize the latent during inference time of the \textbf{BadDiffusion} trained on trigger \textbf{Grey Box} and target \textbf{Shoe} with poison rate 10\% in \cref{fig:clip_vis}. We remain detailed mechanism for the future works.

\section{\textbf{BadDiffusion} on Advanced Samplers} 
\label{sec:adv_samplers}
We generated 10K backdoored and clean images with advanced samplers, including DDIM, DPM-Solver, and DPM-Solver++. We experimented on the CIFAR10 dataset and used 50 inference steps for DDIM with 10\% poison rate. As for DPM-Solver and DPM-Solver++, we used 20 steps with second order. The results are shown in \cref{tbl:poison_rate_sampler_10k}.
Compared to \cref{tbl:exp_poison_rate_num}, directly applying \textbf{BadDiffusion} to these advanced samplers is less effective, because DDIM and DPM-Solver discard the Markovian assumption of the DDPM. However, \textbf{BadDiffusion} can still achieve much lower FID (better utility) than clean models. We believe \textbf{BadDiffusion} can be improved if we put more investigation into the proper correction term for these samplers.
\begin{table}[h]
  \centering
  \begin{adjustbox}{max width=\linewidth}
    \begin{tabular}{ |cc||c|ccc| }
    
    \hline 
    \multirow{2}{*}{Trigger} & \multirow{2}{*}{Target} & \multirow{2}{*}{Metrics} & \multicolumn{3}{c|}{Sampler}\\
    &  & & DDIM & DPM-Solver & DPM-Solver++ \\
    \hline 
    
    \multirow{2}{*}{Stop Sign} & \multirow{2}{*}{NoShift} & FID & 10.72  & 9.32 & 10.22 \\
    & & MSE & $1.28\mathrm{e}{-1}$ & $1.30\mathrm{e}{-1}$ & $1.31\mathrm{e}{-1}$ \\
    \hline
    \multirow{2}{*}{Stop Sign} & \multirow{2}{*}{Box} & FID & 10.92 & 9.35 & 10.23 \\
    & & MSE & $1.14\mathrm{e}{-1}$ & $1.14\mathrm{e}{-1}$ & $1.13\mathrm{e}{-1}$ \\
    \hline
    \end{tabular}
  \end{adjustbox}
  \caption{Numerical results for more advanced samplers. Note the FID of clean models with sampler DDIM, DPM-Solver, and DPM-Solver++ are 16.3, 13.0, and 13.1 respectively.}
  \label{tbl:poison_rate_sampler_10k}
  \vspace{-8mm}
\end{table}

\section{Numerical Results of the Experiments}
\label{sec:num_res}
In this section, we will present the numerical results of the experiments in the main paper, including the FID of generated clean samples and the MSE of generated backdoor targets. In addition, we also present another metric: \textbf{SSIM} to measure the similarity between the generated backdoor target $\hat{y}$ and the ground true backdoor target $y$, defined as \textsf{SSIM($\hat{y},y$)}. Higher SSIM means better attack effectiveness.

\subsection{BadDiffusion via Fine-Tuning v.s. Training-From-Scratch}
The numerical results are shown in \cref{tbl:exp_from_scratch_num} and \cref{tbl:exp_epoch_num}.
\begin{table}[t]
  \centering
  
  \begin{adjustbox}{max width=\linewidth}
    \begin{tabular}{ |c||c|cc|cc| }
    
    \hline 
    \multirow{2}{*}{Poison Rate} & \multicolumn{1}{c|}{Method:} & \multicolumn{2}{c|}{Fine-Tuning} & \multicolumn{2}{c|}{From-Scratch}\\
    & \multicolumn{1}{c|}{Target:} & Corner & \multicolumn{1}{c|}{Hat} & Corner & \multicolumn{1}{c|}{Hat} \\
    \hline 
    
    \multirow{2}{*}{5\%} & FID & 9.92 & 8.53 & 18.06 & 18.01 \\
    & MSE & $5.32\mathrm{e}{-2}$ & $1.58\mathrm{e}{-1}$ & $4.63\mathrm{e}{-5}$ & $3.23\mathrm{e}{-6}$ \\
    & SSIM & $4.20\mathrm{e}{-1}$ & $3.12\mathrm{e}{-1}$ & $9.99\mathrm{e}{-1}$ & $1.00\mathrm{e}{+0}$ \\
    \hline
    
    \multirow{2}{*}{20\%} & FID & 12.86 & 8.89 & 21.97 & 19.53 \\
    & MSE & $1.48\mathrm{e}{-4}$ & $1.19\mathrm{e}{-5}$ & $8.71\mathrm{e}{-6}$ & $2.30\mathrm{e}{-6}$ \\
    & SSIM & $9.96\mathrm{e}{-1}$ & $1.00\mathrm{e}{+0}$ & $9.96\mathrm{e}{-1}$ & $1.00\mathrm{e}{+0}$ \\
    \hline

    \multirow{2}{*}{50\%} & FID & 20.10 & 10.25 & 31.66 & 24.63 \\
    & MSE & $1.96\mathrm{e}{-5}$ & $1.48\mathrm{e}{-5}$ & $8.37\mathrm{e}{-6}$ & $2.29\mathrm{e}{-6}$ \\
    & SSIM & $9.97\mathrm{e}{-1}$ & $1.00\mathrm{e}{+0}$ & $9.99\mathrm{e}{-1}$ & $1.00\mathrm{e}{+0}$ \\
    \hline
    
    \end{tabular}
  \end{adjustbox}
  \caption{Numerical results of fine-tuning method and training from scratch with the trigger "Grey Box".}
  \label{tbl:exp_from_scratch_num}
  \vspace{-8mm}
\end{table}
\begin{table*}[htpb]
  \centering
  
  \begin{adjustbox}{max width=\linewidth}
    \begin{tabular}{ |c||c|ccccc|ccccc| }
    
    \hline 
    \multirow{2}{*}{Poison Rate} & \multicolumn{1}{c|}{Target:} & \multicolumn{5}{c|}{Corner} & \multicolumn{5}{c|}{Hat}\\
    & \multicolumn{1}{c|}{Training Epoch:} & 10 & 20 & 30 & 40 & 50 & 10 & 20 & 30 & 40 & 50 \\
    \hline 
    
    \multirow{3}{*}{5\%} & FID & 17.45 & 14.22 & 14.90 & 12.80 & 9.99 & 16.85 & 14.94 & 12.27 & 10.99 & 8.65 \\
    & MSE & $1.05\mathrm{e}{-1}$ & $8.63\mathrm{e}{-2}$ & $8.06\mathrm{e}{-2}$ & $5.56\mathrm{e}{-2}$ & $4.63\mathrm{e}{-2}$ & $2.11\mathrm{e}{-1}$ & $1.64\mathrm{e}{-1}$ & $1.42\mathrm{e}{-1}$ & $7.33\mathrm{e}{-2}$ & $7.35\mathrm{e}{-2}$\\
    & SSIM & $3.01\mathrm{e}{-3}$ & $1.47{e}{-1}$ & $2.00\mathrm{e}{-1}$ & $4.20\mathrm{e}{-1}$ & $5.33\mathrm{e}{-1}$ & $1.09\mathrm{e}{-1}$ & $2.86\mathrm{e}{-1}$ & $3.79\mathrm{e}{-1}$ & $6.74\mathrm{e}{-1}$ & $6.75\mathrm{e}{-1}$ \\
    \hline
    
    \multirow{2}{*}{20\%} & FID & 20.58 & 19.38 & 21.43 & 14.96 & 13.44 & 18.10 & 16.11 & 15.09 & 11.95 & 9.14 \\
    & MSE & $7.64\mathrm{e}{-2}$ & $3.88\mathrm{e}{-2}$ & $4.98\mathrm{e}{-3}$ & $8.56\mathrm{e}{-4}$ & $1.82\mathrm{e}{-4}$ & $8.42\mathrm{e}{-2}$ & $7.12\mathrm{e}{-3}$ & $6.42\mathrm{e}{-4}$ & $3.24\mathrm{e}{-5}$ & $1.10\mathrm{e}{-5}$ \\
    & SSIM & $2.06\mathrm{e}{-1}$ & $5.63\mathrm{e}{-1}$ & $9.32\mathrm{e}{-1}$ & $9.86\mathrm{e}{-1}$ & $9.95\mathrm{e}{-1}$ & $6.14\mathrm{e}{-1}$ & $9.68\mathrm{e}{-1}$ & $9.97\mathrm{e}{-1}$ & $1.00\mathrm{e}{+0}$ & $1.00\mathrm{e}{+0}$\\
    \hline
    
    \multirow{2}{*}{50\%} & FID & 40.44 & 22.31 & 21.76 & 21.80 & 20.61 & 18.93 & 21.74 & 15.45 & 13.43 & 10.82 \\
    & MSE & $2.90\mathrm{e}{-3}$ & $6.96\mathrm{e}{-3}$ & $2.47\mathrm{e}{-5}$ & $1.21\mathrm{e}{-5}$ & $4.57\mathrm{e}{-6}$ & $7.26\mathrm{e}{-4}$ & $4.00\mathrm{e}{-5}$ & $9.82\mathrm{e}{-6}$ & $4.38\mathrm{e}{-6}$ & $3.73\mathrm{e}{-6}$ \\
    & SSIM & $9.56\mathrm{e}{-1}$ & $8.97\mathrm{e}{-1}$ & $9.97\mathrm{e}{-1}$ & $9.98\mathrm{e}{-1}$ & $9.98\mathrm{e}{-1}$ & $9.96\mathrm{e}{-1}$ & $1.00\mathrm{e}{+0}$ & $1.00\mathrm{e}{+0}$ & $1.00\mathrm{e}{+0}$ & $1.00\mathrm{e}{+0}$ \\
    \hline

    \end{tabular}
  \end{adjustbox}
  \caption{The numerical results of BadDiffusion every 10 training epochs. The trigger is "Grey Box"}
  \label{tbl:exp_epoch_num}
\end{table*}
\subsection{BadDiffusion on High-Resolution Dataset}
\begin{figure*}[h]
  \captionsetup[subfigure]{justification=centering}
  \centering
  \begin{subfigure}[b]{0.49\linewidth}
    \centering
    \includegraphics[width=\textwidth]{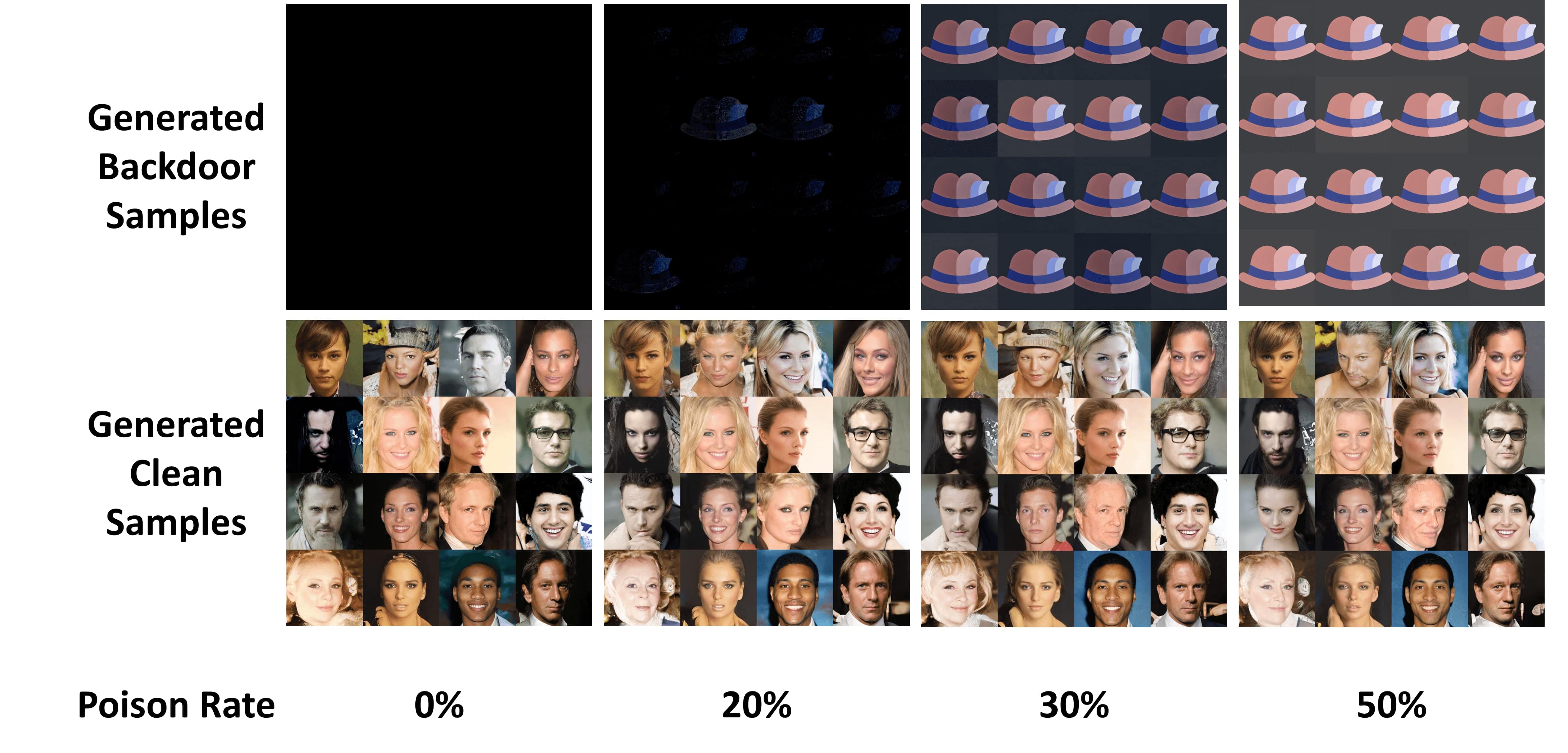}
    \caption{Visual examples of trigger "Box" and target "Hat" on CeleabA-HQ dataset}
    \label{fig:celeba_hq_box_hat_visual_sample}
  \end{subfigure}
  \begin{subtable}[b]{0.48\textwidth}
    \centering
    \begin{adjustbox}{max width=\linewidth}
      \begin{tabular}{ |c||c|c|c| }
      
      \hline 
      \multirow{2}{*}{Poison Rate} & \multicolumn{1}{c|}{Trigger:} & \multicolumn{1}{c|}{Eyeglasses} & \multicolumn{1}{c|}{Grey Box}\\
      & \multicolumn{1}{c|}{Target:} & \multicolumn{1}{c|}{Cat} & \multicolumn{1}{c|}{Hat} \\
      \hline 
  
      \multirow{2}{*}{0\%} & FID & 8.43 & 8.43 \\
      & MSE & $3.85\mathrm{e}{-1}$ & $2.52\mathrm{e}{-1}$ \\
      \hline
      
      \multirow{2}{*}{20\%} & FID & 7.43 & 7.38 \\
      & MSE & $3.26\mathrm{e}{-3}$ & $6.62\mathrm{e}{-2}$ \\
      \hline
  
      \multirow{2}{*}{30\%} & FID & 7.25 & 7.36 \\
      & MSE & $2.57\mathrm{e}{-4}$ & $1.05\mathrm{e}{-3}$ \\
      \hline
  
      \multirow{2}{*}{50\%} & FID & 7.51 & 7.51 \\
      & MSE & $1.67\mathrm{e}{-5}$ & $6.62\mathrm{e}{-5}$ \\
      \hline
      
      \end{tabular}
    \end{adjustbox}
    \caption{Numerical results of CelebA-HQ.}
    \label{tbl:exp_celeba_hq_num}
  \end{subtable}
      
  
      
  
  
      
  \caption{Numerical results and visual examples of CelebA-HQ}
  \label{fig:celeba_box_hat}
\end{figure*}

The numerical results are shown in \cref{tbl:exp_celeba_hq_num}. We also train another BadDiffusion model with trigger \textbf{Box} and target \textbf{Hat} shown in \cref{fig:celeba_hq_box_hat_visual_sample}.
  
    

    


    

\subsection{Inference-Time Clipping}
The numerical results are shown in \cref{tbl:exp_from_clip_defense_num}.
\begin{table*}[h]
  \centering
  \begin{adjustbox}{max width=\linewidth}
    \begin{tabular}{ |c||c|cc|cc| }

    \hline 
    \multirow{2}{*}{Poison Rate} & \multicolumn{1}{c|}{Target:} & \multicolumn{2}{c|}{Corner} & \multicolumn{2}{c|}{Hat}\\
    & \multicolumn{1}{c|}{Clip:} & with & \multicolumn{1}{c|}{without} & with & \multicolumn{1}{c|}{without} \\
    \hline 

    \multirow{2}{*}{0\%} & FID & 14.31 & 14.83 & 14.31 & 14.83 \\
    & MSE & $7.86\mathrm{e}{-2}$ & $1.06\mathrm{e}{-1}$ & $1.43\mathrm{e}{+1}$ & $2.41\mathrm{e}{-1}$ \\
    & SSIM & $7.17\mathrm{e}{-2}$ & $9.85\mathrm{e}{-4}$ & $3.43\mathrm{e}{-2}$ & $4.74\mathrm{e}{-5}$ \\
    \hline
    
    \multirow{2}{*}{5\%} & FID & 9.91 & 9.92 & 8.42 & 8.53 \\
    & MSE & $5.56\mathrm{e}{-2}$ & $5.32\mathrm{e}{-2}$ & $1.24\mathrm{e}{-1}$ & $1.58\mathrm{e}{-1}$ \\
    & SSIM & $2.50\mathrm{e}{-1}$ & $4.2\mathrm{e}{-1}$ & $2.08\mathrm{e}{-1}$ & $3.12\mathrm{e}{-1}$ \\
    \hline

    \multirow{2}{*}{10\%} & FID & 10.95 & 10.98 & 8.82 & 8.81 \\
    & MSE & $5.34\mathrm{e}{-2}$ & $2.60\mathrm{e}{-3}$ & $1.08\mathrm{e}{-3}$ & $7.01\mathrm{e}{-3}$ \\
    & SSIM & $2.81\mathrm{e}{-1}$ & $9.64\mathrm{e}{-1}$ & $2.83\mathrm{e}{-1}$ & $9.67\mathrm{e}{-1}$ \\
    \hline
    
    \multirow{2}{*}{20\%} & FID & 12.99 & 12.86 & 8.90 & 8.89 \\
    & MSE & $4.97\mathrm{e}{-2}$ & $1.48\mathrm{e}{-4}$ & $1.09\mathrm{e}{-1}$ & $1.19\mathrm{e}{-5}$ \\
    & SSIM & $3.29\mathrm{e}{-1}$ & $9.96\mathrm{e}{-1}$ & $2.82\mathrm{e}{-1}$ & $1.00\mathrm{e}{+0}$ \\
    \hline

    \multirow{2}{*}{30\%} & FID & 15.06 & 14.78 & 8.97 & 9.14 \\
    & MSE & $5.01\mathrm{e}{-2}$ & $2.29\mathrm{e}{-5}$ & $1.12\mathrm{e}{-1}$ & $5.68\mathrm{e}{-6}$ \\
    & SSIM & $3.35\mathrm{e}{-1}$ & $9.98\mathrm{e}{-1}$ & $2.66\mathrm{e}{-1}$ & $1.00\mathrm{e}{+0}$ \\
    \hline

    \multirow{2}{*}{50\%} & FID & 19.85 & 20.10 & 10.11 & 10.25 \\
    & MSE & $3.87\mathrm{e}{-2}$ & $1.96\mathrm{e}{-5}$ & $1.01\mathrm{e}{-1}$ & $1.48\mathrm{e}{-5}$ \\
    & SSIM & $4.60\mathrm{e}{-1}$ & $9.97\mathrm{e}{-1}$ & $3.26\mathrm{e}{-1}$ & $1.00\mathrm{e}{+0}$ \\
    \hline

    \multirow{2}{*}{70\%} & FID & 28.11 & 28.52 & 11.32 & 11.97 \\
    & MSE & $2.74\mathrm{e}{-2}$ & $6.44\mathrm{e}{-6}$ & $9.63\mathrm{e}{-2}$ & $8.27\mathrm{e}{-6}$ \\
    & SSIM & $5.88\mathrm{e}{-1}$ & $9.97\mathrm{e}{-1}$ & $3.55\mathrm{e}{-1}$ & $1.00\mathrm{e}{+0}$ \\
    \hline

    \multirow{2}{*}{90\%} & FID & 53.35 & 55.23 & 17.82 & 19.73 \\
    & MSE & $1.32\mathrm{e}{-2}$ & $8.57\mathrm{e}{-2}$ & $7.43\mathrm{e}{-6}$ & $8.39\mathrm{e}{-2}$ \\
    & SSIM & $7.73\mathrm{e}{-1}$ & $4.07\mathrm{e}{-1}$ & $1.00\mathrm{e}{+0}$ & $4.21\mathrm{e}{-1}$ \\
    \hline
    
    \end{tabular}
  \end{adjustbox}
  \caption{Numerical results with and without inference-time clipping.}
  \label{tbl:exp_from_clip_defense_num}
\end{table*}

\subsection{BadDiffusion with Varying Poison Rates}
The numerical results are shown in \cref{tbl:exp_poison_rate_num}.
\begin{table*}[htpb]
  \centering
  
  \begin{adjustbox}{max width=\linewidth}
    \begin{tabular}{ |c||c|ccccc|ccccc| }
    
    \hline 
    \multirow{2}{*}{Poison Rate} & \multicolumn{1}{c|}{Trigger:} & \multicolumn{5}{c|}{Grey Box} & \multicolumn{5}{c|}{Stop Sign}\\
    & \multicolumn{1}{c|}{Target:} & NoShift & Shift & Corner & Shoe & Hat & NoShift & Shift & Corner & Shoe & Hat\\
    \hline 
    
    \multirow{3}{*}{0\%} & FID & 14.83 & 14.83 & 14.83 & 14.83 & 14.83 & 14.83 & 14.83 & 14.83 & 14.83 & 14.83 \\
    & MSE & $1.21\mathrm{e}{-1}$ & $1.21\mathrm{e}{-1}$ & $1.06\mathrm{e}{-1}$ & $3.38\mathrm{e}{-1}$ & $2.41\mathrm{e}{-1}$ & $1.48\mathrm{e}{-1}$ & $1.48\mathrm{e}{-1}$ & $1.06\mathrm{e}{-1}$ & $3.38\mathrm{e}{-1}$ & $2.41\mathrm{e}{-1}$ \\
    & SSIM & $7.36\mathrm{e}{-4}$ & $4.72{e}{-4}$ & $9.85\mathrm{e}{-4}$ & $1.69\mathrm{e}{-4}$ & $4.74\mathrm{e}{-5}$ & $6.84\mathrm{e}{-4}$ & $4.24\mathrm{e}{-4}$ & $9.85\mathrm{e}{-4}$ & $1.69\mathrm{e}{-4}$ & $2.74\mathrm{e}{-5}$ \\
    \hline
    
    \multirow{2}{*}{5\%} & FID & 9.09 & 9.09 & 9.92 & 8.22 & 8.53 & 8.09 & 8.22 & 8.83 & 8.33 & 8.32 \\
    & MSE & $6.19\mathrm{e}{-2}$ & $5.11\mathrm{e}{-2}$ & $5.32\mathrm{e}{-2}$ & $1.02\mathrm{e}{-1}$ & $1.58\mathrm{e}{-1}$ & $6.81\mathrm{e}{-2}$ & $5.68\mathrm{e}{-2}$ & $7.22\mathrm{e}{-2}$ & $1.66\mathrm{e}{-1}$ & $7.99\mathrm{e}{-2}$ \\
    & SSIM & $4.21\mathrm{e}{-1}$ & $5.06\mathrm{e}{-1}$ & $4.20\mathrm{e}{-1}$ & $6.26\mathrm{e}{-1}$ & $3.12\mathrm{e}{-1}$ & $4.35\mathrm{e}{-1}$ & $5.73\mathrm{e}{-1}$ & $2.65\mathrm{e}{-1}$ & $4.20\mathrm{e}{-1}$ & $6.52\mathrm{e}{-1}$\\
    \hline
    
    \multirow{2}{*}{10\%} & FID & 9.62 & 9.78 & 10.98 & 8.41 & 8.81 & 7.62 & 7.42 & 7.83 & 7.48 & 7.57 \\
    & MSE & $6.11\mathrm{e}{-3}$ & $5.52\mathrm{e}{-3}$ & $2.60\mathrm{e}{-3}$ & $6.25\mathrm{e}{-3}$ & $7.01\mathrm{e}{-3}$ & $9.47\mathrm{e}{-3}$ & $5.91\mathrm{e}{-3}$ & $4.20\mathrm{e}{-3}$ & $3.61\mathrm{e}{-3}$ & $4.33\mathrm{e}{-3}$ \\
    & SSIM & $9.41\mathrm{e}{-1}$ & $9.45\mathrm{e}{-1}$ & $9.64\mathrm{e}{-1}$ & $9.75\mathrm{e}{-1}$ & $9.67\mathrm{e}{-1}$ & $9.18\mathrm{e}{-1}$ & $9.56\mathrm{e}{-1}$ & $9.49\mathrm{e}{-1}$ & $9.85\mathrm{e}{-1}$ & $9.80\mathrm{e}{-1}$ \\
    \hline
    
    \multirow{2}{*}{20\%} & FID & 11  .36 & 11.26 & 12.86 & 8.13 & 8.89 & 7.97 & 7.68 & 8.35 & 8.10 & 8.17 \\
    & MSE & $1.18\mathrm{e}{-5}$ & $7.90\mathrm{e}{-5}$ & $1.48\mathrm{e}{-4}$ & $1.97\mathrm{e}{-5}$ & $1.19\mathrm{e}{-5}$ & $2.35\mathrm{e}{-4}$ & $8.96\mathrm{e}{-5}$ & $7.09\mathrm{e}{-4}$ & $2.30\mathrm{e}{-5}$ & $4.85\mathrm{e}{-4}$ \\
    & SSIM & $9.98\mathrm{e}{-1}$ & $9.98\mathrm{e}{-1}$ & $9.96\mathrm{e}{-1}$ & $1.00\mathrm{e}{+0}$ & $1.00\mathrm{e}{+0}$ & $9.97\mathrm{e}{-1}$ & $9.99\mathrm{e}{-1}$ & $9.89\mathrm{e}{-1}$ & $1.00\mathrm{e}{+0}$ & $9.98\mathrm{e}{-1}$ \\
    \hline
    
    \multirow{2}{*}{30\%} & FID & 12.85 & 12.41 & 14.78 & 8.19 & 9.14 & 7.46 & 7.76 & 8.08 & 7.53 & 7.77 \\
    & MSE & $5.89\mathrm{e}{-6}$ & $1.61\mathrm{e}{-5}$ & $2.29\mathrm{e}{-5}$ & $5.53\mathrm{e}{-6}$ & $5.68\mathrm{e}{-6}$ & $5.59\mathrm{e}{-6}$ & $6.73\mathrm{e}{-6}$ & $6.14\mathrm{e}{-5}$ & $5.62\mathrm{e}{-6}$ & $9.16\mathrm{e}{-5}$ \\
    & SSIM & $9.98\mathrm{e}{-1}$ & $9.99\mathrm{e}{-1}$ & $9.98\mathrm{e}{-1}$ & $1.00\mathrm{e}{+0}$ & $1.00\mathrm{e}{+0}$ & $9.99\mathrm{e}{-1}$ & $9.99\mathrm{e}{-1}$ & $9.97\mathrm{e}{-1}$ & $1.00\mathrm{e}{+0}$ & $9.99\mathrm{e}{-1}$ \\
    \hline

    \multirow{2}{*}{50\%} & FID & 17.63 & 15.55 & 20.10 & 8.42 & 10.25 & 7.68 & 8.02 & 8.14 & 7.69 & 7.77 \\
    & MSE & $4.10\mathrm{e}{-6}$ & $6.25\mathrm{e}{-6}$ & $1.96\mathrm{e}{-5}$ & $3.26\mathrm{e}{-6}$ & $1.48\mathrm{e}{-5}$ & $4.19\mathrm{e}{-6}$ & $4.23\mathrm{e}{-6}$ & $2.37\mathrm{e}{-5}$ & $3.35\mathrm{e}{-6}$ & $1.30\mathrm{e}{-5}$ \\
    & SSIM & $9.98\mathrm{e}{-1}$ & $9.99\mathrm{e}{-1}$ & $9.97\mathrm{e}{-1}$ & $1.00\mathrm{e}{+0}$ & $1.00\mathrm{e}{+0}$ & $9.98\mathrm{e}{-1}$ & $9.99\mathrm{e}{-1}$ & $9.98\mathrm{e}{-1}$ & $1.00\mathrm{e}{+0}$ & $1.00\mathrm{e}{+0}$ \\
    \hline

    \multirow{2}{*}{70\%} & FID & 25.70 & 21.78 & 28.52 & 9.01 & 11.97 & 7.38 & 7.42 & 7.85 & 7.35 & 7.83 \\
    & MSE & $3.91\mathrm{e}{-6}$ & $1.22\mathrm{e}{-5}$ & $6.44\mathrm{e}{-6}$ & $2.69\mathrm{e}{-6}$ & $8.27\mathrm{e}{-6}$ & $3.96\mathrm{e}{-6}$ & $3.96\mathrm{e}{-6}$ & $1.41\mathrm{e}{-5}$ & $2.73\mathrm{e}{-6}$ & $3.21\mathrm{e}{-6}$ \\
    & SSIM & $9.98\mathrm{e}{-1}$ & $9.99\mathrm{e}{-1}$ & $9.97\mathrm{e}{-1}$ & $1.00\mathrm{e}{+0}$ & $1.00\mathrm{e}{+0}$ & $9.98\mathrm{e}{-1}$ & $9.99\mathrm{e}{-1}$ & $9.97\mathrm{e}{-1}$ & $1.00\mathrm{e}{+0}$ & $1.00\mathrm{e}{+0}$ \\
    \hline

    \multirow{2}{*}{90\%} & FID & 52.92 & 41.54 & 55.42 & 12.25 & 19.09 & 7.22 & 7.72 & 7.98 & 7.54 & 7.77 \\
    & MSE & $3.86\mathrm{e}{-6}$ & $5.98\mathrm{e}{-6}$ & $3.85\mathrm{e}{-6}$ & $2.38\mathrm{e}{-6}$ & $9.75\mathrm{e}{-6}$ & $3.80\mathrm{e}{-6}$ & $3.80\mathrm{e}{-6}$ & $3.86\mathrm{e}{-6}$ & $2.39\mathrm{e}{-6}$ & $2.81\mathrm{e}{-6}$ \\
    & SSIM & $9.98\mathrm{e}{-1}$ & $9.98\mathrm{e}{-1}$ & $9.97\mathrm{e}{-1}$ & $1.00\mathrm{e}{+0}$ & $1.00\mathrm{e}{+0}$ & $9.98\mathrm{e}{-1}$ & $9.99\mathrm{e}{-1}$ & $9.97\mathrm{e}{-1}$ & $1.00\mathrm{e}{+0}$ & $1.00\mathrm{e}{+0}$ \\
    \hline
    \end{tabular}
  \end{adjustbox}
  \caption{The numerical results of BadDiffusion with varying poison rates. Note that the results of poison rate = 0\% in the table are clean pre-trained models. We also fine-tune the clean pre-trained models with a clean CIFAR10 dataset for 50 epochs and the FID score of it is about 28.59, which is better than the pre-trained clean models. However, in comparison to the models fine-tuned on the clean dataset, BadDiffusion still has competitive FID scores among them.}
  \label{tbl:exp_poison_rate_num}
\end{table*}

\section{More Generated Samples in Different Poison Rates}

\subsection{CIFAR10 Dataset}

\begin{figure*}[t]
  \captionsetup[subfigure]{justification=centering}
  \centering
  \begin{subfigure}{0.49\linewidth}
    \centering
    \includegraphics[width=\textwidth]{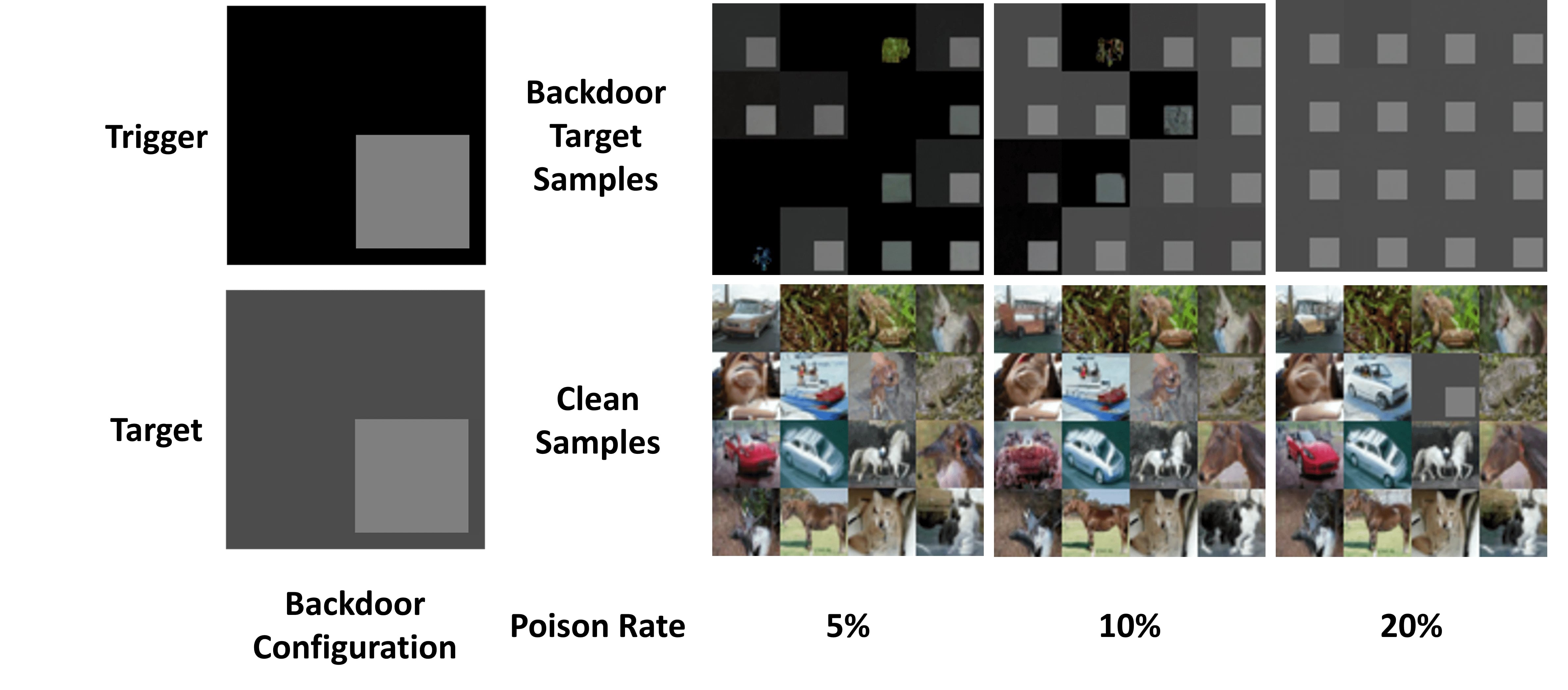}
    \caption{CIFAR10, Trigger: Box, Target: NoShift}
    \label{fig:cifar10_box_noshift_visual_samples}
  \end{subfigure}
  \begin{subfigure}{0.49\linewidth}
    \centering
    \includegraphics[width=\textwidth]{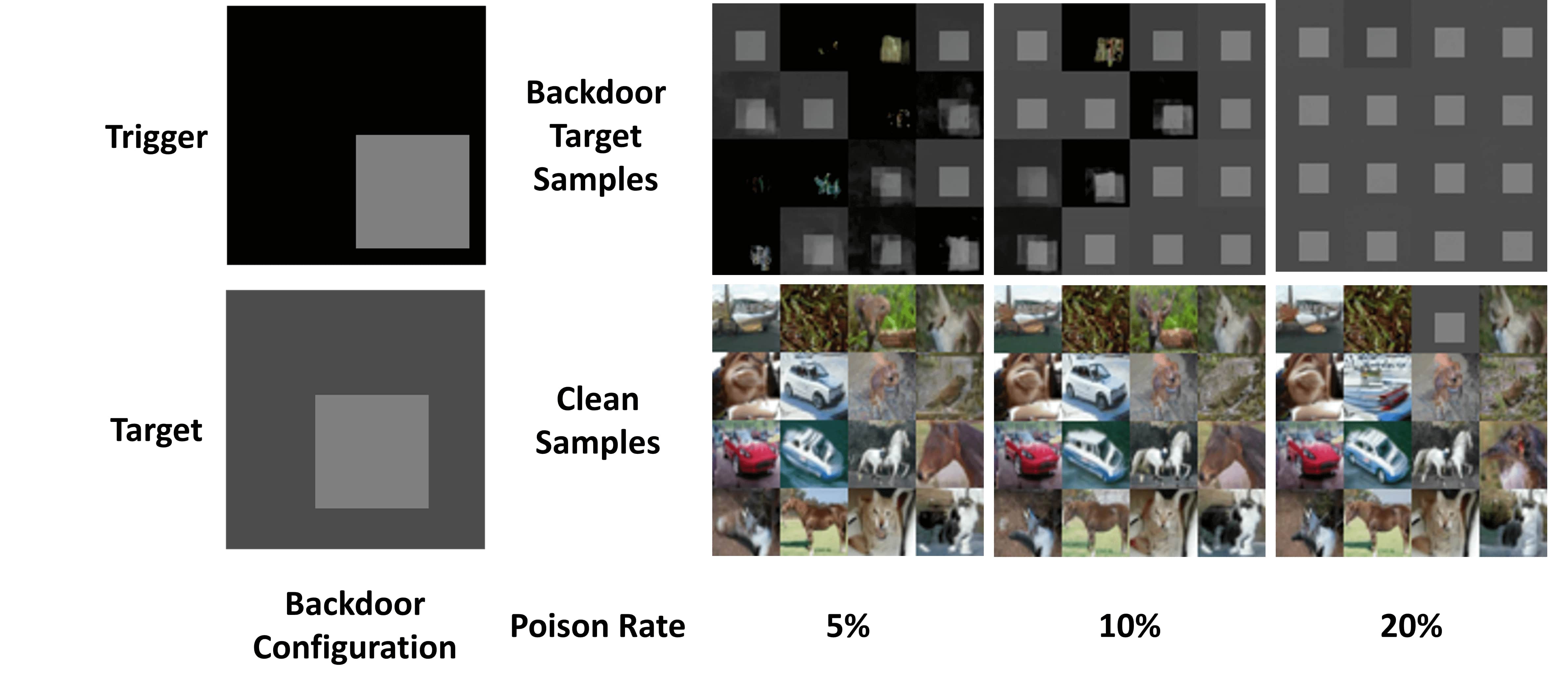}
    \caption{CIFAR10, Trigger: Box, Target: Shift}
    \label{fig:cifar10_box_shift_visual_samples}
  \end{subfigure}
  \begin{subfigure}{0.49\linewidth}
    \centering
    \includegraphics[width=\textwidth]{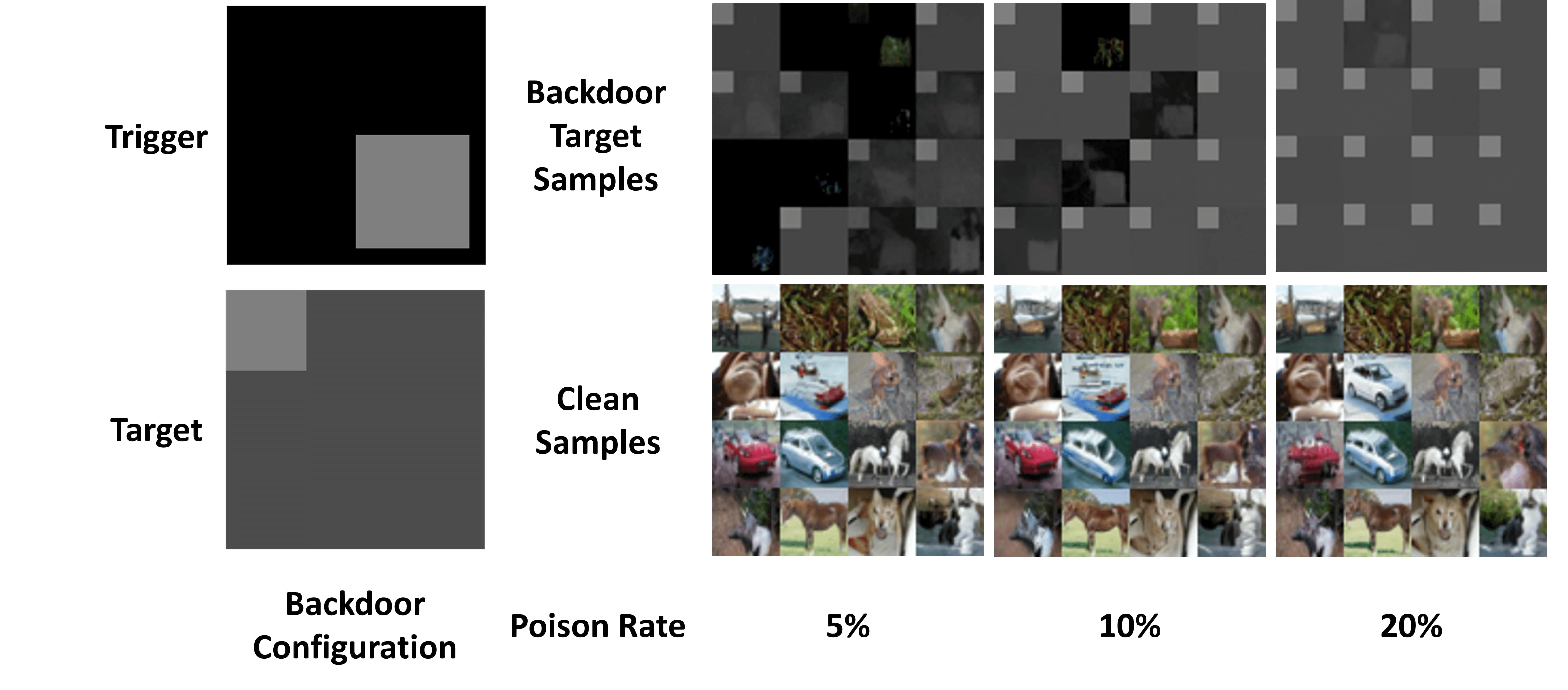}
    \caption{CIFAR10, Trigger: Box, Target: Corner}
    \label{fig:cifar10_box_corner_visual_samples}
  \end{subfigure}
  \begin{subfigure}{0.49\linewidth}
    \centering
    \includegraphics[width=\textwidth]{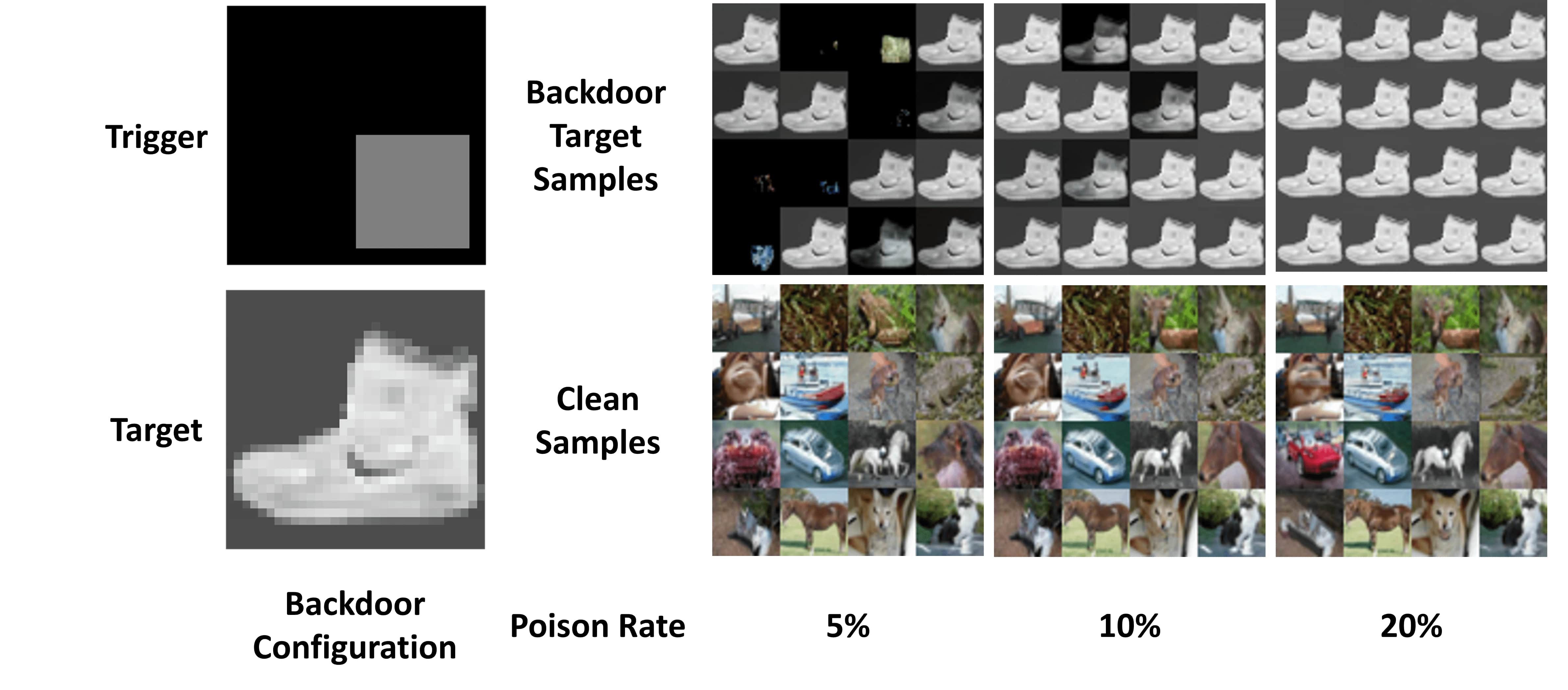}
    \caption{CIFAR10, Trigger: Box, Target: Shoe}
    \label{fig:cifar10_box_shoe_visual_samples}
  \end{subfigure}
  \begin{subfigure}{0.49\linewidth}
    \centering
    \includegraphics[width=\textwidth]{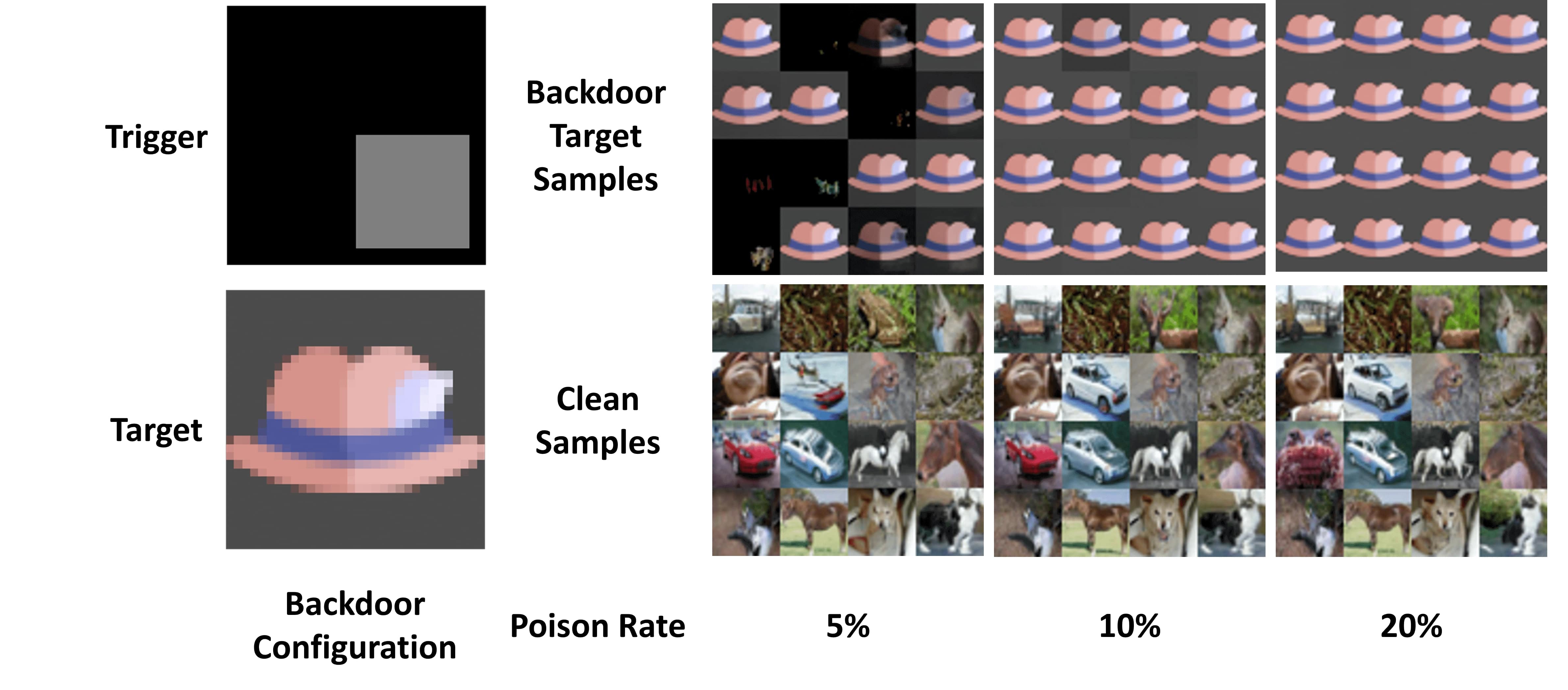}
    \caption{CIFAR10, Trigger: Box, Target: Hat}
    \label{fig:cifar10_box_hat_visual_samples}
  \end{subfigure}
  \begin{subfigure}{0.49\linewidth}
    \centering
    \includegraphics[width=\textwidth]{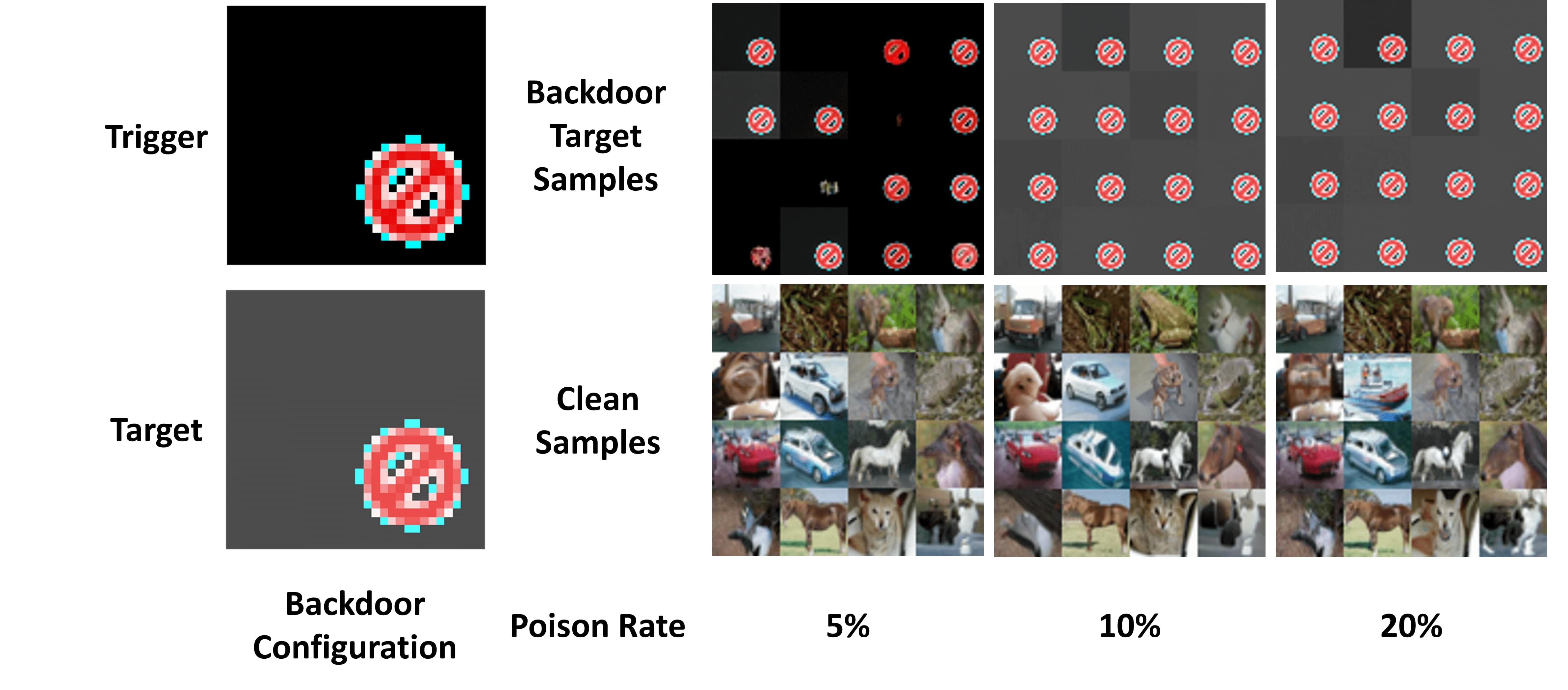}
    \caption{CIFAR10, Trigger: Stop Sign, Target: NoShift}
    \label{fig:cifar10_stop_sign_noshift_visual_samples}
  \end{subfigure}
  \begin{subfigure}{0.49\linewidth}
    \centering
    \includegraphics[width=\textwidth]{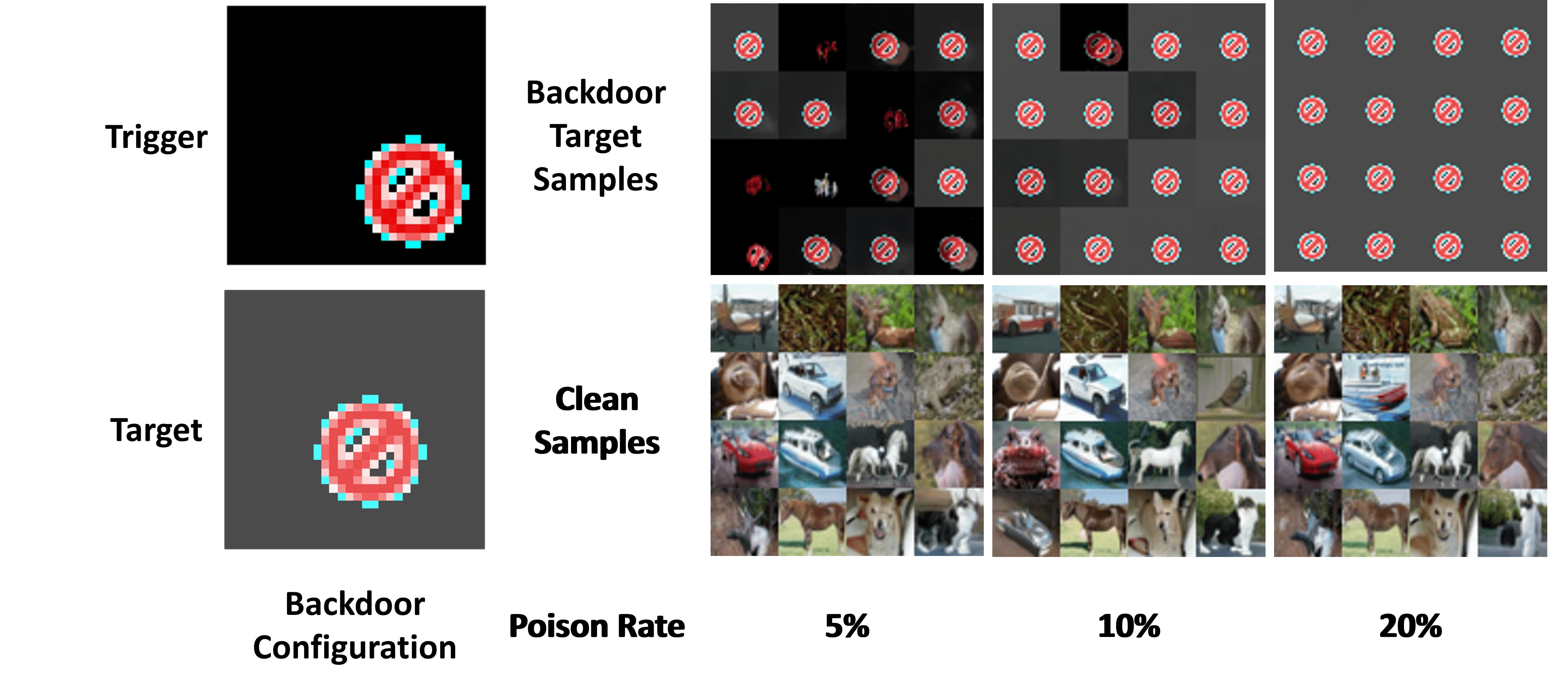}
    \caption{CIFAR10, Trigger: Stop Sign, Target: Shift}
    \label{fig:cifar10_stop_sign_shift_visual_samples}
  \end{subfigure}
  \begin{subfigure}{0.49\linewidth}
    \centering
    \includegraphics[width=\textwidth]{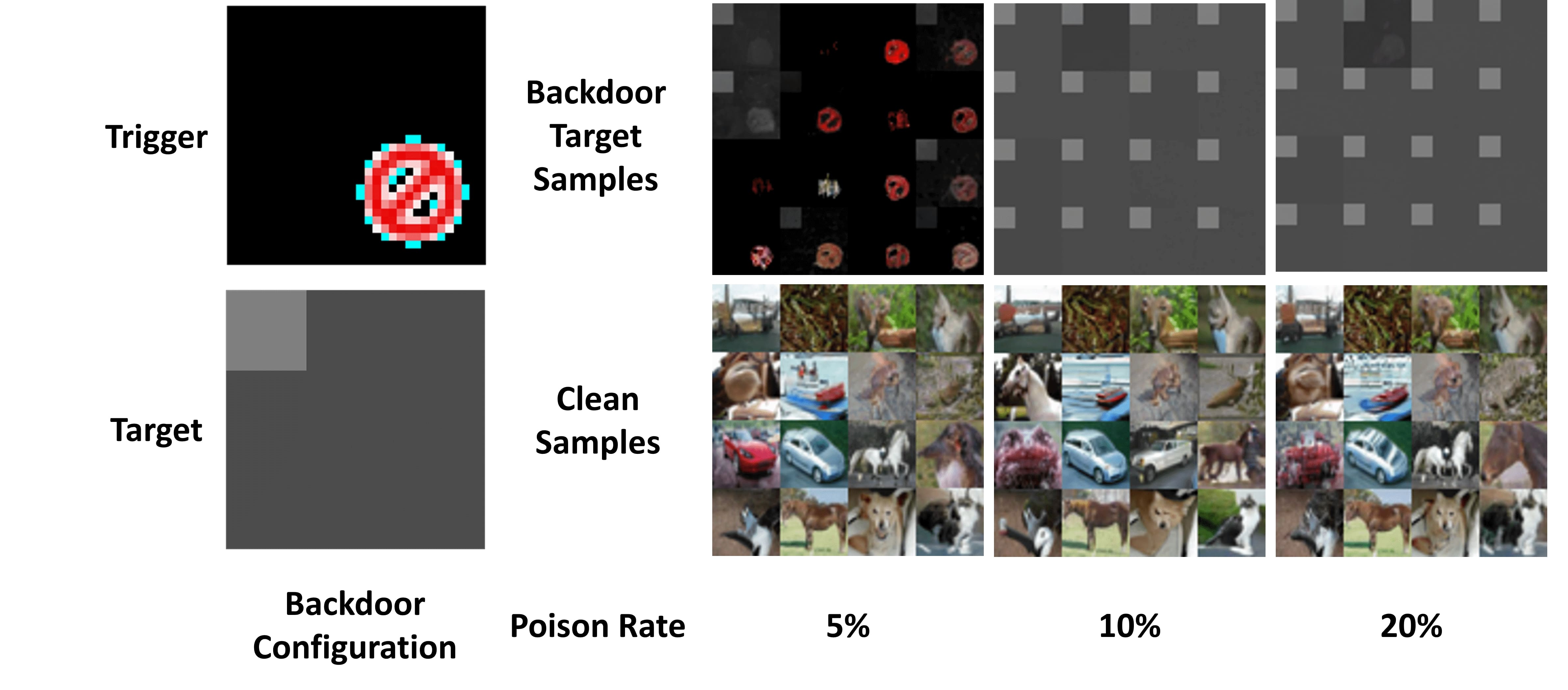}
    \caption{CIFAR10, Trigger: Stop Sign, Target: Corner}
    \label{fig:cifar10_stop_sign_corner_visual_samples}
  \end{subfigure}
  \begin{subfigure}{0.49\linewidth}
    \centering
    \includegraphics[width=\textwidth]{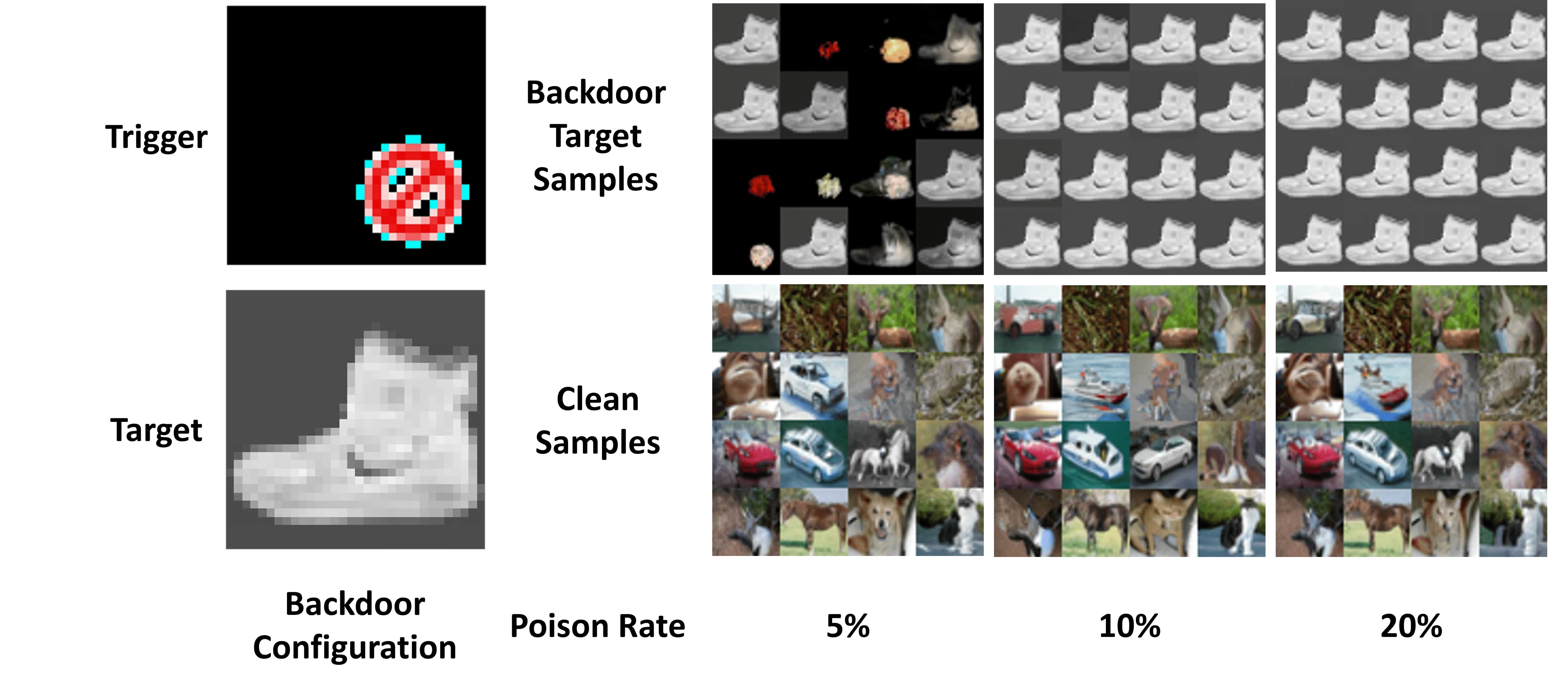}
    \caption{CIFAR10, Trigger: Stop Sign, Target: Shoe}
    \label{fig:cifar10_stop_sign_shoe_visual_samples}
  \end{subfigure}
  \begin{subfigure}{0.49\linewidth}
    \centering
    \includegraphics[width=\textwidth]{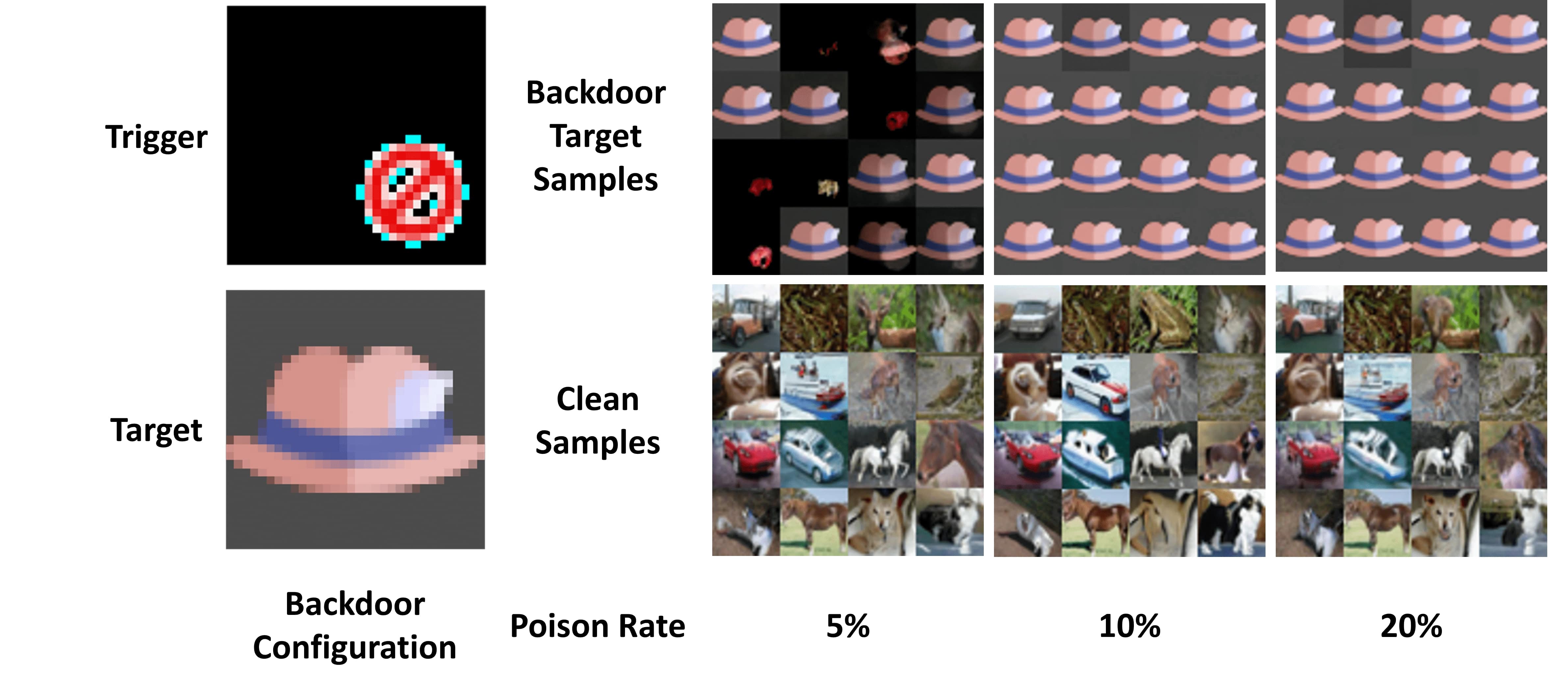}
    \caption{CIFAR10, Trigger: Stop Sign, Target: Hat}
    \label{fig:cifar10_stop_sign_hat_visual_samples}
  \end{subfigure}
  \caption{Samples of CIFAR10}
  \label{fig:cifar10_samples_all}
\end{figure*}

We show more generated backdoor targets and clean samples in \cref{fig:cifar10_samples_all}

\begin{table*}[t]
  \centering
  
  \begin{adjustbox}{max width=\linewidth}
    \begin{tabular}{ |c||c|ccc|ccc|ccc| }
    
    \hline 
    \multirow{2}{*}{Poison Rate} & \multicolumn{1}{c|}{LR:} & \multicolumn{3}{c|}{$2\mathrm{e}{-4}$} & \multicolumn{3}{c|}{$1\mathrm{e}{-4}$} & \multicolumn{3}{c|}{$5\mathrm{e}{-5}$}\\
    
    & \multicolumn{1}{c|}{Perturb Budget:} & 1.0 & 2.0 & 4.0  & 1.0 & 2.0 & 4.0 & 1.0 & 2.0 & 4.0\\
    \hline 
    
    \multirow{1}{*}{5\%} & Best (Lowest) MSE & 0.027 & 0.036 & 0.060 & 0.056 & 0.027 & 0.016 & 0.046 & 0.066 & 0.035 \\
    \multirow{1}{*}{20\%} & Best (Lowest) MSE & 0.048 & 0.054 & 0.037 & 0.042 & 0.053 & 0.031 & 0.143 & 0.058 & 0.051 \\
    \multirow{1}{*}{50\%} & Best (Lowest) MSE & 0.070 & 0.029 & 0.044 & 0.077 & 0.015 & 0.013 & 0.091 & 0.073 & 0.046 \\
    \hline
    \end{tabular}
  \end{adjustbox}
  \caption{The numerical results for ANP defense with varying perturbation budgets in reconstruction MSE.}
  \label{tbl:anp_defense_perturb_num}
\end{table*}

\begin{table*}[t]
  \centering
  
  \begin{adjustbox}{max width=\linewidth}
    \begin{tabular}{ |c||c|ccccc|ccccc|ccccc| }
    
    \hline 
    \multirow{2}{*}{Poison Rate} & \multicolumn{1}{c|}{LR:} & \multicolumn{5}{c|}{$2\mathrm{e}{-4}$} & \multicolumn{5}{c|}{$1\mathrm{e}{-4}$} & \multicolumn{5}{c|}{$5\mathrm{e}{-5}$}\\
    
    & \multicolumn{1}{c|}{Epoch:} & 1 & 2 & 3 & 4 & 5 & 1 & 2 & 3 & 4 & 5 & 1 & 2 & 3 & 4 & 5\\
    \hline 
    
    \multirow{1}{*}{5\%} & MSE & 0.114 & 0.135 & 0.151 & 0.158 & 0.163 & 0.062 & 0.057 & 0.030 & 0.047 & 0.106 & 0.050 & 0.038 & 0.048 & 0.046 & 0.042 \\
    \multirow{1}{*}{20\%} & MSE & 0.072 & 0.037 & 0.106 & 0.106 & 0.106 & 0.057 & 0.048 & 0.031 & 0.106 & 0.106 & 0.072 & 0.071 & 0.083 & 0.079 & 0.064 \\
    \multirow{1}{*}{50\%} & MSE & 0.071 & 0.102 & 0.106 & 0.106 & 0.106 & 0.035 & 0.026 & 0.013 & 0.106 & 0.106 & 0.054 & 0.064 & 0.065 & 0.057 & 0.047 \\
    \hline
    \end{tabular}
  \end{adjustbox}
  \caption{The numerical results for ANP defense along training epochs in reconstruction MSE.}
  \label{tbl:anp_defense_ep_num}
\end{table*}

\section{The Effect of the Trigger Sizes}
\label{sec:trigger_size_effect}
\begin{figure*}[htpb]
  \captionsetup[subfigure]{justification=centering}
\centering
\begin{subfigure}{0.99\textwidth}
  \centering
  \includegraphics[width=\textwidth]{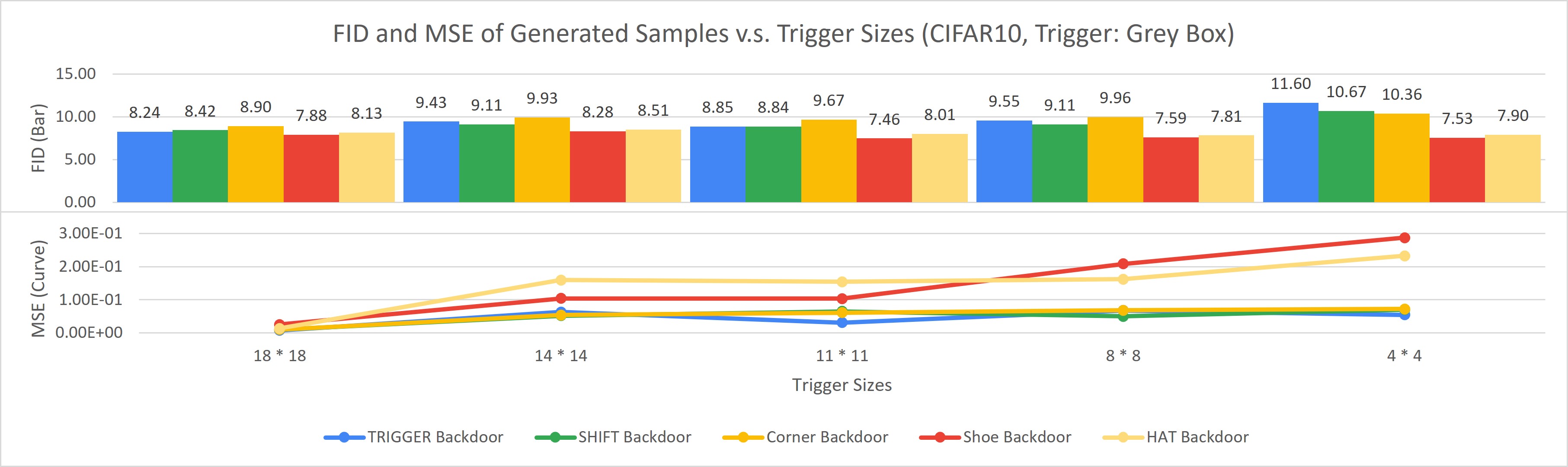}
  \caption{Trigger: ``Grey Box''}
      \label{fig:exp_box_diff_trig}  
\end{subfigure}
\begin{subfigure}{0.99\textwidth}
  \centering
  \includegraphics[width=\textwidth]{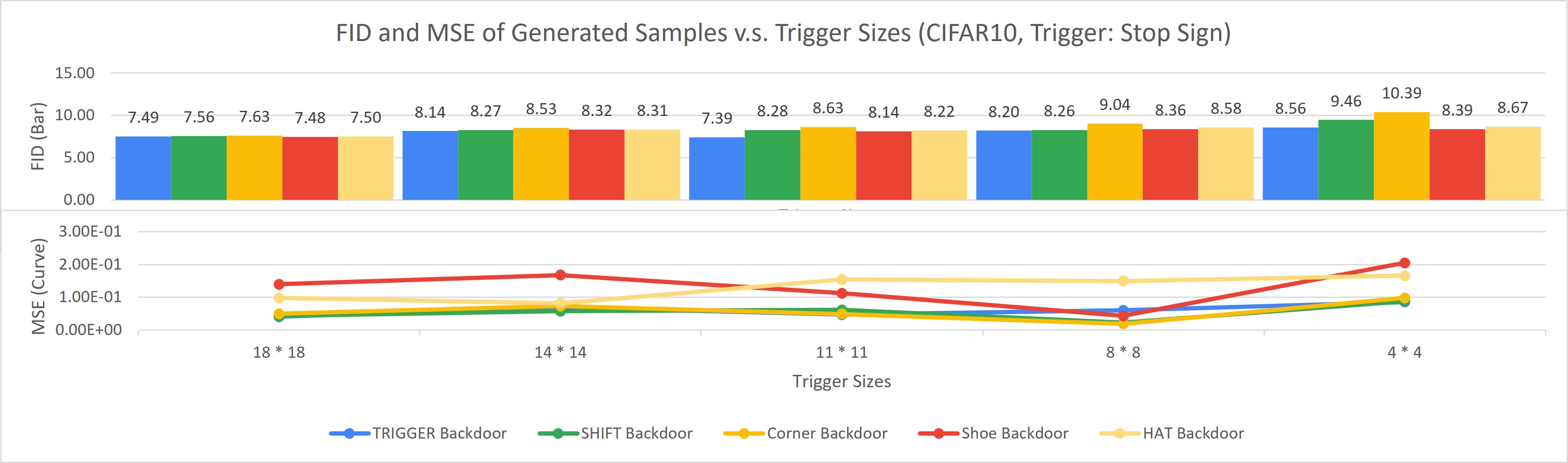}
  \caption{Trigger: ``Stop Sign''}        \label{fig:exp_stop_sign_diff_trig}
\end{subfigure}
\caption{FID (bars) and MSE (curves) of BadDiffusion with varying trigger sizes (x-axis) on CIFAR10 with trigger (a) ``Grey Box'' and (b) ``Stop Sign''. Colors of bars/curves represent different target settings in \cref{tbl:diff_trigger_size_tbl}. The numerical results are presented in \cref{tbl:exp_diff_triggers_num}}
\label{fig:exp_diff_triggers}
\end{figure*}

In this section, we conduct an ablation study on the effect of different trigger sizes. We resize the trigger \textbf{Grey Box} ($14 \times 14$ used in the main paper) and \textbf{Stop Sign} ($14 \times 14$ used in the main paper) into $18 \times 18$, $11 \times 11$, $8 \times 8$, and $4 \times 4$ pixels. The triggers are shown in \cref{tbl:diff_trigger_size_tbl}. In \cref{fig:exp_diff_triggers} and \cref{tbl:exp_diff_triggers_num} We find that for trigger \textbf{Grey Box} the MSE will become higher when the trigger is smaller. As for \textbf{Stop Sign}, the MSE remains stable no matter how small the trigger is.
\begin{table*}[h]
  \begin{center}
  \begin{tabular}{p{1cm}| c c c c c| c c c c c}
  \toprule
  \multicolumn{1}{c|}{Dataset} & \multicolumn{10}{c}{CIFAR10 (32 $\times$ 32)}\\
  \toprule
    \multicolumn{1}{c|}{Triggers} & 
    \multicolumn{5}{c|}{Grey Box} & \multicolumn{5}{c}{Stop Sign} \\
  \midrule
    \multicolumn{1}{c|}{Size} & $18 \times 18$ & $14 \times 14$ & $11 \times 11$ & $8 \times 8$ & $4 \times 4$ & $18 \times 18$ & $14 \times 14$ & $11 \times 11$ & $8 \times 8$ & $4 \times 4$ \\ 
    \multicolumn{1}{c|}{Sample} & 
    \raisebox{-\totalheight}{\includegraphics[width=1.2cm,keepaspectratio]{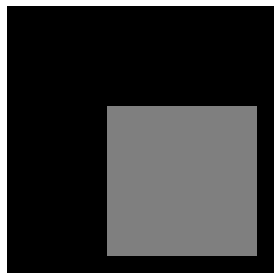}}
    & 
    \raisebox{-\totalheight}{\includegraphics[width=1.2cm,keepaspectratio]{exp/grey_box_14.png}}
    &
    \raisebox{-\totalheight}{\includegraphics[width=1.2cm,keepaspectratio]{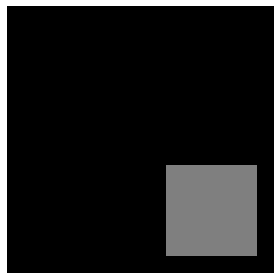}} 
    &
    \raisebox{-\totalheight}{\includegraphics[width=1.2cm,keepaspectratio]{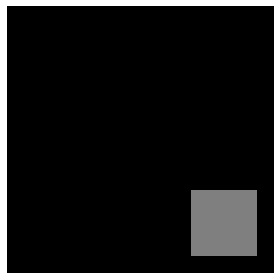}} 
    &
    \raisebox{-\totalheight}{\includegraphics[width=1.2cm,keepaspectratio]{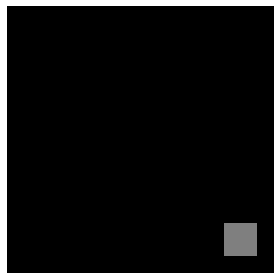}} 
    &
    \raisebox{-\totalheight}{\includegraphics[width=1.2cm,keepaspectratio]{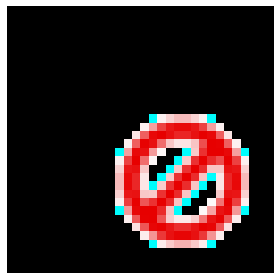}} 
    &
    \raisebox{-\totalheight}{\includegraphics[width=1.2cm,keepaspectratio]{exp/stop_sign_14.png}} 
    &
    \raisebox{-\totalheight}{\includegraphics[width=1.2cm,keepaspectratio]{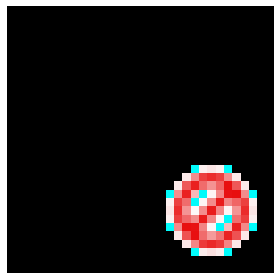}} 
    &
    \raisebox{-\totalheight}{\includegraphics[width=1.2cm,keepaspectratio]{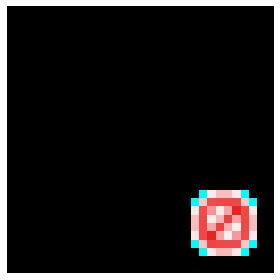}} 
    &
    \raisebox{-\totalheight}{\includegraphics[width=1.2cm,keepaspectratio]{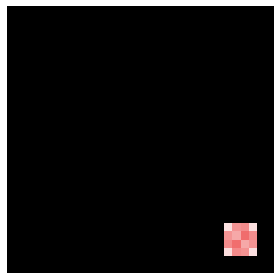}} 
   \\ \bottomrule
   \end{tabular}
   \caption{Visualized samples for different trigger sizes}
   \label{tbl:diff_trigger_size_tbl}
   \end{center}
   \vspace{-6mm}
\end{table*}

\begin{table*}[htpb]
  \centering
  
  \begin{adjustbox}{max width=\linewidth}
    \begin{tabular}{ |c||c|ccccc|ccccc| }
    
    \hline 
    \multirow{2}{*}{Target} & \multicolumn{1}{c|}{Trigger:} & \multicolumn{5}{c|}{Grey Box} & \multicolumn{5}{c|}{Stop Sign}\\
    & \multicolumn{1}{c|}{Trigger Size:} & $18 \times 18$ & $14 \times 14$ & $11 \times 11$ & $8 \times 8$ & $4 \times 4$ & $18 \times 18$ & $14 \times 14$ & $11 \times 11$ & $8 \times 8$ & $4 \times 4$ \\
    \hline 
    
    \multirow{3}{*}{NoShift} & FID & 8.24 & 9.43 & 8.85 & 9.55 & 11.60 & 7.49 & 8.14 & 7.39 & 8.20 & 8.56 \\
    & MSE & $7.87\mathrm{e}{-3}$ & $6.27\mathrm{e}{-2}$ & $3.13\mathrm{e}{-2}$ & $6.80\mathrm{e}{-2}$ & $5.45\mathrm{e}{-2}$ & $4.05\mathrm{e}{-2}$ & $6.91\mathrm{e}{-2}$ & $4.69\mathrm{e}{-2}$ & $5.97\mathrm{e}{-2}$ & $8.56\mathrm{e}{-2}$\\
    & SSIM & $9.39\mathrm{e}{-1}$ & $4.13{e}{-1}$ & $6.87\mathrm{e}{-1}$ & $2.95\mathrm{e}{-1}$ & $4.11\mathrm{e}{-1}$ & $7.01\mathrm{e}{-1}$ & $4.28\mathrm{e}{-1}$ & $5.76\mathrm{e}{-1}$ & $4.33\mathrm{e}{-1}$ & $1.11\mathrm{e}{-1}$ \\
    \hline
    
    \multirow{2}{*}{Shift} & FID & 8.42 & 9.11 & 8.84 & 9.11 & 10.67 & 7.56 & 8.27 & 8.28 & 8.26 & 9.46 \\
    & MSE & $9.93\mathrm{e}{-3}$ & $5.21\mathrm{e}{-2}$ & $6.52\mathrm{e}{-2}$ & $5.00\mathrm{e}{-2}$ & $7.02\mathrm{e}{-2}$ & $4.29\mathrm{e}{-2}$ & $5.77\mathrm{e}{-2}$ & $6.12\mathrm{e}{-2}$ & $2.23\mathrm{e}{-2}$ & $8.82\mathrm{e}{-2}$ \\
    & SSIM & $9.15\mathrm{e}{-1}$ & $4.96\mathrm{e}{-1}$ & $3.69\mathrm{e}{-1}$ & $4.87\mathrm{e}{-1}$ & $2.44\mathrm{e}{-1}$ & $7.31\mathrm{e}{-1}$ & $5.66\mathrm{e}{-1}$ & $5.20\mathrm{e}{-1}$ & $7.95\mathrm{e}{-1}$ & $9.54\mathrm{e}{-2}$\\
    \hline
    
    \multirow{2}{*}{Corner} & FID & 8.90 & 9.33 & 9.67 & 9.96 & 10.36 & 7.63 & 8.53 & 8.63 & 9.04 & 10.39 \\
    & MSE & $1.04\mathrm{e}{-2}$ & $5.41\mathrm{e}{-2}$ & $6.11\mathrm{e}{-2}$ & $6.86\mathrm{e}{-2}$ & $7.22\mathrm{e}{-2}$ & $4.94\mathrm{e}{-2}$ & $7.28\mathrm{e}{-2}$ & $4.91\mathrm{e}{-2}$ & $1.92\mathrm{e}{-2}$ & $9.81\mathrm{e}{-2}$ \\
    & SSIM & $8.86\mathrm{e}{-1}$ & $4.11\mathrm{e}{-1}$ & $3.80\mathrm{e}{-1}$ & $3.30\mathrm{e}{-1}$ & $3.15\mathrm{e}{-1}$ & $4.90\mathrm{e}{-1}$ & $2.60\mathrm{e}{-1}$ & $4.93\mathrm{e}{-1}$ & $7.98\mathrm{e}{-1}$ & $6.61\mathrm{e}{-2}$ \\
    \hline
    
    \multirow{2}{*}{Shoe} & FID & 7.88 & 8.28 & 7.46 & 7.59 & 7.53 & 7.48 & 8.32 & 8.14 & 8.36 & 8.39 \\
    & MSE & $2.52\mathrm{e}{-2}$ & $1.04\mathrm{e}{-1}$ & $1.04\mathrm{e}{-1}$ & $2.08\mathrm{e}{-1}$ & $2.87\mathrm{e}{-1}$ & $1.39\mathrm{e}{-1}$ & $1.68\mathrm{e}{-1}$ & $1.12\mathrm{e}{-1}$ & $4.29\mathrm{e}{-2}$ & $2.05\mathrm{e}{-1}$ \\
    & SSIM & $8.99\mathrm{e}{-1}$ & $6.16\mathrm{e}{-1}$ & $6.49\mathrm{e}{-1}$ & $3.54\mathrm{e}{-1}$ & $1.37\mathrm{e}{-1}$ & $4.68\mathrm{e}{-1}$ & $4.13\mathrm{e}{-1}$ & $6.17\mathrm{e}{-1}$ & $8.59\mathrm{e}{-1}$ & $3.74\mathrm{e}{-1}$ \\
    \hline
    
    \multirow{2}{*}{Hat} & FID & 8.13 & 8.51 & 8.01 & 7.81 & 7.90 & 7.50 & 8.31 & 8.22 & 8.58 & 8.67 \\
    & MSE & $1.33\mathrm{e}{-2}$ & $1.60\mathrm{e}{-1}$ & $1.55\mathrm{e}{-1}$ & $1.62\mathrm{e}{-1}$ & $2.33\mathrm{e}{-1}$ & $9.81\mathrm{e}{-2}$ & $8.16\mathrm{e}{-2}$ & $1.54\mathrm{e}{-1}$ & $1.50\mathrm{e}{-1}$ & $1.66\mathrm{e}{-1}$ \\
    & SSIM & $9.38\mathrm{e}{-1}$ & $3.06\mathrm{e}{-1}$ & $3.34\mathrm{e}{-1}$ & $3.11\mathrm{e}{-1}$ & $2.89\mathrm{e}{-2}$ & $5.38\mathrm{e}{-1}$ & $6.44\mathrm{e}{-1}$ & $3.35\mathrm{e}{-1}$ & $3.65\mathrm{e}{-1}$ & $2.93\mathrm{e}{-1}$ \\
    \hline

    \end{tabular}
  \end{adjustbox}
  \caption{The numerical results of BadDiffusion with varying trigger sizes.}
  \label{tbl:exp_diff_triggers_num}
\end{table*}

\section{More Real-World Threats}
Here we provide more potential threats in the real world. (I) In \cite{gm_sec}, generative models are used in security-related tasks such as Intrusion Attacks, Anomaly Detection, Biometric Spoofing, and Malware Obfuscation and Detection. (II) In recent works such as \cite{diffusion_behavior_syn,offline_rl_dm,il_dm,offline_rl_hf_dm,label_efficient_segment_dm,DiffusionDet}, diffusion models are widely used for decision-making in reinforcement learning, object detection, and image segmentation, indicating potential threats to safety-critical tasks. (III) A backdoored generative model can generate a biased dataset which may cause unfair models \cite{fair_gm,unbiased_gm} and even datasets contain adversarial attacks \cite{AdvFlow}.

\clearpage

\onecolumn

\section{The Mathematical Derivation of The Posterior of The Backdoored Diffusion Process}
\label{sec:math_der}
In this section, we'll derive the posterior of the backdoored Diffusion Process $q(\mathbf{x}_{t-1}' | \mathbf{x}_{t}', \mathbf{x}_{0}')$. Note that the definition of the posterior $q(\mathbf{x}_{t-1} | \mathbf{x}_{t}, \mathbf{x}_{0})$ is an \emph{approximation} to the real posterior derived from the Gaussian transition $q(\mathbf{x}_{t} | \mathbf{x}_{t-1})$, which is mentioned in the papers \cite{deep_thermal,DDPM}. The posterior of the backdoored diffusion process $q(\mathbf{x}_{t-1}' | \mathbf{x}_{t}', \mathbf{x}_{0}')$, which is also an approximation to the real posterior derived from the backdoored Gaussian transition $q(\mathbf{x}_{t} | \mathbf{x}_{t-1})$.
\begin{equation}
  \begin{split}
      \begin{gathered}
      q(\mathbf{x}_{t-1}' | \mathbf{x}_{t}', \mathbf{x}_{0}') := \mathcal{N}(\mathbf{x}_{t-1}'; \tilde{\mu}_{t}'(\mathbf{x}_{t}', \mathbf{x}_{0}', \mathbf{r}), \tilde{\beta} \mathbf{I})) \\
      \tilde{\mu}_{t}'(\mathbf{x}_t', \mathbf{x}_{0}', \mathbf{r}) = \frac{1}{\sqrt{\alpha_t}} \left( \mathbf{x}_{t}'(\mathbf{x}_{0}', \mathbf{r}, \mathbf{\epsilon}) - \rho_{t} \mathbf{r} - \frac{\beta_t}{\delta_t} \mathbf{\epsilon}  \right) \\
      \tilde{\beta}_t 
      = \frac{1 - \bar{\alpha}_{t-1}}{1 - \bar{\alpha}_t} \beta_t
      \end{gathered}
  \end{split}
  \label{eq:ddpm_mu_backdoor}
\end{equation}
where $\rho_{t} = (1 - \sqrt{\alpha_t})$, $\delta_t = \sqrt{1 - \bar{\alpha}_t}$, and $\mathbf{x}_{t}'(\mathbf{x}_{0}', \mathbf{r}, \mathbf{\epsilon}) = \sqrt{\bar{\alpha}_t} \mathbf{x}_t + \delta_t \mathbf{r} + \sqrt{1 - \bar{\alpha}_t} \mathbf{\epsilon}$ for $\mathbf{\epsilon} \sim \mathcal{N}(0, \mathbf{I})$, which is a reparametrization of $\mathbf{x}_{t}'$.

We can derive the posterior from scratch.
\begin{equation}
  \begin{split}
      \begin{gathered}
        q(\mathbf{x}_{t-1}' | \mathbf{x}_{t}', \mathbf{x}_{0}') = q(\mathbf{x}_{t}' | \mathbf{x}_{t-1}', \mathbf{x}_{0}') \frac{q(\mathbf{x}_{t-1}' | \mathbf{x}_{0}')}{q(\mathbf{x}_{t}' | \mathbf{x}_{0}')} \\
        \propto \text{exp}\left(- \frac{1}{2} \left( \frac{(\mathbf{x}_{t}' - \rho_{t} \mathbf{r} - \sqrt{\alpha_{t}}\mathbf{x}_{t-1}')^2}{\beta_{t}} - \frac{(\mathbf{x}_{t-1}'  - (1 - \sqrt{\bar{\alpha}_{t-1}}) \mathbf{r} - \sqrt{\bar{\alpha}_{t-1}} \mathbf{x}_{0}')^2}{1 - \bar{\alpha}_{t-1}} + \frac{(\mathbf{x}_{t}' - (1 - \sqrt{\bar{\alpha}_{t}}) \mathbf{r} - \sqrt{\bar{\alpha}_{t}} \mathbf{x}_{0}')^2}{1 - \bar{\alpha}_{t}} \right) \right) \\
      \end{gathered}
  \end{split}
  \label{eq:posterior}
\end{equation}

We gather the terms related to $\mathbf{x}_{t-1}'$ and represent the terms that not involving $\mathbf{x}_{t-1}'$ as $C(\mathbf{x}_{t}', \mathbf{x}_{0}')$
\begin{equation}
  \begin{split}
    \begin{gathered}
      = \exp \left( -\frac{1}{2} \left( \left(\frac{\alpha_t}{\beta_t} + \frac{1}{1 - \bar{\alpha}_{t-1}} \right) \mathbf{x}_{t-1}'^2 
      - 2 \left(\frac{\mathbf{x}_t' \sqrt{\alpha_t}}{\beta_t}
      + \frac{\mathbf{x}_{0}' \sqrt{\bar{\alpha}_{t-1}}}{1 - \bar{\alpha}_{t-1}} + 
      \left( \frac{(1 - \sqrt{\bar{\alpha}_{t-1}})}{1 - \bar{\alpha}_{t-1}} - \frac{\sqrt{\alpha_t} (1 - \sqrt{\alpha_t})}{\beta_t} \right) \mathbf{r} \right) \mathbf{x}_{t-1}' 
      \color{black}{ + C(\mathbf{x}_t', \mathbf{x}_0')} \right) \right)
    \end{gathered}
  \end{split}
  \label{eq:derive1}
\end{equation}

Since we take $q(\mathbf{x}_{t-1}' | \mathbf{x}_{t}', \mathbf{x}_{0}')$ as a Gaussian distribution, we approximate the distribution with mean $\tilde{\mu}_{t}'(\mathbf{x}_{t}', \mathbf{x}_{0}')$ and variance $\tilde{\beta}_{t}$ defined as
\begin{equation}
  \begin{split}
    \begin{gathered}
      \tilde{\beta}_t 
      := \frac{1}{\frac{\alpha_t}{\beta_t} + \frac{1}{1 - \bar{\alpha}_{t-1}}}
      = \frac{1}{\frac{\alpha_t - \bar{\alpha}_t + \beta_t}{\beta_t(1 - \bar{\alpha}_{t-1})}}
      = \frac{1 - \bar{\alpha}_{t-1}}{1 - \bar{\alpha}_t} \beta_t
    \end{gathered}
  \end{split}
  \label{eq:derive2}
\end{equation}

To derive the mean, we reparametrize the random variable $\mathbf{x}_{t}' = \mathbf{x}_{t}'(\mathbf{x}_{0}', \mathbf{r}, \mathbf{\epsilon})$. Here we mark the additional terms of BadDiffusion in red. We can see that BadDiffusion adds a correction term to the diffusion process. We mark the correction term of BadDiffusion as red.
\begin{equation}
  \begin{split}
    \begin{gathered}
      \tilde{\mathbf{\mu}}_t' (\mathbf{x}_t', \mathbf{x}_0')
      := \left(\left(\frac{\sqrt{\alpha_t}}{\beta_t} \mathbf{x}_{t}'(\mathbf{x}_{0}', \mathbf{r}, \mathbf{\epsilon}) + \frac{\sqrt{\bar{\alpha}_{t-1} }}{1 - \bar{\alpha}_{t-1}} \mathbf{x}_0' \right) 
      + \color{red}{\left( \frac{1 - \sqrt{\bar{\alpha}_{t-1}}}{1 - \bar{\alpha}_{t-1}} - \frac{\sqrt{\alpha_t} (1 - \sqrt{\alpha_t})}{\beta_t} \right) \mathbf{r}} \right)
      \color{black}{/ (\frac{\alpha_t}{\beta_t} + \frac{1}{1 - \bar{\alpha}_{t-1}})} \\
    \end{gathered}
  \end{split}
  \label{eq:mean1}
\end{equation}
\begin{equation}
  \begin{split}
    \begin{gathered}
      = \left( \left(\frac{\sqrt{\alpha_t}}{\beta_t} \mathbf{x}_{t}'(\mathbf{x}_{0}', \mathbf{r}, \mathbf{\epsilon}) + \frac{\sqrt{\bar{\alpha}_{t-1} }}{1 - \bar{\alpha}_{t-1}} \mathbf{x}_0' \right) 
      + \color{red}{\left(\frac{1 - \sqrt{\bar{\alpha}_{t-1}}}{1 - \bar{\alpha}_{t-1}} - \frac{\sqrt{\alpha_t} (1 - \sqrt{\alpha_t})}{\beta_t} \right) \mathbf{r}} \right)
      \frac{1 - \bar{\alpha}_{t-1}}{1 - \bar{\alpha}_t} \cdot \beta_t \\
    \end{gathered}
  \end{split}
  \label{eq:mean2}
\end{equation}
\begin{equation}
  \begin{split}
    \begin{gathered}
      = \left(\frac{\sqrt{\alpha_t} \left(1 - \bar{\alpha}_{t-1} \right)}{1 - \bar{\alpha}_t} \mathbf{x}_{t}'(\mathbf{x}_{0}', \mathbf{r}, \mathbf{\epsilon}) 
      + \frac{\sqrt{\bar{\alpha}_{t-1}}\beta_t}{1 - \bar{\alpha}_t} \mathbf{x}_0' \right)
      \color{red}{+ \left( \frac{\beta_t (1 - \sqrt{\bar{\alpha}_{t-1}})}{1 - \bar{\alpha}_{t}} - \frac{\sqrt{\alpha_t} (1 - \sqrt{\alpha_t}) (1 - \bar{\alpha}_{t-1})}{1 - \bar{\alpha}_t} \right)} \mathbf{r} \\
    \end{gathered}
  \end{split}
  \label{eq:mean3}
\end{equation}
Replace $\mathbf{x}_{0}'$ with the $\frac{1}{\sqrt{\bar{\alpha}}_t} (\mathbf{x}_{t}'(\mathbf{x}_{0}', \mathbf{r}, \mathbf{\epsilon}) - (1 - \sqrt{\bar{\alpha}_t}) \mathbf{r} - \sqrt{1 - \bar{\alpha}}_t \epsilon)$, which is the reparametrization of $\mathbf{x}_{0}'$ derived from $\mathbf{x}_{t}'(\mathbf{x}_{0}', \mathbf{r}, \mathbf{\epsilon})$.
\begin{equation}
  \begin{split}
    \begin{gathered}
      = \left(\frac{\sqrt{\alpha_t} \left(1 - \bar{\alpha}_{t-1} \right)}{1 - \bar{\alpha}_t} \mathbf{x}_{t}'(\mathbf{x}_{0}', \mathbf{r}, \mathbf{\epsilon}) 
      + \frac{\sqrt{\bar{\alpha}_{t-1}}\beta_t}{1 - \bar{\alpha}_t} \left( \frac{1}{\sqrt{\bar{\alpha}}_t} (\mathbf{x}_{t}'(\mathbf{x}_{0}', \mathbf{r}, \mathbf{\epsilon}) - \sqrt{1 - \bar{\alpha}}_t \epsilon) \right) \right) \\
      \color{red}{+ \left( \frac{\beta_t (1 - \sqrt{\bar{\alpha}_{t-1}})}{1 - \bar{\alpha}_{t}} - \frac{\sqrt{\alpha_t} (1 - \sqrt{\alpha_t}) (1 - \bar{\alpha}_{t-1})}{1 - \bar{\alpha}_t} - \frac{\sqrt{\bar{\alpha}_{t-1}} \beta_t (1 - \sqrt{\bar{\alpha}_{t}})}{(1 - \bar{\alpha}_t) \sqrt{\bar{\alpha}_t}} \right)} \mathbf{r} \\
    \end{gathered}
  \end{split}
  \label{eq:mean4}
\end{equation}
\begin{equation}
  \begin{split}
    \begin{gathered}
      = \frac{1}{\sqrt{\alpha_t}} \left( \mathbf{x}_{t}'(\mathbf{x}_{0}', \mathbf{r}, \mathbf{\epsilon}) - \frac{\beta_t}{\sqrt{1 - \bar{\alpha}_t}} \epsilon \right) \\
      \color{red}{+ \left( \frac{\beta_t (\sqrt{\alpha}_{t} - \cancel{\sqrt{\bar{\alpha}_{t}}}) - \alpha_{t} (1 - \sqrt{\alpha_{t}}) (1 - \bar{\alpha}_{t-1}) - \beta_{t} (1 - \cancel{\sqrt{\bar{\alpha}_{t}}})}{(1 - \bar{\alpha}_{t}) \sqrt{\alpha_{t}}} \right)} \mathbf{r} \\
    \end{gathered}
  \end{split}
  \label{eq:mean5}
\end{equation}
\begin{equation}
  \begin{split}
    \begin{gathered}
      = \frac{1}{\sqrt{\alpha_t}} \left( \mathbf{x}_{t}'(\mathbf{x}_{0}', \mathbf{r}, \mathbf{\epsilon}) - \frac{\beta_t}{\sqrt{1 - \bar{\alpha}_t}} \epsilon \right) \color{red}{+ \left( \frac{\beta_t (\sqrt{\alpha}_{t} - 1) - (\sqrt{\alpha_{t}} - 1) (\bar{\alpha}_{t} - \alpha_{t})}{(1 - \bar{\alpha}_{t}) \sqrt{\alpha_{t}}} \right)} \mathbf{r} \\
    \end{gathered}
  \end{split}
  \label{eq:mean6}
\end{equation}
\begin{equation}
  \begin{split}
    \begin{gathered}
      = \frac{1}{\sqrt{\alpha_t}} \left( \mathbf{x}_{t}'(\mathbf{x}_{0}', \mathbf{r}, \mathbf{\epsilon}) - \frac{\beta_t}{\sqrt{1 - \bar{\alpha}_t}} \epsilon \right) \color{red}{+ \left( \frac{(\sqrt{\alpha_{t}} - 1) \cancel{(1 - \bar{\alpha}_{t})}}{\cancel{(1 - \bar{\alpha}_{t})} \sqrt{\alpha_{t}}} \right)} \mathbf{r} \\
    \end{gathered}
  \end{split}
  \label{eq:mean7}
\end{equation}
\begin{equation}
  \begin{split}
    \begin{gathered}
      = \frac{1}{\sqrt{\alpha_t}} \left( \mathbf{x}_{t}'(\mathbf{x}_{0}', \mathbf{r}, \mathbf{\epsilon}) - \frac{\beta_t}{\sqrt{1 - \bar{\alpha}_t}} \epsilon \right) \color{red}{- \frac{1}{\sqrt{\alpha_t}} (1 - \sqrt{\alpha_t}) \mathbf{r}} \\
    \end{gathered}
  \end{split}
  \label{eq:mean8}
\end{equation}
Denote $\rho_t = 1 - \sqrt{\alpha_t}$ and we get
\begin{equation}
  \begin{split}
    \begin{gathered}
      = \frac{1}{\sqrt{\alpha_t}} \left( \mathbf{x}_{t}'(\mathbf{x}_{0}', \mathbf{r}, \mathbf{\epsilon}) -  \color{red}{\rho_{t} \mathbf{r}} \color{black}{ - \frac{\beta_t}{\sqrt{1 - \bar{\alpha}_t}}} \epsilon \right) \\
    \end{gathered}
  \end{split}
  \label{eq:mean9}
\end{equation}

\end{document}